\newif\ifanonymized
\anonymizedfalse

\newif\ifconfidential
\confidentialfalse

\newif\ifauthorversion
\authorversiontrue

\ifanonymized
    \documentclass[acmtog, anonymous, review]{acmart}
\else
    \ifauthorversion
        \documentclass[acmtog, authorversion, screen]{acmart}
    \else
        \documentclass[acmtog, screen]{acmart}
    \fi
\fi

\usepackage{subcaption} 
\usepackage{stfloats}

\usepackage{mathtools}

\DeclarePairedDelimiter{\norm}{\lVert}{\rVert}

\DeclareMathOperator*{\minimize}{minimize}
\DeclareMathOperator*{\smooth}{smooth}

\usepackage{tikz}  
\usetikzlibrary{calc,spy}
\tikzstyle{closeup} = [
  opacity=1.0,          
  height=2.5cm,
  width=2.5cm,
  connect spies, red  
]
\tikzstyle{largewindow} = [red, line width=0.20mm]
\tikzstyle{smallwindow} = [red, line width=0.40mm]
\tikzstyle{aalsmallwindow1} = [red, line width=0.40mm]
\tikzstyle{aalsmallwindow2} = [blue, line width=0.40mm]
\tikzstyle{aalcloseup1} = [
  opacity=1.0,          
  height=2.25cm,
  width=2.25cm,
  connect spies, red  
]
\tikzstyle{aalcloseup2} = [
  opacity=1.0,          
  height=2.25cm,
  width=2.25cm,
  connect spies, blue  
]
\tikzstyle{aallargewindow1} = [red, line width=0.50mm]
\tikzstyle{aalsmallwindow1} = [red, line width=0.40mm]
\tikzstyle{aallargewindow2} = [blue, line width=0.50mm]
\tikzstyle{aalsmallwindow2} = [blue, line width=0.40mm]
\tikzstyle{aapsmallwindow1} = [red, line width=0.40mm]
\tikzstyle{aapsmallwindow2} = [blue, line width=0.40mm]
\tikzstyle{aapsmallwindow3} = [green, line width=0.40mm]
\tikzstyle{aapcloseup1} = [
  opacity=1.0,          
  height=2.25cm,
  width=2.25cm,
  connect spies, red  
]
\tikzstyle{aapcloseup2} = [
  opacity=1.0,          
  height=2.25cm,
  width=2.25cm,
  connect spies, blue  
]
\tikzstyle{aapcloseup3} = [
  opacity=1.0,          
  height=4.65cm,
  width=2.65cm,
  connect spies, green  
]
\tikzstyle{aaplargewindow1} = [red, line width=0.50mm]
\tikzstyle{aapsmallwindow1} = [red, line width=0.40mm]
\tikzstyle{aaplargewindow2} = [blue, line width=0.50mm]
\tikzstyle{aapsmallwindow2} = [blue, line width=0.40mm]
\tikzstyle{aaplargewindow3} = [green, line width=0.50mm]
\tikzstyle{aapsmallwindow3} = [green, line width=0.40mm]

\definecolor{darkgreen}{rgb}{0,0.5,0}
\definecolor{orange}{rgb}{1,0.5,0}
\definecolor{purple}{rgb}{0.5,0.2,0.8}
\definecolor{darkred}{rgb}{0.5,0,0}
\definecolor{Yellow}{rgb}{1,1, 0.6}
\definecolor{Orange}{rgb}{1,0.8, 0.6}
\definecolor{Red}{rgb}{1, 0.6, 0.6}

\newcommand{\sect}[1]{Section~\ref{#1}}
\newcommand{\fig}[1]{Figure~\ref{#1}}

\newcommand{\tacc}[1]{#1}

\newcommand{\myvspace}[1]{\vspace{#1}}

\newcommand{\gconfidential}{\color{red}{GOOGLE CONFIDENTIAL}}

\newcommand{\ourpapertitle}{Computational Long Exposure Mobile Photography}

\setcopyright{acmlicensed}

\ifanonymized
    \acmSubmissionID{305}
\fi

\acmJournal{TOG}
\acmYear{2023}
\acmVolume{42}
\acmNumber{4}
\acmArticle{48}
\acmMonth{8}
\acmPrice{15.00}

\acmDOI{10.1145/3592124}

\ifconfidential
    \title[\ourpapertitle - \gconfidential]{\ourpapertitle}
\else
    \title{\ourpapertitle}
\fi

\ifanonymized
\else
    \author{Eric Tabellion}
\author{Nikhil Karnad}
\author{Noa Glaser}
\author{Ben Weiss}
\author{David E. Jacobs}
\author{Yael Pritch}

\affiliation{
    \institution{\\Google Research}
    \streetaddress{1600 Amphitheatre Parkway}
    \city{Mountain View}
    \state{CA}
    \postcode{94043}
    \country{USA}
}

\authorsaddresses{%
    Authors’ address:
    \href{https://www.tabellion.org/et/index.html}{Eric Tabellion},
    Nikhil Karnad, Noa Glaser, Ben Weiss, David E. Jacobs, Yael Pritch,
    Google Inc., 1600 Amphitheatre Parkway, Mountain View, CA, 94043, USA
}

\fi

\begin{document}

\begin{teaserfigure}
\centering
    \begin{subfigure}[t]{1.0\textwidth}
        \begin{subfigure}[t]{0.49\columnwidth}
            \centering
            \begin{subfigure}[t]{\columnwidth}
                \centering
                \includegraphics[width=0.495\columnwidth]{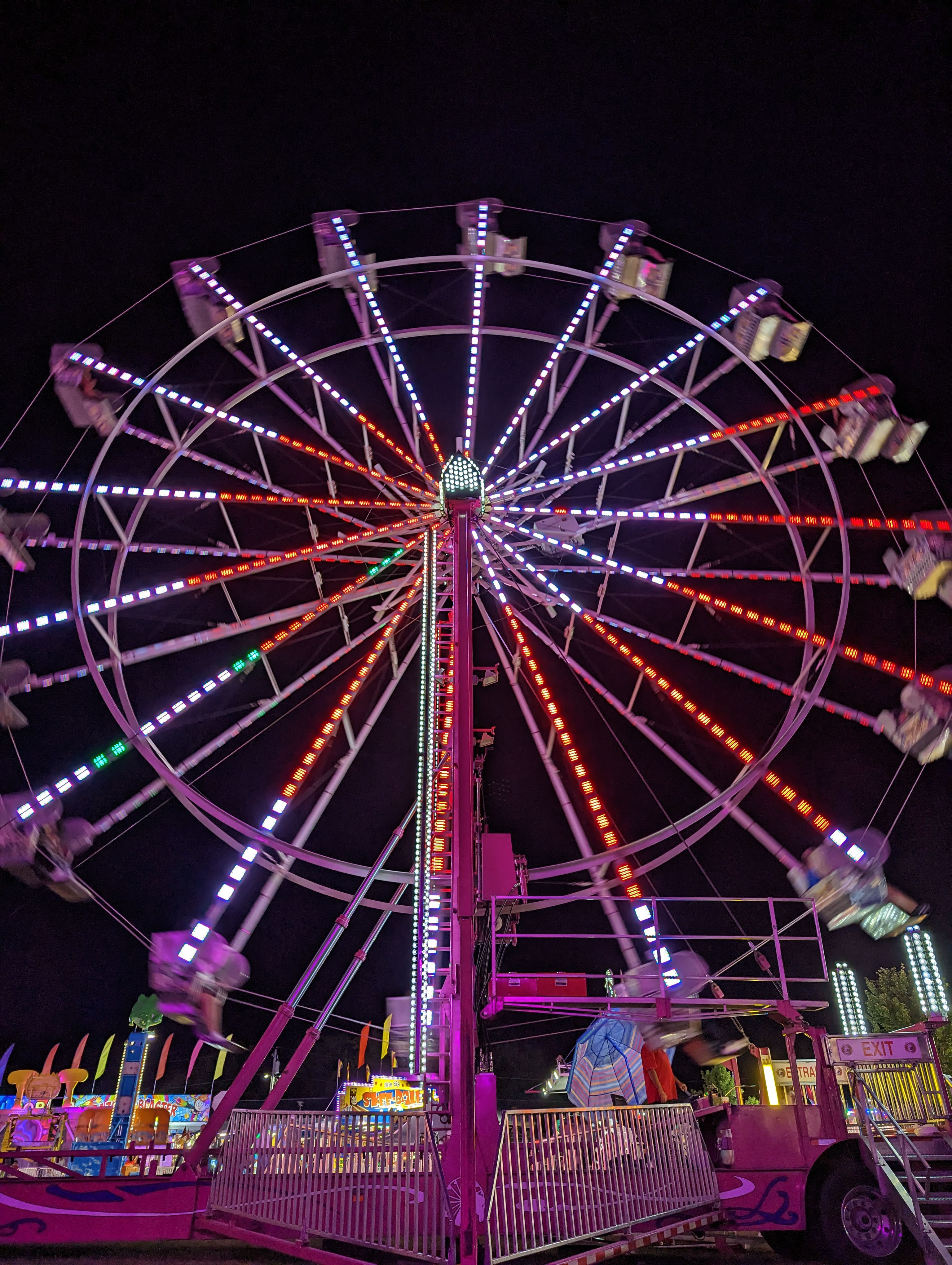}
                \hfill
                \includegraphics[width=0.495\columnwidth]{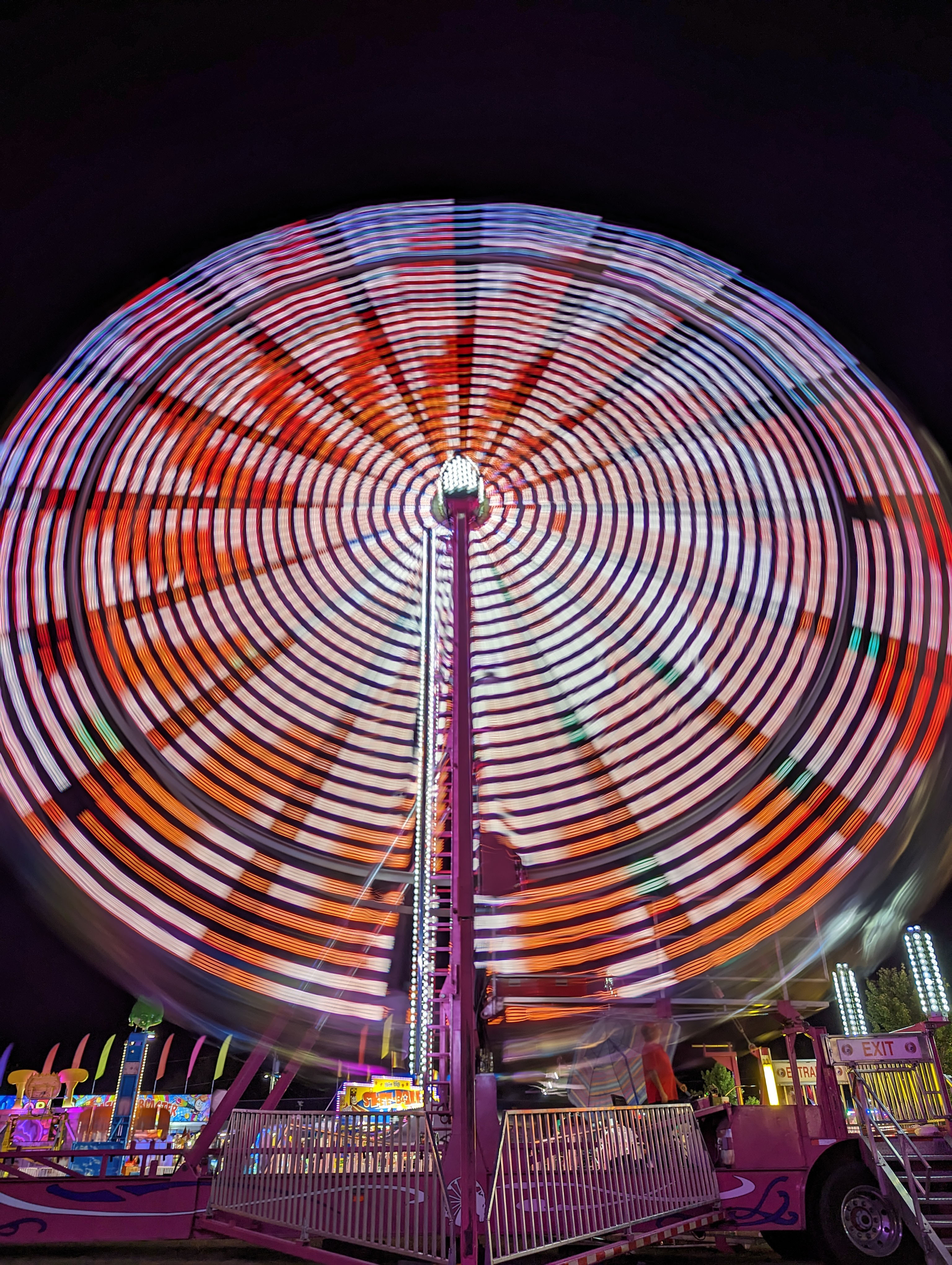}
                \myvspace{-1.1em}
            \end{subfigure}
            \begin{subfigure}[t]{\columnwidth}
                \centering
                \begin{subfigure}[t]{0.495\columnwidth}
                    \includegraphics[width=1.0\columnwidth]{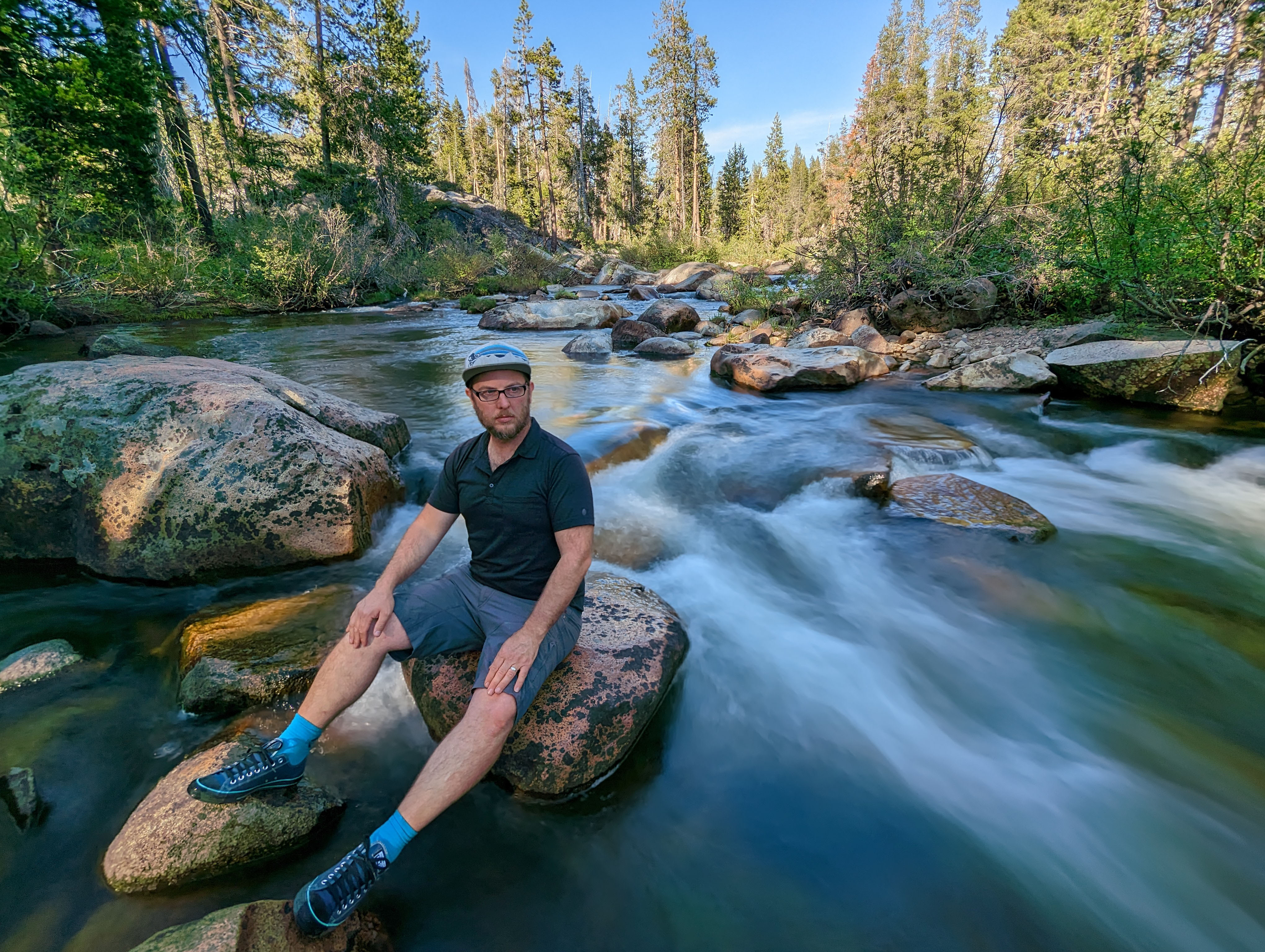}
                \end{subfigure}
                \hfill
                \begin{subfigure}[t]{0.495\columnwidth}
                    \includegraphics[width=1.0\columnwidth]{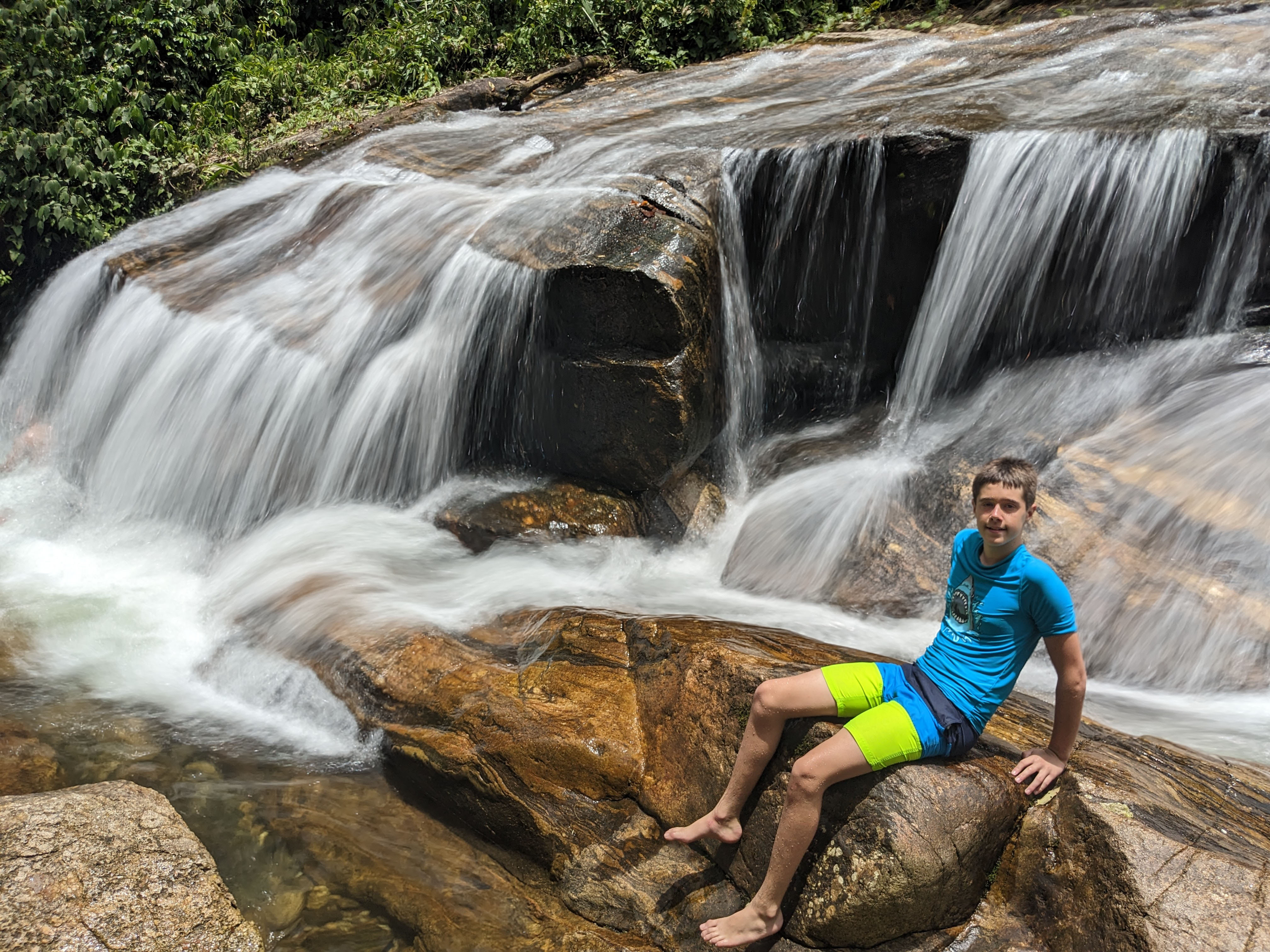}
                \end{subfigure}
            \end{subfigure}
            \myvspace{-1.5em}
            \subcaption{Foreground blur examples}
        \end{subfigure}
        \hfill
        \begin{subfigure}[t]{0.49\columnwidth}
            \centering
            \begin{subfigure}[t]{\columnwidth}
                \centering
                \includegraphics[width=0.495\columnwidth]{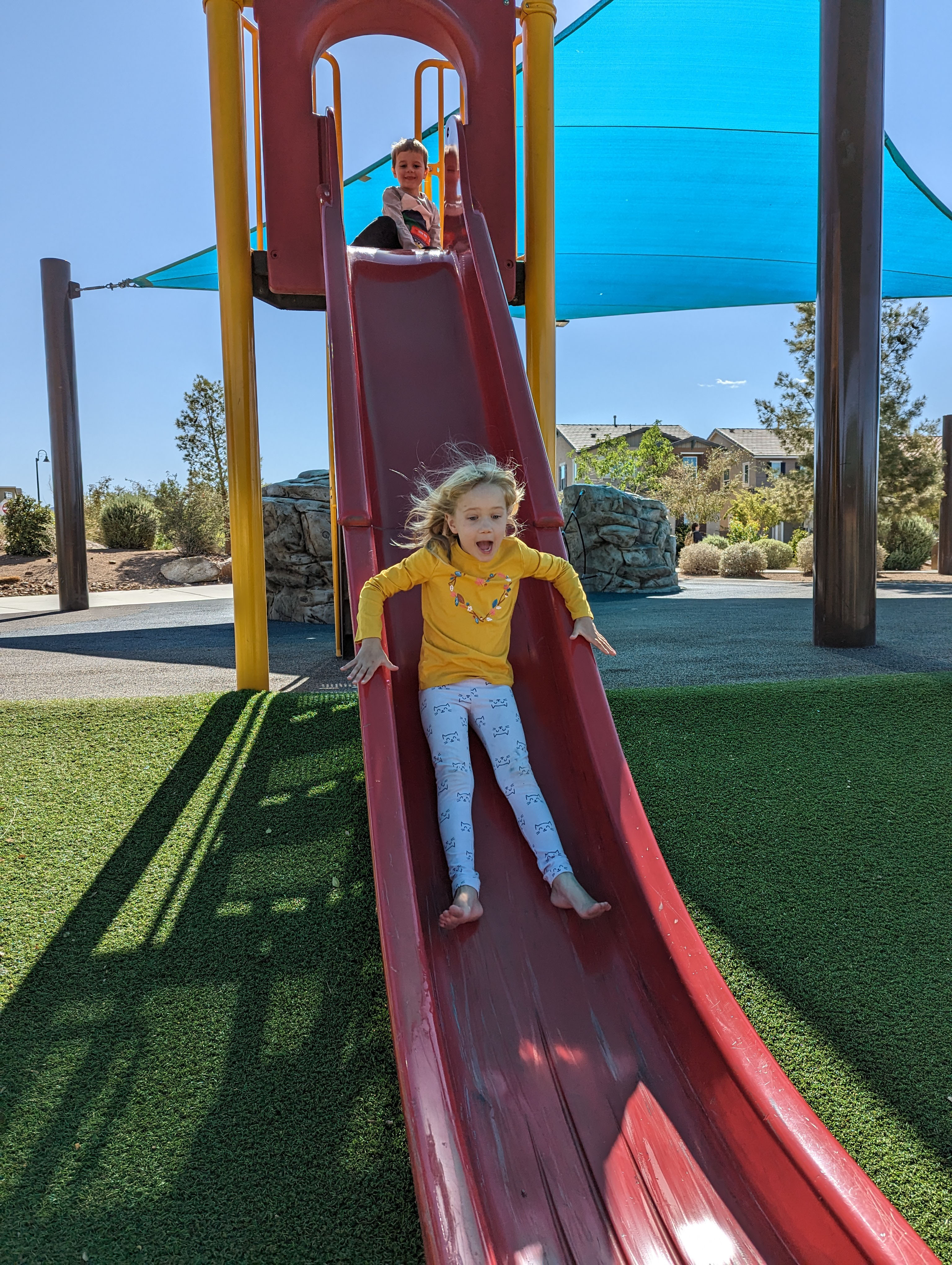}
                \hfill
                \includegraphics[width=0.495\columnwidth]{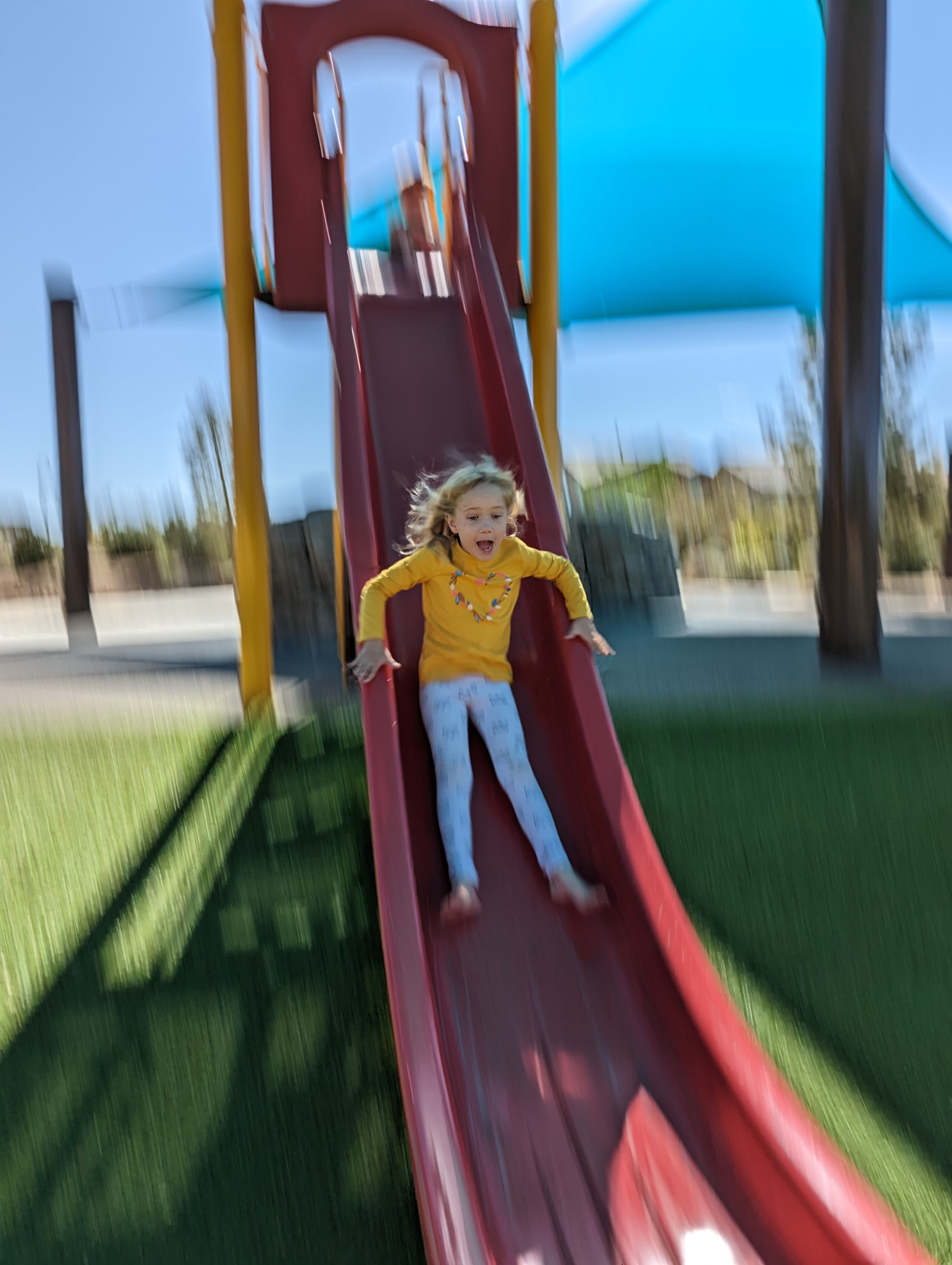}
                \myvspace{-1.1em}
            \end{subfigure}
            \begin{subfigure}[t]{\columnwidth}
                \centering
                \begin{subfigure}[t]{0.495\columnwidth}
                    \includegraphics[width=1.0\columnwidth]{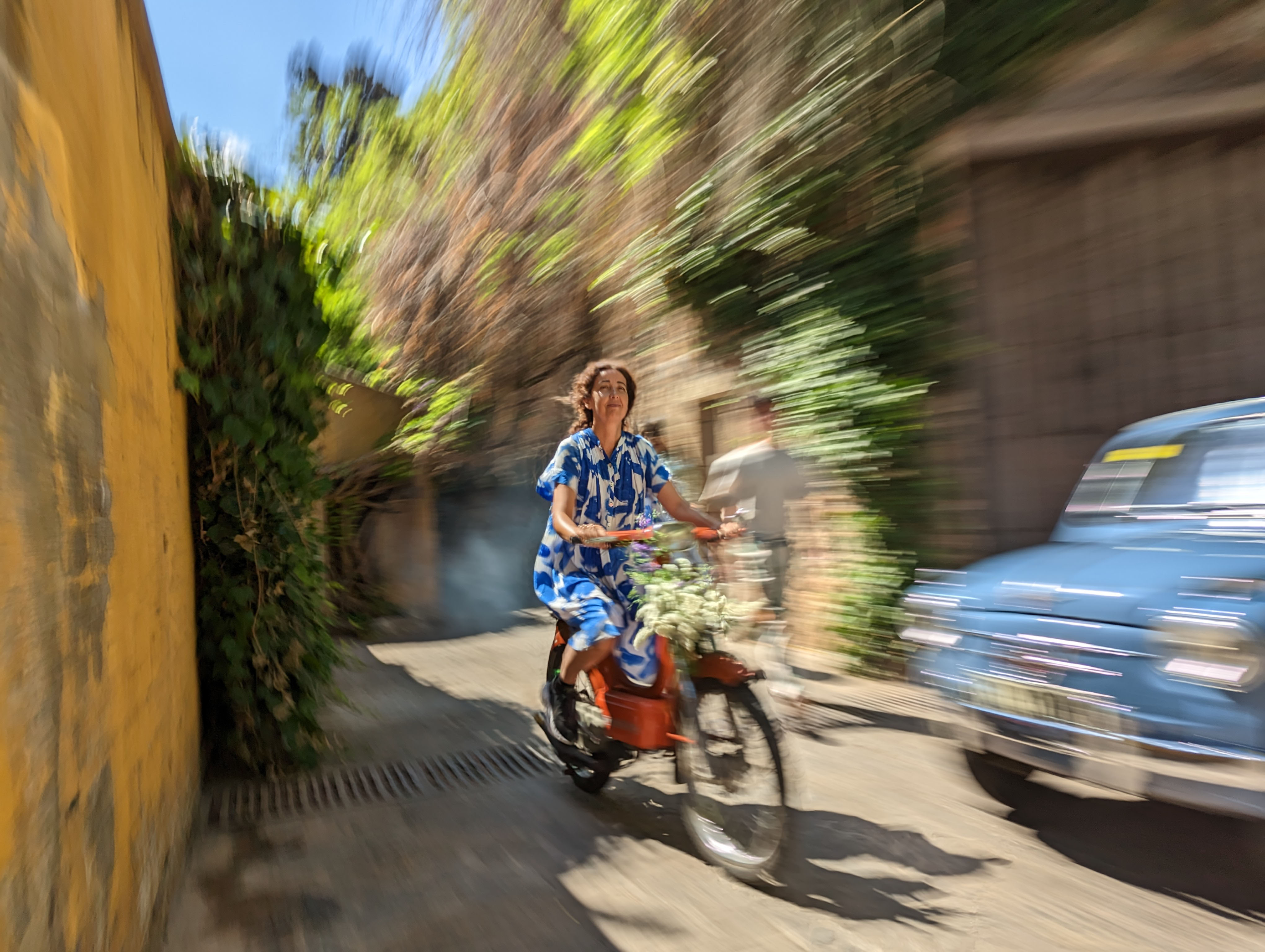}
                \end{subfigure}
                \hfill
                \begin{subfigure}[t]{0.495\columnwidth}
                    \includegraphics[width=1.0\columnwidth]{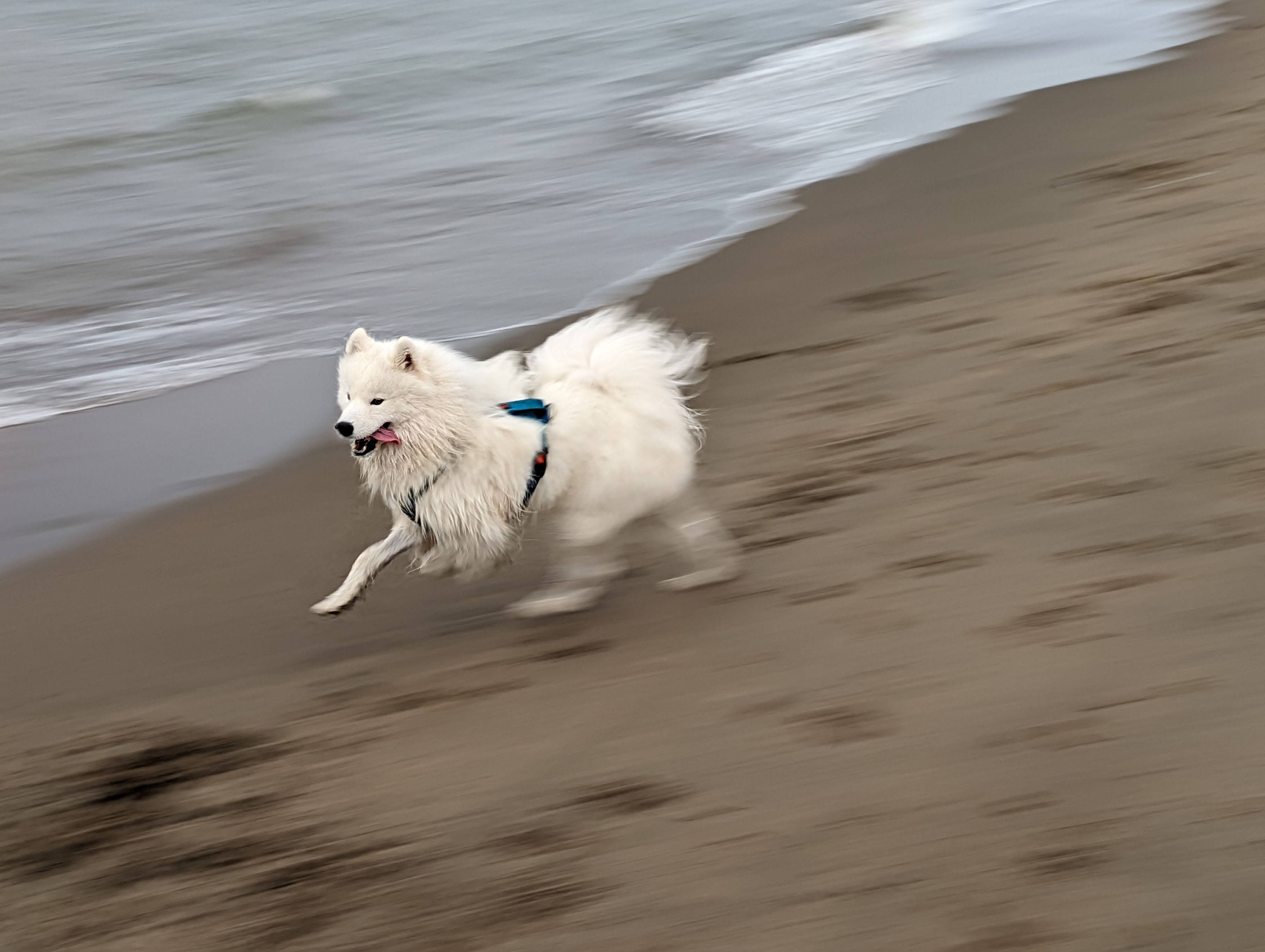}
                \end{subfigure}
            \end{subfigure}
            \myvspace{-1.5em}
            \subcaption{Background blur examples}
        \end{subfigure}
    \end{subfigure}
    \myvspace{-1em}
    \caption{At the tap of the shutter button, our smartphone camera system captures, processes and outputs both conventional and long exposure corresponding photographs in a few seconds, as shown in the top row. Examples of foreground blur captured hand-held are shown in (a), and examples of background blur captured without precise tracking, are shown in (b). Our computational long exposure photography pipeline handles both use cases fully automatically.}
    \myvspace{+0.15em}
    \label{fig:teaser}
\end{teaserfigure}

\begin{abstract}
Long exposure photography produces stunning imagery, representing moving elements in a scene with motion-blur. It is generally employed in two modalities, producing either a foreground or a background blur effect. Foreground blur images are traditionally captured on a tripod-mounted camera and portray blurred moving foreground elements, such as silky water or light trails, over a perfectly sharp background landscape. Background blur images, also called panning photography, are captured while the camera is tracking a moving subject, to produce an image of a sharp subject over a background blurred by relative motion. Both techniques are notoriously challenging and require additional equipment and advanced skills. In this paper, we describe a computational burst photography system that operates in a hand-held smartphone camera app, and achieves these effects fully automatically, at the tap of the shutter button. Our approach first detects and segments the salient subject. We track the scene motion over multiple frames and align the images in order to preserve desired sharpness and to produce aesthetically pleasing motion streaks. We capture an under-exposed burst and select the subset of input frames that will produce blur trails of controlled length, regardless of scene or camera motion velocity. We predict inter-frame motion and synthesize motion-blur to fill the temporal gaps between the input frames. Finally, we composite the blurred image with the sharp regular exposure to protect the sharpness of faces or areas of the scene that are barely moving, and produce a final high resolution and high dynamic range (HDR) photograph. Our system democratizes a capability previously reserved to professionals, and makes this creative style accessible to most casual photographers.

\ifauthorversion
    \tacc{
    \myvspace{1em}
    More information can be found on our project webpage: \url{https://motion-mode.github.io/}.
    }
\fi
\end{abstract}



\maketitle

\section{Introduction}
\label{sec:introduction}

Mobile photography is ever present in consumers' daily lives and is often superseding traditional photography. Using image burst capture and post-processing techniques, modern mobile phones' imaging pipelines produce very high quality results, providing high dynamic range tone mapping, exceptional low light performance and simulating depth-of-field bokeh effects, which were previously achievable only with much bigger and heavier cameras and lenses.

Despite these outstanding improvements, long exposure mobile photography remains poorly treated to the best of our knowledge. Existing solutions don't help users produce results where moving and static scene elements appear blurry and sharp respectively. This juxtaposition of sharp against blurry is a key property of a compelling image, that cannot be achieved by simply exposing a hand-held camera sensor for a longer duration.

Traditional long exposure photography is typically performed in one of two ways, according to the scene and situation. One approach produces a foreground blur effect (e.g. silky waterfall, light trails, etc.) over a sharp background, using very long exposure times that can last up to several seconds. This requires using a tripod, as even a slight camera shake can cause undesired loss of background sharpness. Additionally, a neutral density (ND) filter must be added to the lens, to avoid over-exposing the sensor. A second approach, called panning photography, produces a rendition with a sharp moving subject over a background that is blurred with motion relative to the subject. It is achieved by tracking the moving subject with the camera, while keeping the shutter open with the exposure time increased modestly, e.g. half a second, and the aperture slightly reduced to avoid over-exposing the image. The photographer must track the subject motion as precisely as possible to avoid undesired loss of subject sharpness, while also pressing the shutter button at the right moment. Both approaches require advanced skills, practice and choosing the camera shutter speed manually, taking into account how fast the scene is moving to achieve the desired result.

The main contribution of this paper is a computational long exposure mobile photography system, implemented in two variants, which democratize the two aforementioned use cases. It is implemented in a
\ifanonymized
new smartphone camera mode (anonymized during review),
\else
new camera mode called "Motion Mode" on Google Pixel 6 and 7 smartphones,
\fi
which allows the user to easily capture these effects, without the need for a tripod or lens filter, nor the need to track the moving subject precisely or at all. Our method is fully automatic end-to-end within each variant: after the user chooses which of foreground or background blur result they wish to produce, we generate long exposure 12 megapixel photographs at the tap of the shutter button, while compensating for camera and/or subject motion, thereby preserving desired background and subject sharpness. The main components of our system are:
\begin{itemize}
    \item Capture schedule and frame selection, producing normalized blur trail lengths independent of scene or camera velocity,
    \item Subject detection that combines gaze saliency with people and pets face region predictions, and tracking of their motion,
    \item Alignment of input images to cancel camera hand-shake, stabilize the background in the presence of moving foreground elements, or to annul subject motion while producing pleasing background motion blur trails,
    \item Dense motion prediction and blur synthesis, spanning multiple high resolution input frames and producing smooth curved motion blur trails with highlight preservation.
\end{itemize}

Furthermore, our system architecture, which includes several neural networks, performs efficiently on a mobile device under constrained compute and memory budgets, implementing an HDR imaging pipeline that produces both related conventional and long exposure results in just a few seconds.
\ifauthorversion
\else
    More information can be found on our project webpage: \url{https://motion-mode.github.io/}.
\fi

\section{Related Work}
\label{sec:related_work}


\subsection{Mobile Phone Computational Photography}
Many computational photography advances in recent years define today's mobile photography capabilities. The work from \citet{Hasinoff16} describes a mobile camera pipeline that captures, aligns and merges bursts of under-exposed raw images. Combined with the work of~\citet{Wronski19}, they are able to strongly improve the Signal to Noise Ratio (SNR), dynamic range and image detail, overcoming the limitations of small smartphone sensors and lenses. Our system is built on top of such a computational imaging foundation.

To handle very low light situations without using a flash, \citet{Liba19} employ a scene motion metering approach to adjust the number of input burst frames and determine their exposure time.
Similarly, we adjust the frame capture schedule based on scene motion, estimated when the shutter button is pressed, for the purpose of normalizing the resulting amount of motion-blur.

Since the small camera lenses used on smartphones cannot produce shallow depth-of-field effects optically, \citet{Wadhwa18} design a synthetic bokeh approach, that relies on a semantic person segmentation neural network, with the intent to isolate a subject from a distracting background. Our system is analogous, as we strive to isolate a subject from the relative motion of the background, while attempting to emphasize the dynamic nature of the scene.


\subsection{Auto-tracking a Subject (background blur)}
Determining the subject of a background blur capture is a hard problem. Many synthetic long exposure pipelines avoid it altogether by requiring manually tagging the subject region, or using a heuristic such as the middle region of the image~\cite{Lancelle19,Mikamo21,Luo18}. In contrast, we present a pipeline which determines the subject region automatically by predicting visual saliency and face regions.

Using saliency-driven image edits to highlight a main subject from a distracting background was introduced in~\cite{Aberman22}. Existing methods to detect and track subject motion over time include~\cite{Stengel15}, which use gaze saliency prediction to detect the subject and optical flow to track its motion, and~\cite{Mikamo21}, which require the user to specify the subject region using a bounding box and similarly track its motion. In our work, we detect the subject using a combination of gaze saliency and semantic segmentation using a highly efficient mobile architecture inspired by~\cite{Bazarevsky19}. We track its motion using feature tracking, and introduce an alignment regularization term to result in more visually pleasing motion-blur trails, which are more consistent with overall subject motion.


\subsection{Stabilizing the Background (foreground blur)}
Images captured by handheld cameras are often shaky and may often contain parallax. In the foreground-blur case, we need to stabilize the background to keep the resulting regions sharp. This can be solved for using structure-from-motion (SFM) techniques~\cite{Liu2014} to construct a 3-d representation of the scene~\cite{Hartley2004}, then a stabilized camera path can be solved for in 3-d, and the scene finally re-rendered using the new smooth path~\cite{Liu2009content}. However, these techniques rely on 3-d reconstruction, for which a fast and robust implementation is challenging~\cite{Liu2011subspace}. At the other end of the spectrum are 2-d stabilization techniques that use much simpler motion models such as global homographies or affine transformations~\cite{Morimoto1998}. These techniques are fast and robust, but cannot model effects such as parallax, rolling-shutter, or lens distortion. There is a large body of work that extends these 2-d methods, such as using gyroscopic sensors only~\cite{karpenko2011digital}, gyroscopes with face detection~\cite{Shi2019steadiface}, targeted crops~\cite{Grundmann2011} and trajectory filtering~\cite{Liu2011subspace}. Our method is analogous to techniques that start with 2-d feature trajectories to estimate per-frame global transformations and refine this estimate with spatially varying image warps~\cite{Zaragoza2013apap, Liu2013bundled} to achieve the desired trade-off between speed and robustness.

\begin{figure*}[htb]
  \centering
  \includegraphics[width=1.0\textwidth]{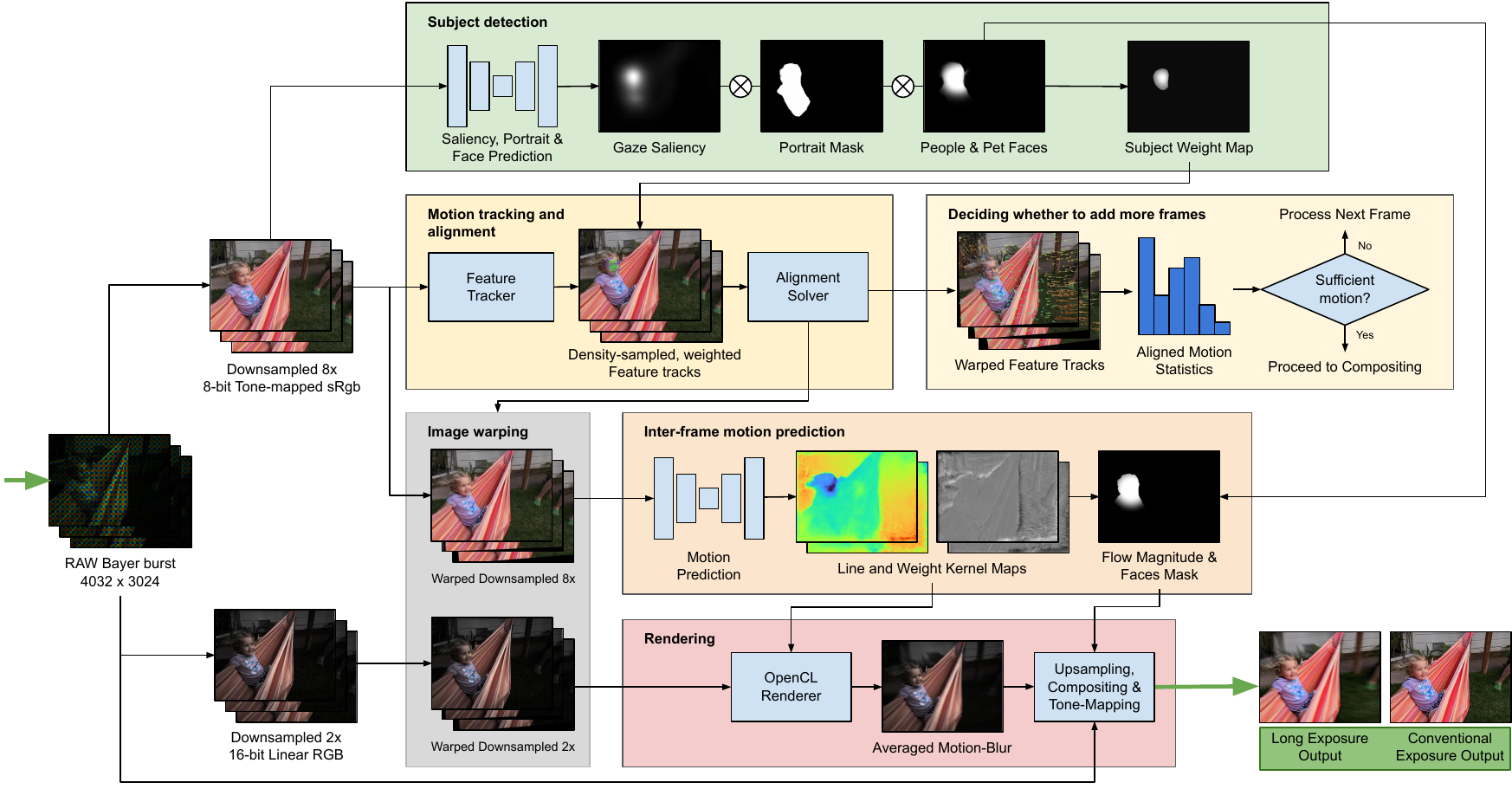}
  \myvspace{-1.75em}
  \caption{Our system processes raw burst images incrementally, from left to right on this diagram, first at low resolution (8x downsampling) for initial subject predictions, feature tracking, motion analysis and motion prediction. Images are also processed at half resolution (2x downsampling) to synthesize motion-blur before being upsampled back to full 12 megapixel resolution for compositing and tone-mapping the final results. More detail is provided in \sect{sec:system_overview}.
  }
  \label{fig:system_overview}
  \myvspace{-0.5em}
\end{figure*}


\subsection{Synthesizing Motion Trails}

There is a large body of prior work on synthesizing motion-blur, in the context of non-photorealistic rendering~\cite{Lee09}, stop-motion animation~\cite{Brostow01}, or 3D computer graphics rendering in real-time~\cite{Ronnow21} or offline~\cite{Lehtinen11,Navarro11}. There is work describing single photograph post-processing interactive applications to create artistic motion effects~\cite{Teramoto10,Luo18,Luo20} or that process multiple previously stabilized images and can achieve non-physical renditions~\cite{Salamon19}.

Our work compares more directly to prior work on computational long exposure from multiple photographs or video frames. \citet{Lancelle19} describe a pipeline that can handle both foreground and background blur effects, but requires substantial user interaction to handle all the cases. Like other approaches, they require significant compute time in an offline processing application, as they rely on expensive optical-flow based image warping or frame interpolation, to synthesize smooth motion-blur spanning the input frames pair-wise. In contrast, our pipeline is fully automatic, is integrated in a smartphone photo camera and produces the result in a few seconds at 12 megapixel resolution. To synthesize motion-blur, we use a line kernel prediction neural network, derived from~\cite{Brooks19}, combined with a GPU rendering algorithm that can handle the input image alignment warping implicitly, while producing smooth and curved motion-blur trails spanning multiple input frames.

\section{System overview}
\label{sec:system_overview}

A diagram of our computational long-exposure system is shown in \fig{fig:system_overview}. The stream of captured raw images is processed incrementally at two different resolutions through four stages, each corresponding to a row of the diagram in \fig{fig:system_overview}: initial subject detection, motion analysis, motion prediction and rendering. The initial saliency and face prediction stage (\sect{sec:subject_detection}) computes the main signals for our subject detection, producing a normalized weight map. The motion analysis stage is responsible for tracking (\sect{sec:motion_tracking}) and aligning (\sect{sec:image_alignment}) a set of feature points corresponding to the detected subject or to the background, and for \tacc{selecting frames based on motion statistics (\sect{sec:frame_selection})}. The motion prediction stage (\sect{sec:motion_prediction}) predicts dense line kernel and weight maps, that are used in the rendering stage (\sect{sec:rendering}) to produce smooth motion-blur spanning a given input frame pair. \tacc{The final compositing stage (\sect{sec:compositing}) layers the final results while preserving the sharpness of important areas in the final image.}

The first three stages use as their input, images that have been tone-mapped and converted to sRGB, downsampled by a factor of 8 to a low resolution of 504 x 376. This resolution is chosen to achieve low latency when processing frames, which is dominated by the dense motion prediction neural network. This also ensures that the receptive field covers most practical motion disparities in the input full resolution images. \tacc{The last stage however, uses the intentionally under-exposed burst raw images converted to 16-bit linear RGB. The high bit-depth is necessary to preserve the scene's high dynamic range during rendering, i.e. to avoid saturating the highlights and banding artifacts in the shadows. Images are downsampled by a factor of 2 as a trade-off that preserves enough detail in the final result while operating within a reduced memory footprint.}

The incremental processing loop converts and downsamples an additional burst frame at each iteration, feeding the three last stages of our pipeline and resulting in an accumulated averaged motion-blur image. The loop stops when \tacc{the frame selection criteria is reached, using an estimate of motion-blur trails' length. We then composite the final results, while upsampling images back to full resolution. At the end of our pipeline, images are converted to a low dynamic range 8-bit representation, using tone-mapping to preserve the high dynamic range visual appearance of the scene.}

\section{Implementation}
\label{sec:implementation}

\subsection{Burst Capture}
\label{sec:burst_capture}

\tacc{Our camera system captures frames at a rate of 30 frames per second, using fully automatic aperture, shutter speed, and focus settings. These settings may adjust dynamically to scene changes across frames, and our system performs well with all but abrupt variations.}

In the background blur case, we target scenes with fast moving nearby subjects. When tapping the shutter button, the most recently captured frame is used as the base frame to produce the conventional exposure~\cite{Hasinoff16}, as it depicts the peak of the action chosen by the user. \tacc{Up to 8 a}dditional frames in the past may then be selected by our algorithm to produce the background blur effect.

When producing a foreground blur effect, we target scenes with a much larger range of scene motion velocity, including slow and far away moving content. To produce a compelling effect, this requires \tacc{extending the capture for a duration up to several seconds, according to scene motion.} When the shutter button is pressed, we quickly analyze the scene motion statistics using the last \tacc{5} frames seen through the camera viewfinder and automatically determine \tacc{a subsequent capture duration that aims to satisfy our frame selection criteria. We use a lightweight variant of the motion tracking and image alignment described in the next few sections, that operates in under 50ms, to compute an estimate of scene velocity. With this estimate, under a constant velocity assumption, we trivially derive a capture duration that yields the desired blur trail length (see \sect{sec:frame_selection}). Given the extended capture duration of up to 7 seconds, we also derive a frame processing rate, to select an evenly distributed subset of up to 12 captured frames for processing, balancing the compute budget with a suitable temporal sampling rate. The captured frames selected for processing} are queued up for immediate concurrent processing by the following stages, thereby hiding some of the processing latency during capture.

\myvspace{-0.8em}

\subsection{Automatic Subject Detection}
\label{sec:subject_detection}

In the background blur case, we want the effect of a fixed subject with the rest of the world blurred behind them. Therefore, we automatically detect and track the main subject, and align the input frames to negate its motion. The subject is represented as a weight map, and is used in solving for the inverse subject motion alignment.

The main subject is first predicted using the proxy task of attention saliency. For this task, \tacc{we use a mobile-friendly 3-level U-Net with skip connections, with an encoder comprised of 15 BlazeBlock with 10 base channels~\cite{Bazarevsky19} and a corresponding decoder made of separable convolutions and bi-linear upsampling layers. It is distilled} from a larger model trained on the SALICON dataset~\cite{jiang2015salicon}. \tacc{To focus on the peak of saliency in our signal, we re-normalize predicted values to the interval $[0,1]$ and zero out values below a threshold (we empirically chose 0.43).}

The saliency signal tends to peak on the subject center, so we complement it with a face signal, which helps keep subject faces sharp, which is especially important in subjects with complex articulated motion. We compute the face signal by first predicting human, cat, and dog face regions, then feathering the resulting regions using a \texttt{smootherstep} falloff\tacc{~\cite{Ebert2003}}, and lastly masking it by a whole-subject segmentation similar to that of~\cite{Wadhwa18}.

We combine the saliency and face signals as follows, to produce the subject weight map with per pixel weight \tacc{\(w = s\ (1 + f)\)}, where $s \in [0, 1]$ is the saliency signal value and $f \in [0, 1]$ is the face signal value\tacc{, followed by a re-normalization to the interval $[0,1]$}. The face signal is also used in the compositing step to preserve face sharpness, as described in \sect{sec:compositing}.

\myvspace{-0.8em}

\subsection{Motion Tracking}
\label{sec:motion_tracking}

We use a feature tracking library based on~\cite{Grundmann2011} for extracting the motion tracks used in \tacc{subsequent image} alignment. Motion track statistics are also used to select frames, to determine when sufficient motion has been captured in the scene.

Subject tracking in background blur requires a high concentration of tracks on the subject for stable, high quality alignments. As a latency optimization, we use rejection sampling \tacc{over an image grid with cells of 5 x 5 pixels each, to generate feature tracks with density proportional to the subject weight map (\sect{sec:subject_detection}). We} only attempt to extract feature tracks in cells where a sampled uniform random variable $v \in [0,1]$ is smaller than the corresponding average track-weight at that grid location.

\subsection{Image Alignment}
\label{sec:image_alignment}

\begin{figure}[tb]
  \resizebox{120pt}{90pt}{%
  \begin{subfigure}[t]{\textwidth}
    \centering
    \includegraphics[width=\textwidth]{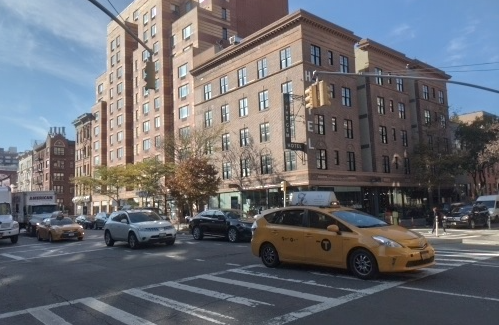}
  \end{subfigure}
  }%
  \resizebox{120pt}{90pt}{%
  \begin{subfigure}[t]{\textwidth}
    \centering
    \includegraphics[width=\textwidth]{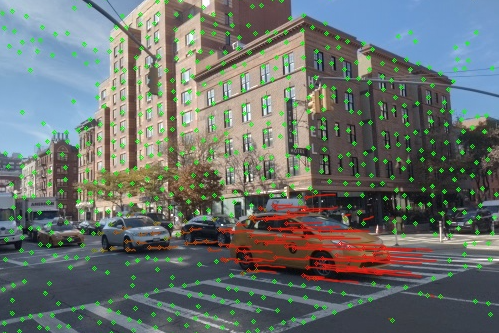}
  \end{subfigure}
  }
  \resizebox{120pt}{90pt}{%
  \begin{subfigure}[t]{\textwidth}
    \centering
    \includegraphics[width=\textwidth]{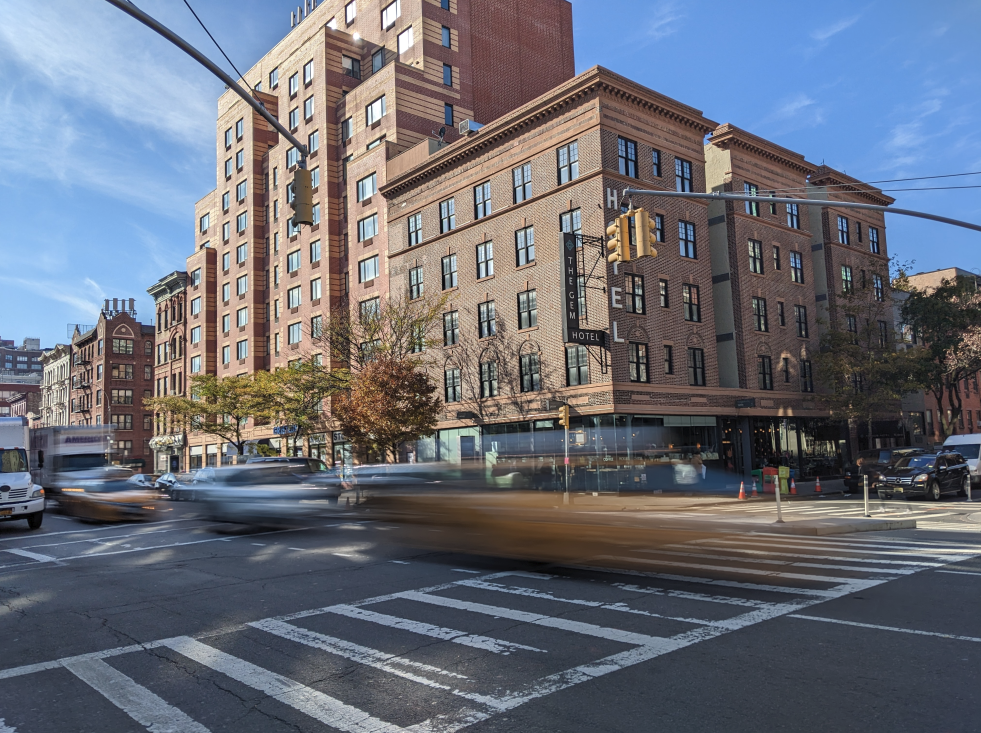}
  \end{subfigure}
  }%
  \resizebox{120pt}{90pt}{%
  \begin{subfigure}[t]{\textwidth}
    \centering
    \includegraphics[width=\textwidth]{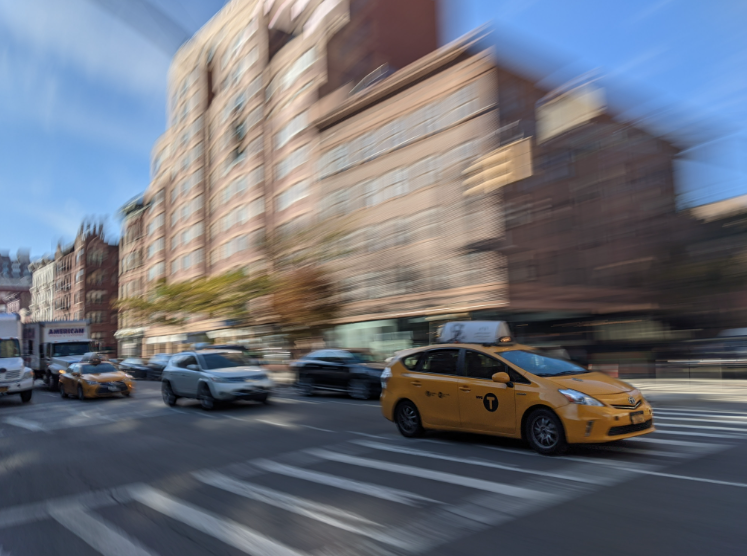}
  \end{subfigure}
  }%
  \myvspace{-0.25\baselineskip}
  \caption{City scene. Top: Traffic moving through a busy city intersection with foreground motion vectors (red) and background motion vectors (green).
  Bottom: Our foreground blur (left) and background blur (right) results.}%
  \myvspace{-1em}
  \label{fig:alignment_taxicab}%
\end{figure}

Given the feature track correspondences from \sect{sec:motion_tracking}, we first estimate global transforms to align all the frames to our reference frame. This cancels out overall camera motion, including both hand-shake and sweeping motions used to track subjects. The remaining image alignment stages are specific to the desired motion-blur effect: foreground or background blur. For the purpose of illustration, we pick an example scene that could be rendered as either: a taxicab passing through a busy city intersection as shown in \fig{fig:alignment_taxicab}.

\subsubsection{Foreground blur}
\label{subsec:alignment_fg_blur}
To keep the background as sharp as possible, we must account for spatial effects such as parallax, rolling shutter, and lens distortion. After applying global transforms, we \tacc{compute a residual vector as the position difference between a transformed tracked feature and its corresponding position in the base frame. We then use the} residuals to estimate local refinement transforms on a grid of vertices across the image. The resulting spatially varying warp cancels motion in the background while retaining motion in the foreground, producing sharp backgrounds as in \fig{fig:alignment_taxicab} (bottom-left).

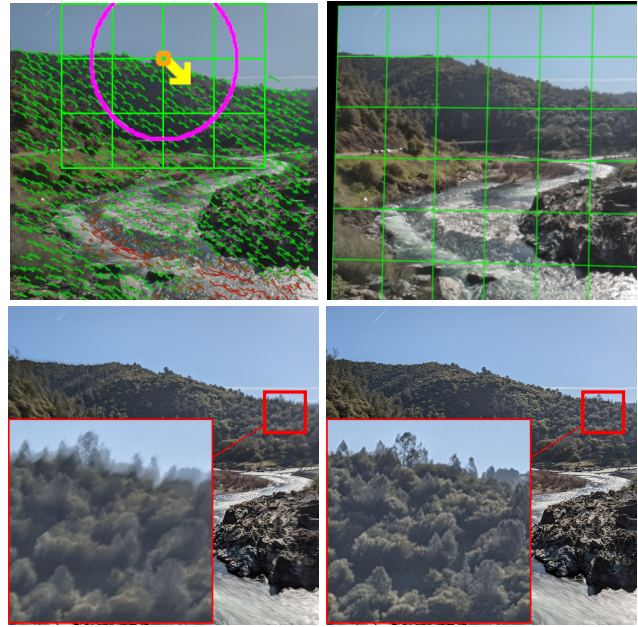
\begin{figure}[tb]
  \centering
  \hspace{-0.3\baselineskip}
  \resizebox{0.48\columnwidth}{!}{%
  \begin{subfigure}[t]{\textwidth}
    \centering
    \includegraphics[width=\textwidth,trim={0 5cm 0 0},clip]{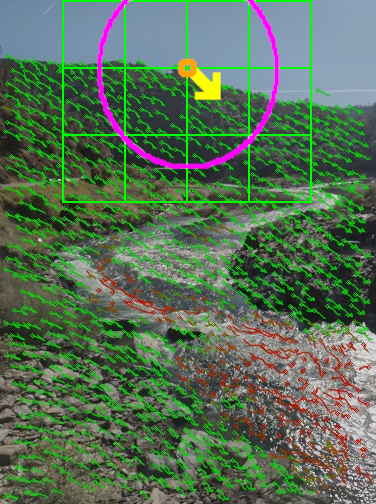}
  \end{subfigure}
  }%
  \hspace{0.07\baselineskip}
  \resizebox{0.485\columnwidth}{!}{%
  \begin{subfigure}[t]{\columnwidth}
    \centering
    \includegraphics[width=\textwidth,trim={0 5cm 0 0},clip]{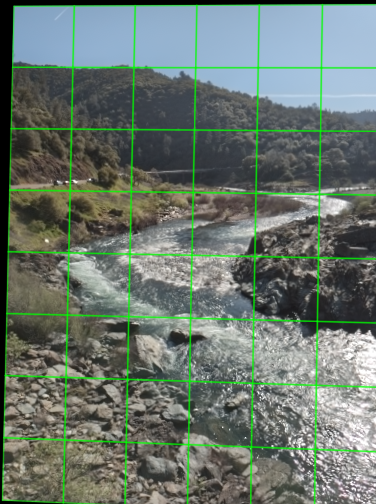}
  \end{subfigure}
  }%
  \myvspace{-0.1\baselineskip}
  \resizebox{0.495\columnwidth}{!}{%
  \begin{subfigure}[t]{\textwidth}
    \centering
        \resizebox{\columnwidth}{!}{%
        \begin{tikzpicture}[node distance=0,outer sep=0,spy using outlines={every spy on node/.append style={smallwindow}}]
          \coordinate (se) at (-1.99, 1.5);
          \coordinate (nw) at (3.8, 4);
          \clip (se) rectangle + (nw);
          \node[anchor=south](FigA) at (0,0) {\includegraphics[height=5.3cm]{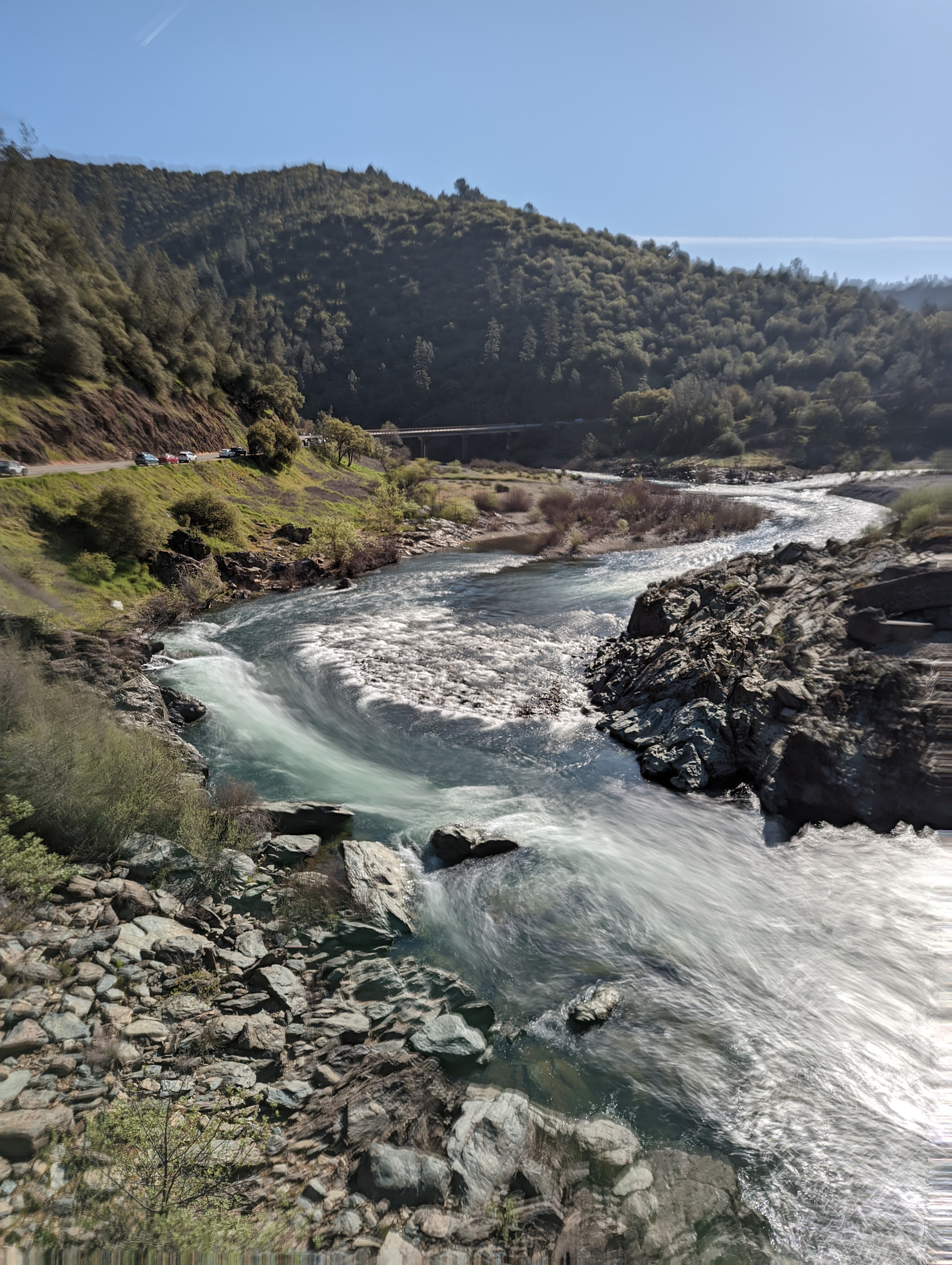}};
          \spy [closeup,magnification=5] on ($(FigA)+( 1.4, 1.35)$) 
            in node[largewindow,anchor=south]       at ($(FigA.south)+(-0.74, 1.52)$);
        \end{tikzpicture}
        }%
    \end{subfigure}
    }%
    \resizebox{0.495\columnwidth}{!}{%
    \begin{subfigure}[t]{\textwidth}
    \centering
        \resizebox{\columnwidth}{!}{%
        \begin{tikzpicture}[node distance=0,outer sep=0,spy using outlines={every spy on node/.append style={smallwindow}}]
          \coordinate (se) at (-1.99, 1.5);
          \coordinate (nw) at (3.8, 4);
          \clip (se) rectangle + (nw);
          \node[anchor=south](FigB) at (0,0) {\includegraphics[height=5.3cm]{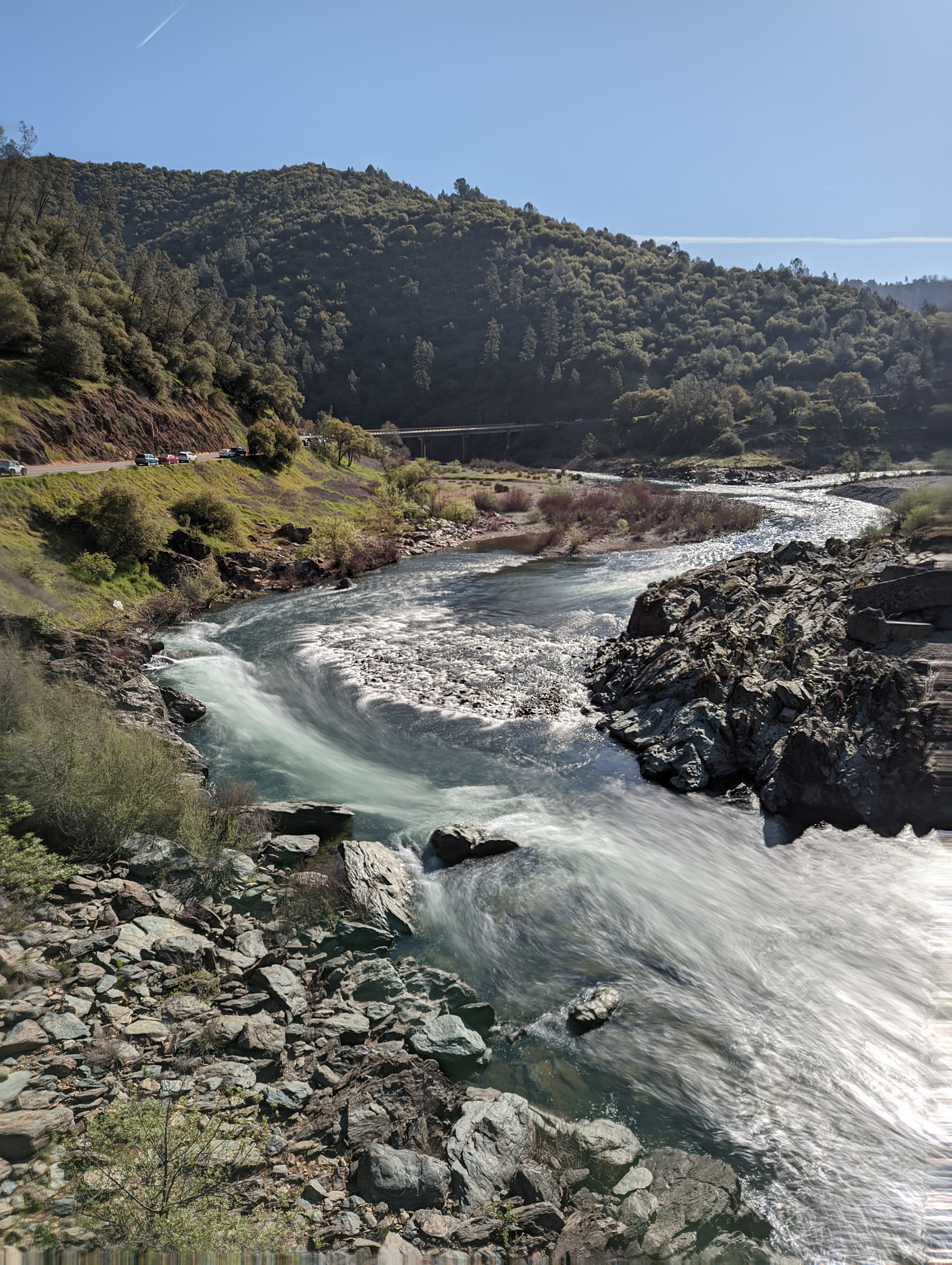}};
          \spy [closeup,magnification=5] on ($(FigA)+( 1.4, 1.35)$) 
            in node[largewindow,anchor=south]       at ($(FigA.south)+(-0.74, 1.52)$);
        \end{tikzpicture}
        }%
    \end{subfigure}
    }%
  \myvspace{-0.25\baselineskip}
  \caption{Spatially varying warp. Top-left: The background flow vectors (green) inside one of the mesh vertex (orange) support regions (magenta) contribute to its local displacement (yellow). Top-right: The resulting mesh and spatially varying warp that aligns the background to that of the reference frame. Foreground blur results using only a single global homography transform (bottom-left) and using our spatially varying mesh warp (bottom-right). Insets are displayed at 5x magnification.}%
  \label{fig:alignment_spatially_varying_warp}%
\end{figure}

In~\cite{Zaragoza2013apap}, the authors weight points by distance from each grid vertex to produce a spatially varying as-projective-as-possible warp. Our approach to placing a grid and estimating local transforms is similar, but we weight our points uniformly and use a hard cut-off for point inclusion during local similarity transform estimation for better performance. The estimation is controlled by the support radius of each vertex (shown as magenta circle in \fig{fig:alignment_spatially_varying_warp}), i.e. the maximum distance from a vertex that a feature point needs to be for inclusion in the local refinement estimation. We found that setting this radius to 1.5 times the cell size of the mesh grid \tacc{and using a grid size of 8 x 6 cells,} was large enough for the local refinement transforms to vary smoothly across the entire field-of-view, yet small enough that disparate scene objects from different parts of the scene do not affect each other. The estimated transforms are applied to each vertex to then generate a spatially varying mesh that aligns the background of any frame to that of the reference frame. To optimize latency, the application of this mesh is folded into the downstream rendering stage by passing a texture of local 2-d displacement vectors to the GPU.

\subsubsection{Background blur}
\label{subsec:alignment_bg_blur}

In this case, as shown in \fig{fig:alignment_taxicab} (bottom-right), we want the foreground to be as sharp as possible. We use the subject mask from \sect{sec:subject_detection} to select the subset of feature tracks that correspond to the foreground subject. With this as a starting point, we further use spectral clustering to select the most salient motion cluster to help discern the right motion segment to track and to remove outliers~\cite{porikli2004learning}. This is especially useful for articulated subjects, such as a running person whose limbs move differently from their torso.

The goal of our objective function is to balance two goals: (1) \tacc{s}ubject sharpness: minimize the overall reprojection error of the salient point correspondences, and\tacc{ (2) t}emporal smoothness: keep the transformed background motion vectors as parallel as possible to those from the previous time step, as shown in \fig{fig:alignment_rotations} and \fig{fig:alignment_ablations}.

Given a pair of frames $i$ and $j$ with point correspondences $\mathbf{x}_i$ and $\mathbf{x}_j$, we define a similarity transform that scales uniformly ($s_{j,i} \in \mathbb{R}^{1}$), rotates ($\mathbf{R_{j,i}} \in SO(2)$) and translates ($\mathbf{t_{j,i}}  \in \mathbb{R}^{2}$) 2-dimensional points from frame $i$ to frame $j$ as follows.
\begingroup
\setlength{\abovedisplayskip}{5pt}
\setlength{\belowdisplayskip}{3pt}
\begin{equation}
\begin{split}
    \mathbf{\hat{x}_j} &= s_{j,i}\mathbf{R_{j,i}}\mathbf{x_{i}} + \mathbf{t_{j,i}} \label{eqn:alignment_transformed_points}
\end{split}
\end{equation}
\endgroup
For simplicity, we omit the from-and-to indices from the transform parameters $s$, $\mathbf{R}$ and $\mathbf{t}$ and define our objective function as follows.
\begingroup
\setlength{\abovedisplayskip}{6pt}
\setlength{\belowdisplayskip}{4.5pt}
\begin{equation}
\begin{split}
    \minimize_{s, \mathbf{R}, \mathbf{t}} \quad \lambda_{f} E_{f}(s, \mathbf{R}, \mathbf{t}) + \lambda_{b} E_{b}(s, \mathbf{R}, \mathbf{t}) \label{eqn:alignment_objective_fn}
\end{split}
\end{equation}
\endgroup
The scalars $\lambda_f$ and $\lambda_b$ are the relative weights of the objective function terms. The subject sharpness term $E_f$ is defined on the foreground points $\mathbf{x} \in \mathcal{X}_{f}$ as the L-2 norm of the reprojection error of transformed point correspondences (using Eq.~\ref{eqn:alignment_transformed_points}) as follows.
\begingroup
\setlength{\abovedisplayskip}{6pt}
\setlength{\belowdisplayskip}{4.5pt}
\begin{equation}
\begin{split}
    E_{f}(s, \mathbf{R}, \mathbf{t}) & = \sum_{\mathbf{x} \in \mathcal{X}_{f}} \norm{\mathbf{x_{j}} - s\mathbf{R}\mathbf{x_{i}} - \mathbf{t}}_2
\end{split}\label{eqn:alignment_term_ef}
\end{equation}
\endgroup
The background term $E_{b}$ is used as a temporal smoothness prior to penalize background flow vectors that are not parallel to their counterparts from the previous frame pair. Given three distinct frame indices $i$, $j$ and $k$, this is defined using vector dot product as a measure of parallelism as follows:
\begingroup
\setlength{\abovedisplayskip}{4pt}
\setlength{\belowdisplayskip}{4pt}
\begin{equation}
\begin{split}
    E_{b}(s_{i,k}, \mathbf{R}_{i,k}, \mathbf{t}_{i,k}) & = \sum_{\mathbf{x} \in \mathcal{X}_{b}} \smooth_{L_1} \left( \frac{| \mathbf{v}_{i,j} \cdot \mathbf{v}_{j, k} |}{\norm{\mathbf{v}_{i,j}} \norm{\mathbf{v}_{j, k}}} \right)
\end{split}\label{eqn:alignment_term_eb}
\end{equation}
\endgroup
where $\mathbf{v}_{p,q} = \mathbf{\hat{x}}_q - \mathbf{x}_q$ is the directional displacement vector between a point in the reference frame $q$ and its transformed point correspondence from frame $p$. The smooth-$L_1$ loss function from~\cite{Huber1964} is used to prevent vanishing gradients during the optimization.
\begin{equation}
\smooth_{L_1}(x) =
    \begin{cases}
        0.5 x^2, & \text{if}\ |x| < 1 \\
        |x| - 0.5, & \text{otherwise}
    \end{cases}    
\end{equation}
Without loss of generality, we chose $i = 0$ as the fixed reference frame index, and set $k = j + 1$ in Eq.~\ref{eqn:alignment_term_eb} to estimate the transforms incrementally for each new frame pair $(j, j+1)$ in a single forward pass through the frame sequence. We solved this non-linear optimization problem using the Ceres library~\cite{Agarwal_Ceres_Solver_2022}. \tacc{We use values $1$ and $10$ for $\lambda_f$ and $\lambda_b$ respectively, to balance the effect of both error terms.}

The temporal regularization term $E_b$ is not defined for the first frame pair, which we handle as a special case using a constraint on the scale and rotation parameters, $s$ and $\mathbf{R}$, from Eq.~\ref{eqn:alignment_objective_fn}. Our key observation comes from subjects of background blur shots that undergo rotational motion in the image plane. Simply inverting this rotation produces undesirable multiple sharp regions in the result, as shown in \fig{fig:alignment_rotations}. \tacc{In traditional panning photography, it is uncommon to attempt the rotation blur effect and exceedingly difficult to achieve subject sharpness in this manner (i.e. rotating the camera around an axis centered away from the subject). Instead, it is typically done panning the camera, tracking the overall trajectory of the subject, and our method aims for these outcomes.}

To estimate the initial scale and rotation, we use the integrated estimation technique from~\cite{zinsser2005point}. We then constrain the estimated rotation, $\mathbf{R}$ in Eq.~\ref{eqn:alignment_objective_fn}, to prevent any additional sharp regions from detracting away from the sharp subject. We empirically found that constraining the roll angle to 25\% of its estimated value helps make the blur field more linear, as shown in \fig{fig:alignment_rotations}, with only the subject kept sharp as desired. More examples of image alignment for background-blur scenes are provided in \sect{sec:alignment_comparison}.

\begingroup
\begin{figure}[htb]
    \centering
    \begin{subfigure}[t]{0.49\columnwidth}
        \includegraphics[width=\textwidth]{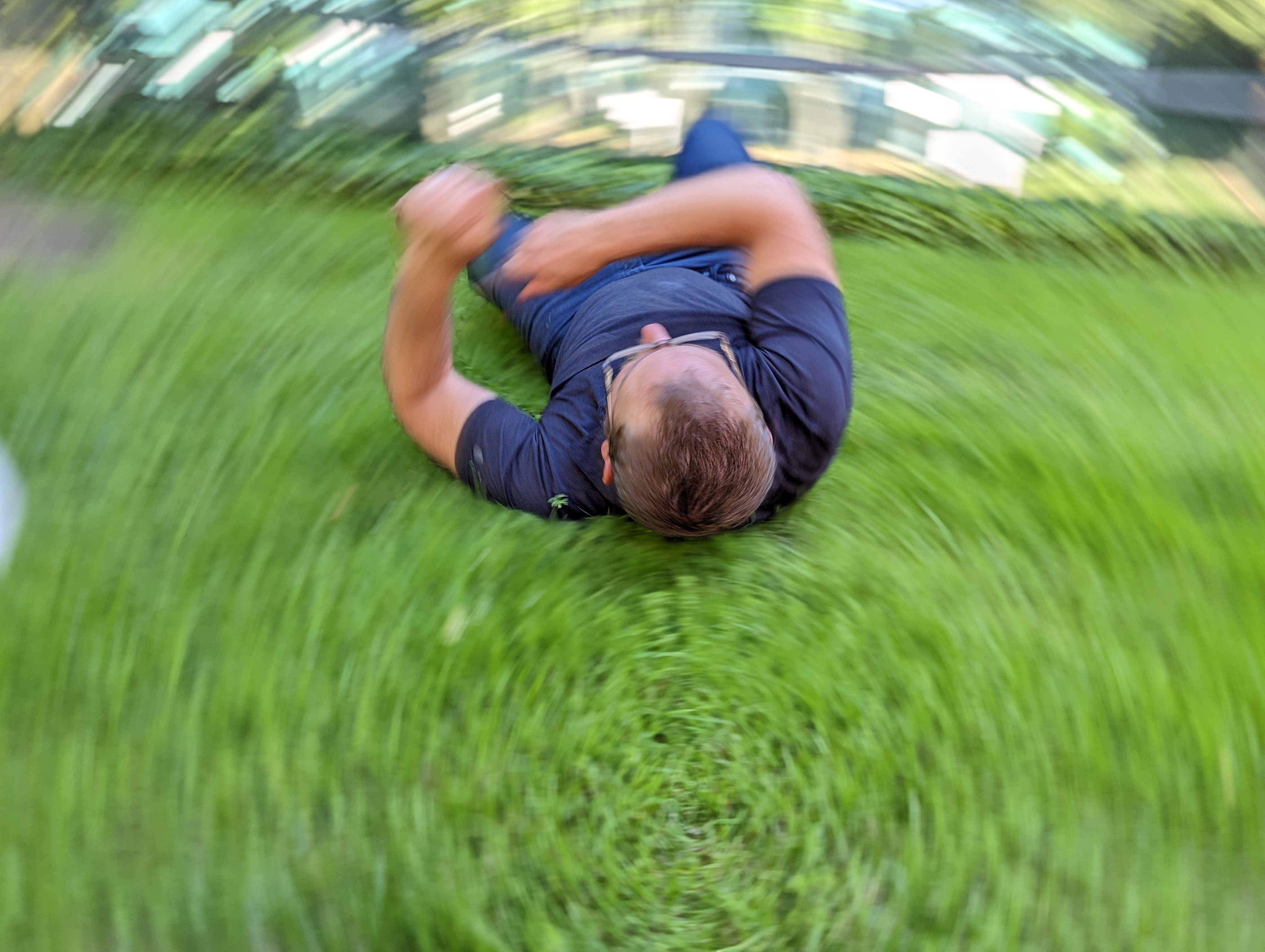}
        \myvspace{-1.3em}
        \subcaption{Unconstrained rotation}
        \label{subfig:unconstrained_rotation}
    \end{subfigure}
    \hfill
    \begin{subfigure}[t]{0.49\columnwidth}
        \includegraphics[width=\textwidth]{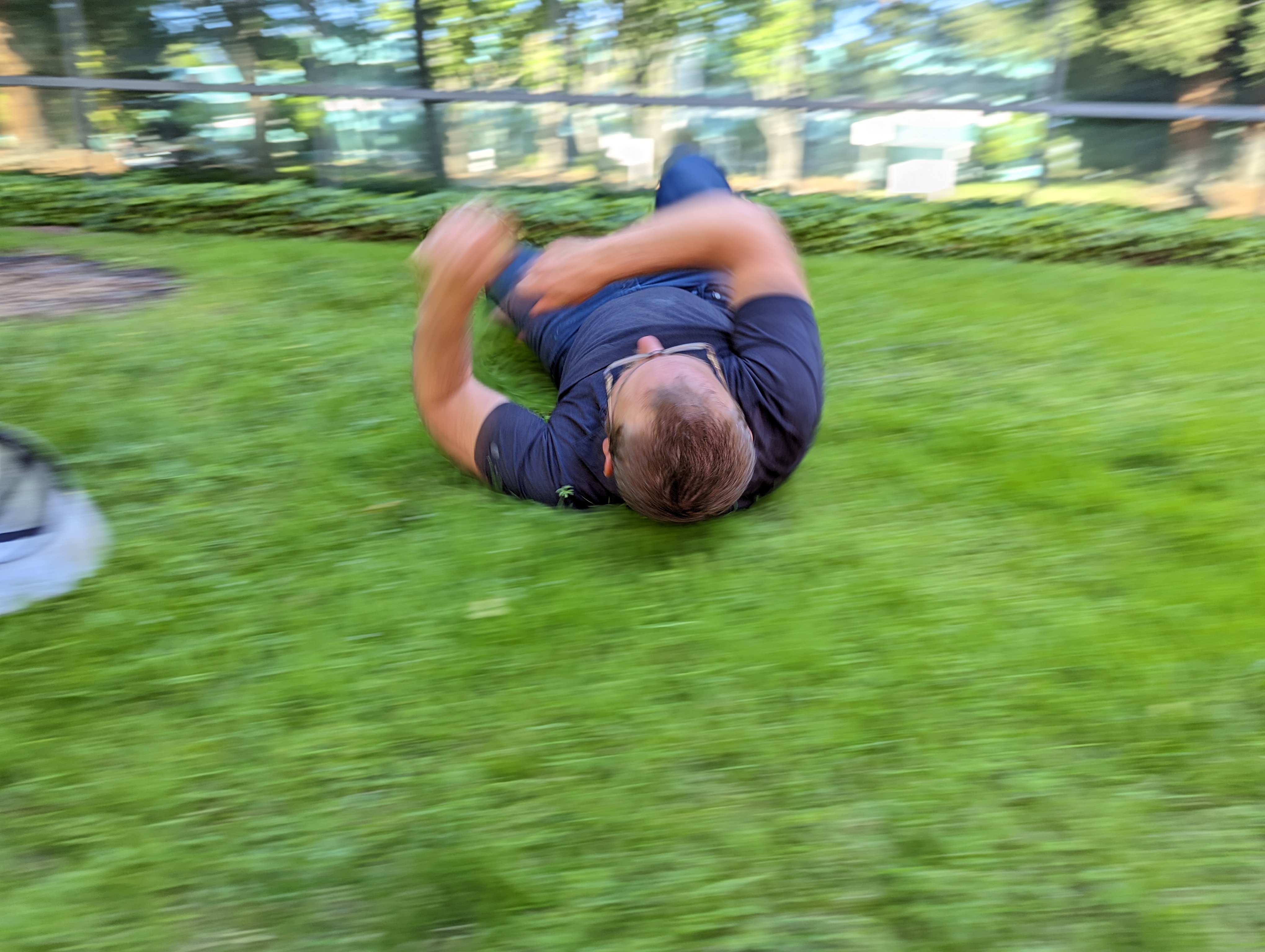}
        \myvspace{-1.3em}
        \subcaption{Constrained rotation}
        \label{subfig:constrained_rotation}
    \end{subfigure}
    \myvspace{-1em}
    \caption{Undesirable rotations. Fully inverting the subject's rotation (\subref{subfig:unconstrained_rotation}) gives us an undesirable result with an additional sharp region below the subject. \tacc{Even though the rotating blur can be a fun effect, the sharpness region at the center of rotation attracts the viewer’s attention away from the main subject and degrades the subject separation from the background, both of which goes against photography composition rules.} We alleviate this by constraining the estimated rotation  (\subref{subfig:constrained_rotation}).}%
    \label{fig:alignment_rotations}%
\end{figure}
\endgroup

\tacc {

\subsection{Frame Selection}
\label{sec:frame_selection}

Our system uses a frame selection mechanism that computes an estimate of motion-blur trails' length, to decide when the incremental frame processing outer-loop should stop (see \sect{sec:system_overview}). First, we use the transformations computed by the alignment solver to transform the motion feature tracks to the reference space of the base frame, where they align spatially with the corresponding tracked features' motion-blur trails in the output image. The length of each aligned track can then be computed, and we use a high percentile of the track length distribution as an estimate of overall blur trail length. This estimate is finally compared to a constant target setting, to decide whether the frame selection criteria is satisfied.

We measure the track length in percentage of image diagonal, a metric that is largely insensitive to image resolution or aspect-ratio. In the case of foreground blur, our criteria is for the 98\textsuperscript{th} percentile to reach a target of 30\%, producing relatively long and smooth blur trails for the fastest moving object. In the background blur case, we use the 80\textsuperscript{th} percentile and a target of 2.8\%, producing short blur trails for a larger area of the background, aiming to preserve subject sharpness and avoid losing the context of the surrounding scene. These settings were derived empirically, iterating over large collections of input bursts.

}

\subsection{Motion Prediction}
\label{sec:motion_prediction}
Once the input low-resolution images are aligned, we feed them through a motion-blur kernel-prediction neural network, one input frame pair at a time, predicting a pair of line and weight kernel maps at each iteration. The low-resolution kernel maps are used to synthesize motion-blur segments at half resolution, spanning the corresponding input frames, as described in \sect{sec:rendering}.

The motion prediction model is responsible for predicting the parameters of two spatial integrals along line segments, which approximate the temporal integral defining the averaging of colors seen through each motion-blurred output pixel, during the corresponding time interval. We use a model based on~\cite{Brooks19}, with further modifications that improve the trade-off between performance and image quality, allowing us to fit within a reasonable memory and compute budget on mobile devices.

Their mathematical formulation predicts weight maps \(W_{i}\) per input frame \(i\) in a given image pair \(k\), with \(N = 17\) channels, which are used to weigh each corresponding texture sample along the predicted line segments. We simplify this model by predicting only a single channel, used to weigh the result of the integral from each input frame. An example gray-scale map can be seen in \fig{fig:system_overview}, showing that the network predicts approximately equal weights everywhere across input images, except in areas of dis-occlusion where the weights favor the result from one of the two inputs. This simplification significantly reduces system complexity and memory use, and allows for more of the network capacity to be devoted to predicting the line segments.

\begin{figure}
  \centering
  \begin{subfigure}[t]{0.491\columnwidth}
    \includegraphics[width=\columnwidth]{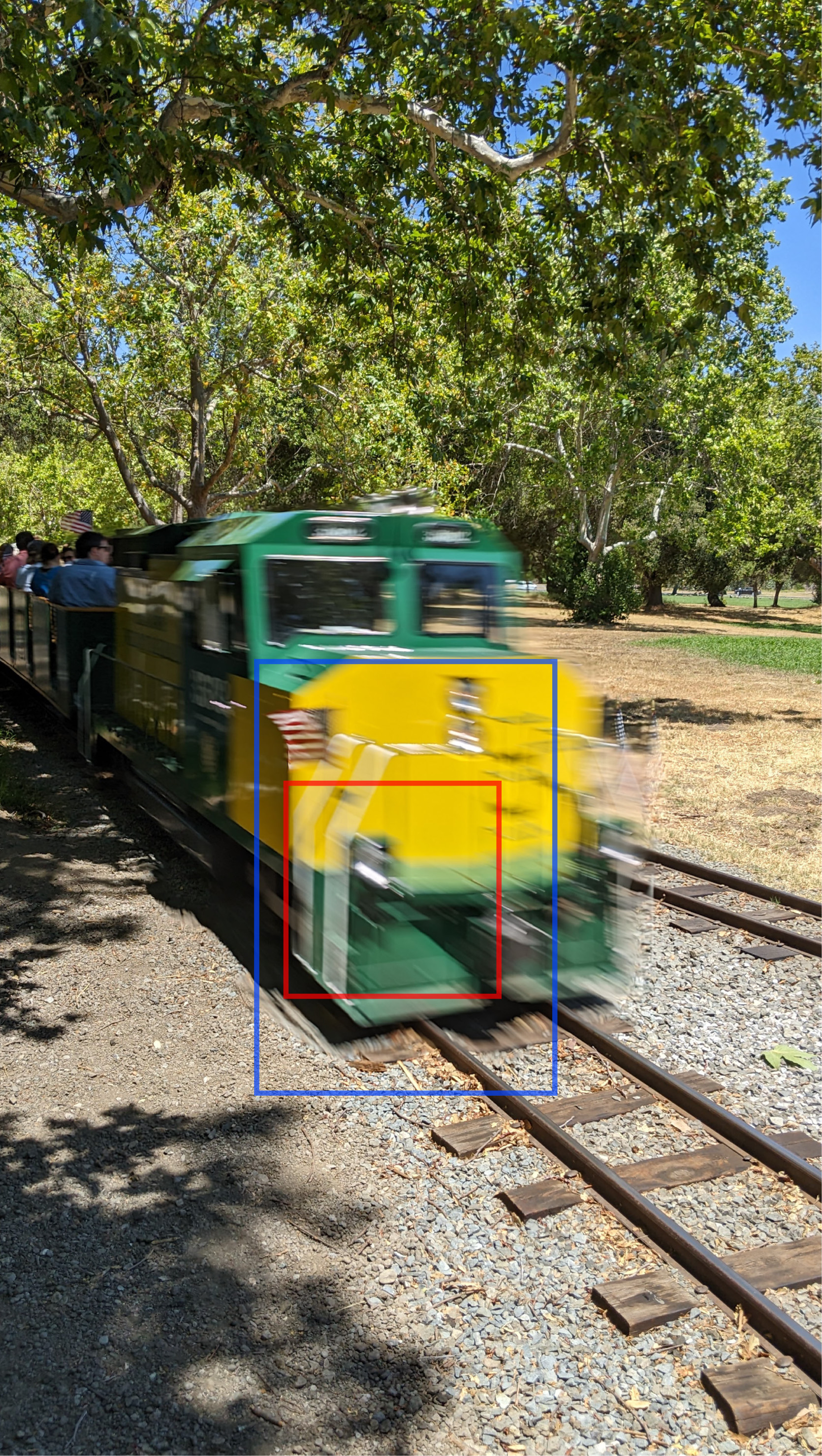}
    \myvspace{-1.25em}
    \subcaption[]{Blur from single image pair}
    \label{subfig:blur_from_single_pair}
  \end{subfigure}
  \hfill
  \begin{subfigure}[t]{0.48\columnwidth}
    \vspace{-23.5em}
    \begin{subfigure}[t]{\columnwidth}
      \includegraphics[width=\columnwidth]{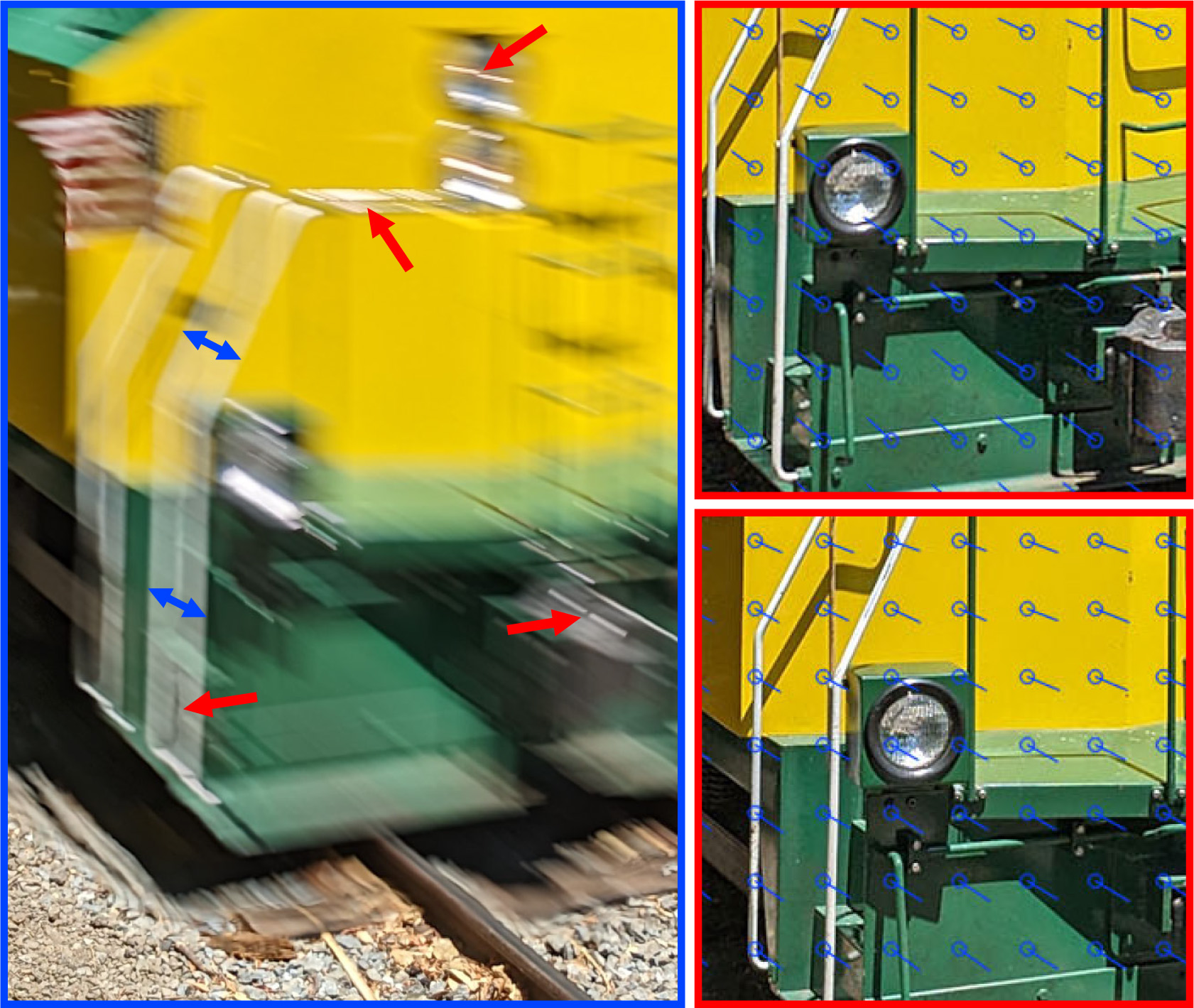}
      \myvspace{-1.6em}
      \subcaption[]{Without weight ramp (Ours-abl.)}
      \myvspace{0.1em}
      \label{subfig:without_weight_ramp}
    \end{subfigure}
    \begin{subfigure}[t]{\columnwidth}
      \includegraphics[width=\columnwidth]{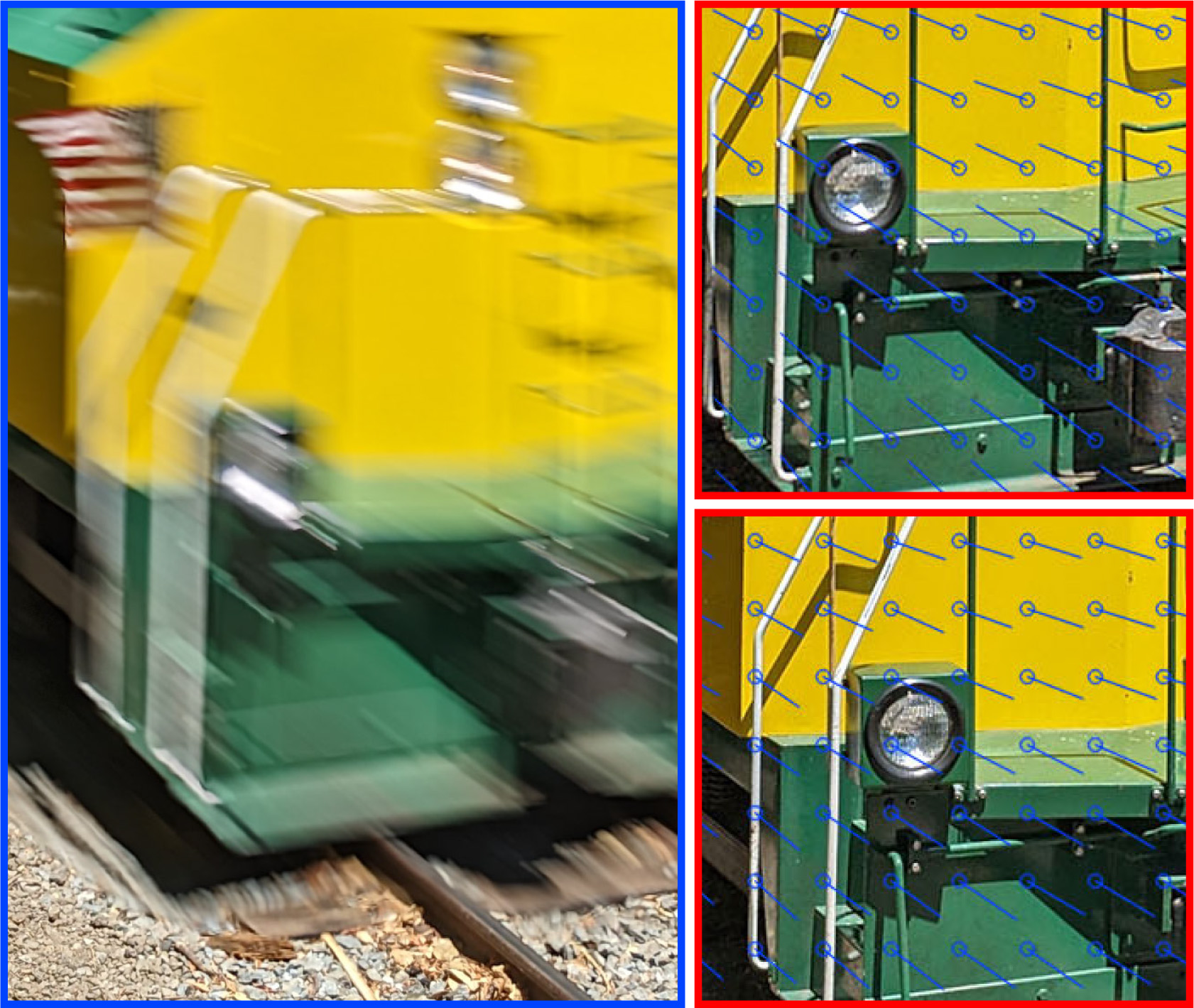}
      \myvspace{-1.25em}
      \subcaption[]{With weight ramp (Ours)}
      \label{subfig:with_weight_ramp}
    \end{subfigure}
  \end{subfigure}
  \myvspace{-0.75em}
  \caption{Motion prediction model ramp function ablation. (\subref{subfig:blur_from_single_pair}) Rendering of a motion-blurred moving train synthesized from a single input image pair. Both a motion-blurred closeup of the front of the train and corresponding input image pair overlaid with a vector field visualization representing the predicted line segments, is shown in (\subref{subfig:without_weight_ramp}) and (\subref{subfig:with_weight_ramp}) using the models "Ours-abl." and "Ours", i.e. without and with the ramp function \(w_{n}\), respectively. In image (\subref{subfig:without_weight_ramp})-left, the blue arrows indicate the full span of motion blur trails and the red arrows showcase the gap discontinuities in the middle of blur trails that are most noticeable.}
  \label{fig:motion_prediction_middle_gap}
\end{figure}

In addition, we eliminate artifacts due to the predicted line segments' endpoint error~\cite{Zhang16}, causing them to meet imperfectly at the end of the spanned time interval, and resulting in very noticeable artifacts in the middle of blur trails, as illustrated in~\fig{fig:motion_prediction_middle_gap}. To avoid this issue, we scale the input image texture samples further by a normalized decreasing linear ramp function \(w_{n}\), that favors samples close to the output pixel and gradually down-weighs samples further away along each predicted line segment. The intensity of the output pixel \((x,y)\) for the input frame pair \(k\) is:
\begin{equation}
  I_{k}(x,y) = \sum_{i\in\{k, k+1\}} \frac{W_{i}(x, y)}{\sum_{n=0}^{N-1}w_{n}}\ \sum_{n=0}^{N-1} w_{n}\ I_{i}(x_{in}, y_{in})
\end{equation}
with \(w_{n} = 1 - n / N\), and with sampled positions:
\[x_{in} = x + (\frac{n}{N - 1})\ \Delta_{i}^{x}(x, y)   \;\;\;\text{and}\;\;\;  
  y_{in} = y + (\frac{n}{N - 1})\ \Delta_{i}^{y}(x, y)\]
where \(\Delta_{i}\) are the predicted line segments.

We also modify the network architecture as follows. First, we replace the leaky ReLU convolution activations throughout, with a parameterized ReLU~\cite{He15}, where the slope coefficient is learned. Next, to avoid common checkerboard artifacts~\cite{Odena16}, we replace the 2x resampling layers to use average pooling for downsampling, and bi-linear upsampling followed by a 2x2 convolution. This results in a model labeled "Ours-large" analyzed in \sect{sec:results}. Furthermore, to improve the balance between the number of floating operations, number of parameters and receptive field, we further reduce the U-Net model topology to only 3 levels, where each level is using a 1x1 convolution, followed by a ResNet block~\cite{He16} with four 3x3 convolution layers. This results in a model labeled "Ours" with significantly fewer learned parameters.

As shown in \fig{fig:motion_prediction_middle_gap}, the ramp function \(w_{n}\) brings a significant benefit to our learned single weight model, as it causes the predicted line segments to span spatially in each input image, the equivalent of the full time interval being integrated. When our model is trained with this term ablated, resulting in the model "Ours-abl.", the network predicts line segments that span approximately half of the time interval on each side, causing the noticeable discontinuity in the middle of blur trails. More examples can be found in the model comparison analysis provided in \sect{sec:results}.

\subsection{Rendering}
\label{sec:rendering}
The line and weight kernel maps output by the motion prediction network are used by a renderer that synthesizes the motion-blurred image. The renderer is implemented in an OpenCL kernel, which runs very efficiently on the mobile device's GPU, leveraging the hardware texturing units while adaptively sampling texture lookups in the half resolution input images (the number of texture samples \(N\) is adjusted proportionally to the length of the predicted line vectors). Motion prediction and rendering iterations can be performed one input frame-pair at a time, producing piecewise-linear motion-blur trails. Kernel maps are up-sampled from low to half-resolution by using bi-linear texture lookups.

\subsubsection{Spline interpolation}
\label{subsec:spline_interpolation}
Piecewise-linear motion interpolation may introduce jagged visual artifacts in motion trails. To interpolate the motion more smoothly, we interpolate the inferred instantaneous flow \({\Delta}_i \) between frames using cubic Hermite splines.

The instantaneous flow \({\delta}_i \) at each pixel is inferred by constructing a vector  \(H({\Delta}_i^+, {\Delta}_i^-) \) parallel to \(({\Delta}_i^+ + {\Delta}_i^-) \), with magnitude equal to the harmonic mean of \(|{\Delta}_i^+| \) and \(|{\Delta}_i^-| \). Superscripts \(+\) and \(-\) refer to time directions. If \({\Delta}_i^+ \) and \({\Delta}_i^- \) deviate by an angle \(\theta \) from a straight-line path, the \tacc{vector} is further scaled by a factor of \((\theta / \sin \theta) \) for smaller angular deviations (< 90°), tapering this adjustment back towards zero for larger deviations (where the path doubles back acutely) to avoid singularities. These correctional factors reduce overshoot, and keep the parametric spline speed more stable for regions of moderate curvature.
\begin{equation}
{\delta}_i = H({\Delta}_i^+, {\Delta}_i^-)\ (\theta / \sin \theta) \times
  \left.
  \begin{cases}
    1, &  \theta \leq \pi/2 \\
    1 - ({{2\theta}/{\pi} - 1})^4, &  \theta > \pi / 2
  \end{cases}
  \right.
\end{equation}

For the accumulated blur of \(I_k \) on the interval [k .. k+1] for output position (x, y), we solve for a parametric 2D cubic spline path \({\rho} \)(x, y, t) satisfying four constraints:

\begin{itemize}
  \item \({\rho} \)(x, y, 0) = (x, y)
  \item \({\rho} \)(x, y, 1) = (x, y) + \({\Delta}_i^+ (x, y) \)
  \item \({\rho}'(x, y, 0) = {\delta}_i ({\rho} (x, y, 0)) \)
  \item \({\rho}'(x, y, 1) = {\delta}_{i+1} ({\rho} (x, y, 1)) \)
\end{itemize}

\begin{figure}
  \begin{minipage}[c]{0.48\linewidth}
    \centering
    \includegraphics[width=0.8\columnwidth]{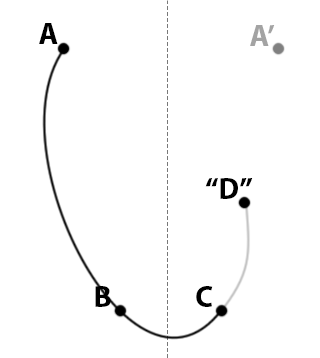}
    \caption{Our spline extrapolation strategy. See \sect{subsec:spline_interpolation}.}
    \label{fig:spline_extrapolation}
  \end{minipage}\hfill
  \begin{minipage}[c]{0.47\linewidth}
    \centering
    \includegraphics[width=\columnwidth]{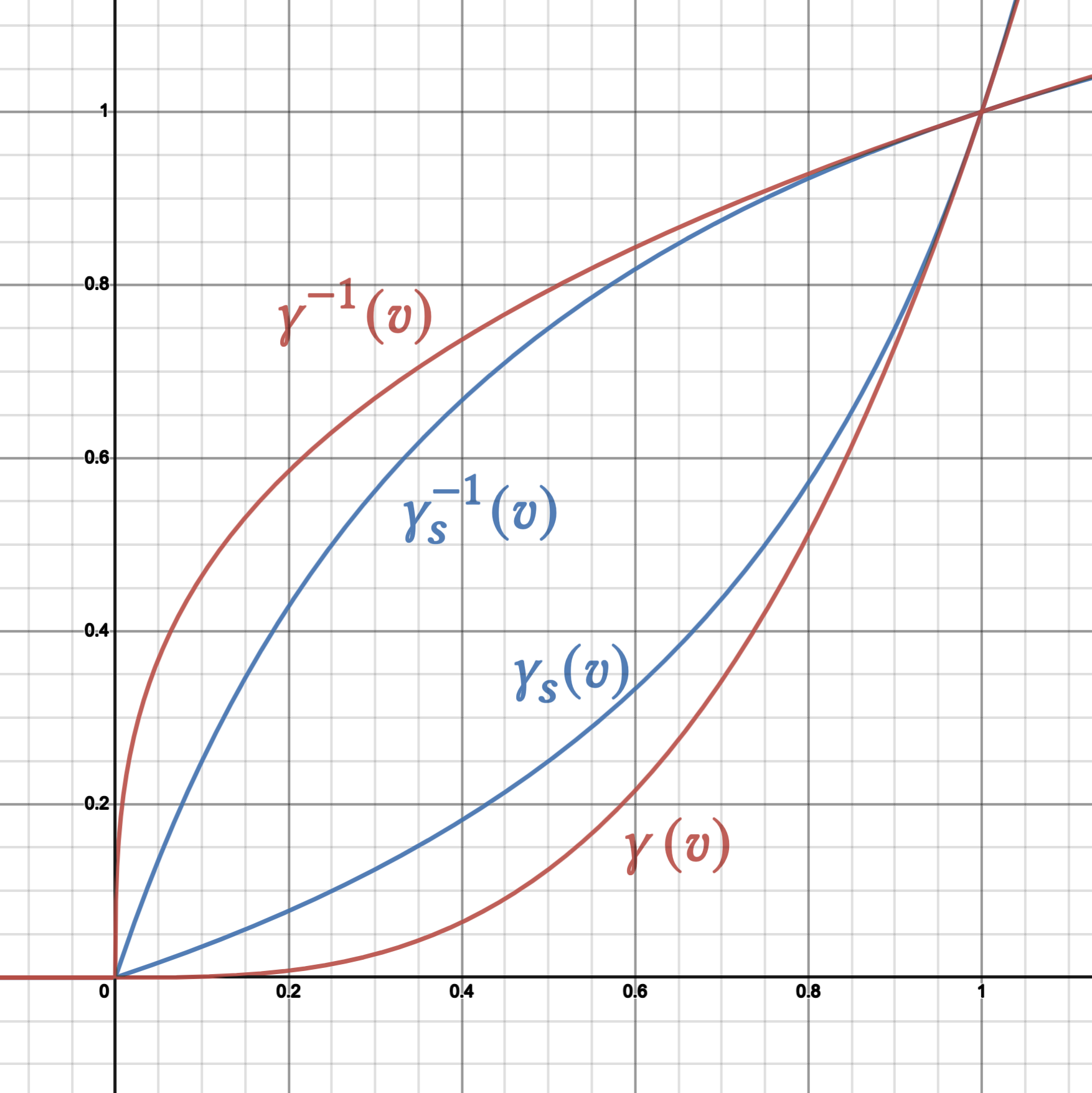}
    \caption{Comparison of traditional gamma \(\gamma \) vs. our soft gamma \({\gamma}_s \). See \sect{subsec:soft_gamma}.}
    \label{fig:soft_gamma}
  \end{minipage}
  \centering
\end{figure}

We then accumulate the blur along this path by sampling uniformly in parameter space, normalizing the weight of each sample to compensate for the non-uniform spatial sampling in image space in order to ensure spatially uniform brightness along motion trails. At the burst endpoints we extrapolate the flow beyond the first and last frames by attempting to preserve the curvature of the flow through those endpoints. As shown in \fig{fig:spline_extrapolation}: if 'C' represents the final frame in a burst, a motion trail position at the "next" frame D is extrapolated by reflecting A in the line bisecting BC (constructing A'), then clamping the magnitude of CA' to |BC| to form CD. The flow at C is then inferred from points \{B,C,D\}.

\subsubsection{Frame accumulation}\label{subsec:frame_accumulation}
In practice, the blur is accumulated in several passes: two passes per frame pair, weighted to fall off linearly between one frame and the next. For an output pixel at position p at frame \(I_i \), the blur between frame \(I_i \) and \(I_{i+1} \) is accumulated by using the aforementioned flow splines to determine the projected position p' in frame \(I_i \) at relative time t. For K frame pairs in the burst, 2K such passes (K forward, K backward) are computed and summed to produce the final blur result. For each temporal direction:
\begin{equation}
  I(x,y) = \sum_{i=0}^{K - 1} \sum_{n=0}^{N - 1} {I_i}({\rho}_{i}(x, y, t_n))\ |{\rho}_{i}'(x, y, t_n)|\ w_n
\end{equation}

\subsubsection{Soft Gamma Colorspace}\label{subsec:soft_gamma}
Very bright highlights (e.g. car headlights) tend to saturate the camera sensor, resulting in their blurred motion trails becoming unrealistically dim even when processed in linear colorspace. \tacc{The clipping is due to the finite range of the input sensor, and the brightness loss becomes noticeable when the clipped input highlight energy is distributed (i.e. synthetically motion-blurred) over many output pixels.}

To work around this limitation, we process the blur in an intentionally non-linear colorspace, using an invertible gamma-like "soft gamma" function \({\gamma}_s \), shown in \fig{fig:soft_gamma}, on the interval \([0..1] \). This adjusts the brightness curve in the opposite direction from a linear-to-sRGB color transformation, emphasizing highlights without crushing shadows, allowing the nonlinear frames to be stored with usable fidelity in 16-bit buffers. The function is applied to the warped downsampled 2x buffers on creation, using a value of 3.0 for \(k \), and is later inverted (by reapplying with \(k = 1.0 / 3.0 \)) after accumulating the blur for all frames. (See ablation in \sect{sec:results}).
\begin{equation}
\begin{split}
  {\gamma}_s (v) = \frac{v}{v + (1 - v)\ k} \hspace{5pt} \approx \hspace{5pt} v^k
  \label{eqn:rendering_soft_gamma}
\end{split}
\end{equation}
This is homologous to the Bias family of functions in \cite{10.5555/180895.180931}, but our reparameterization in Eq.~\ref{eqn:rendering_soft_gamma} makes clearer the connection to the corresponding gamma curve with exponent \(k \). The idea of processing the blur in the modified colorspace was inspired by the Ordinal Transform technique in \cite{10.1145/1141911.1141918}. \tacc{Our goal is similar to the clipped highlight recovery technique in~\cite{Lancelle19}, which in comparison uses a more abrupt discontinuous highlight boosting function, that may further clip the signal.}

\myvspace{1em}

\subsection{Compositing}
\label{sec:compositing}

\begin{figure}[htb]
    \centering
    \begin{subfigure}[t]{0.32\columnwidth}
        \centering
        \includegraphics[width=\textwidth]{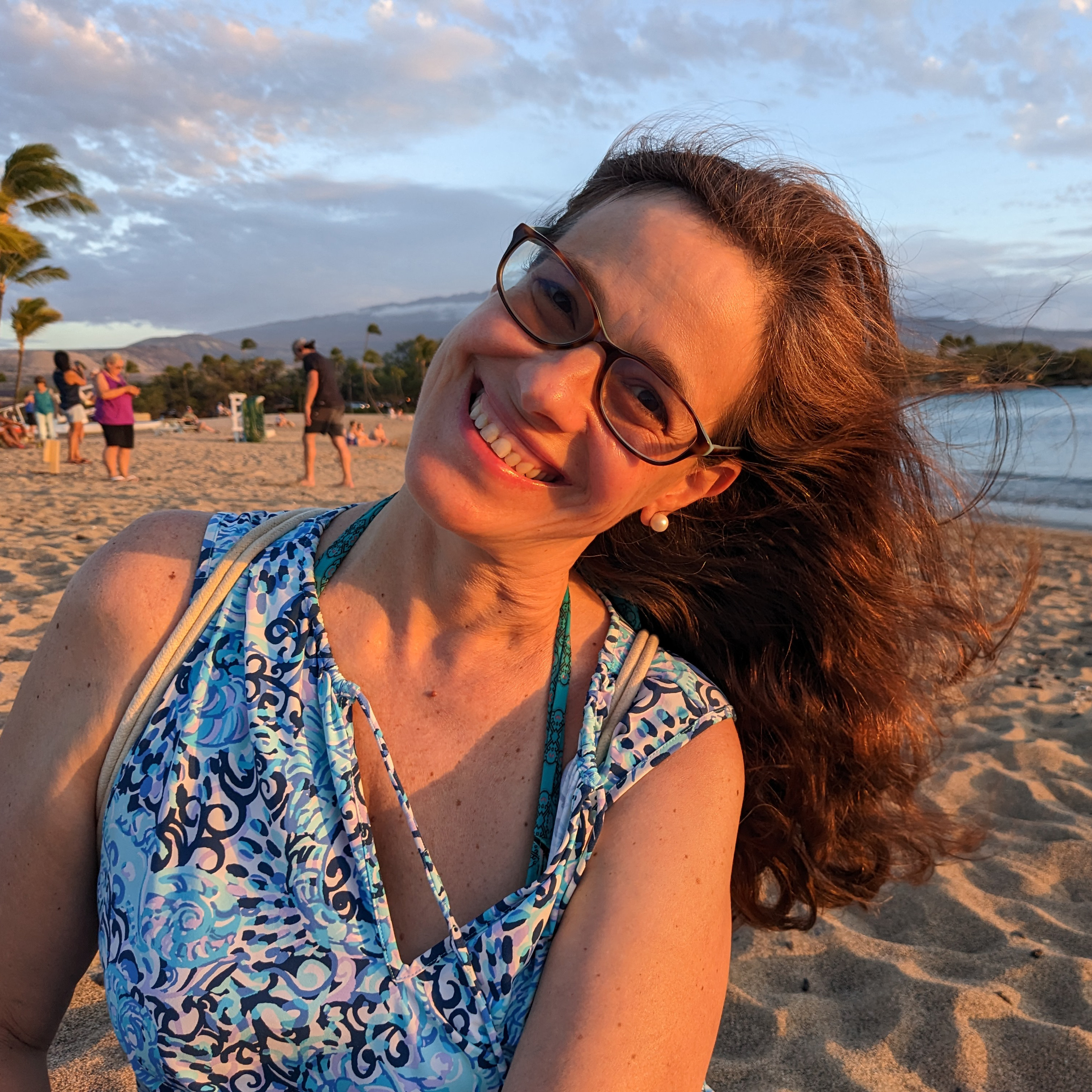}
        \myvspace{-1.5em}
        \subcaption[]{Fully sharp}
        \label{subfig:fully_sharp}
    \end{subfigure}
    \begin{subfigure}[t]{0.32\columnwidth}
        \centering
        \includegraphics[width=\textwidth]{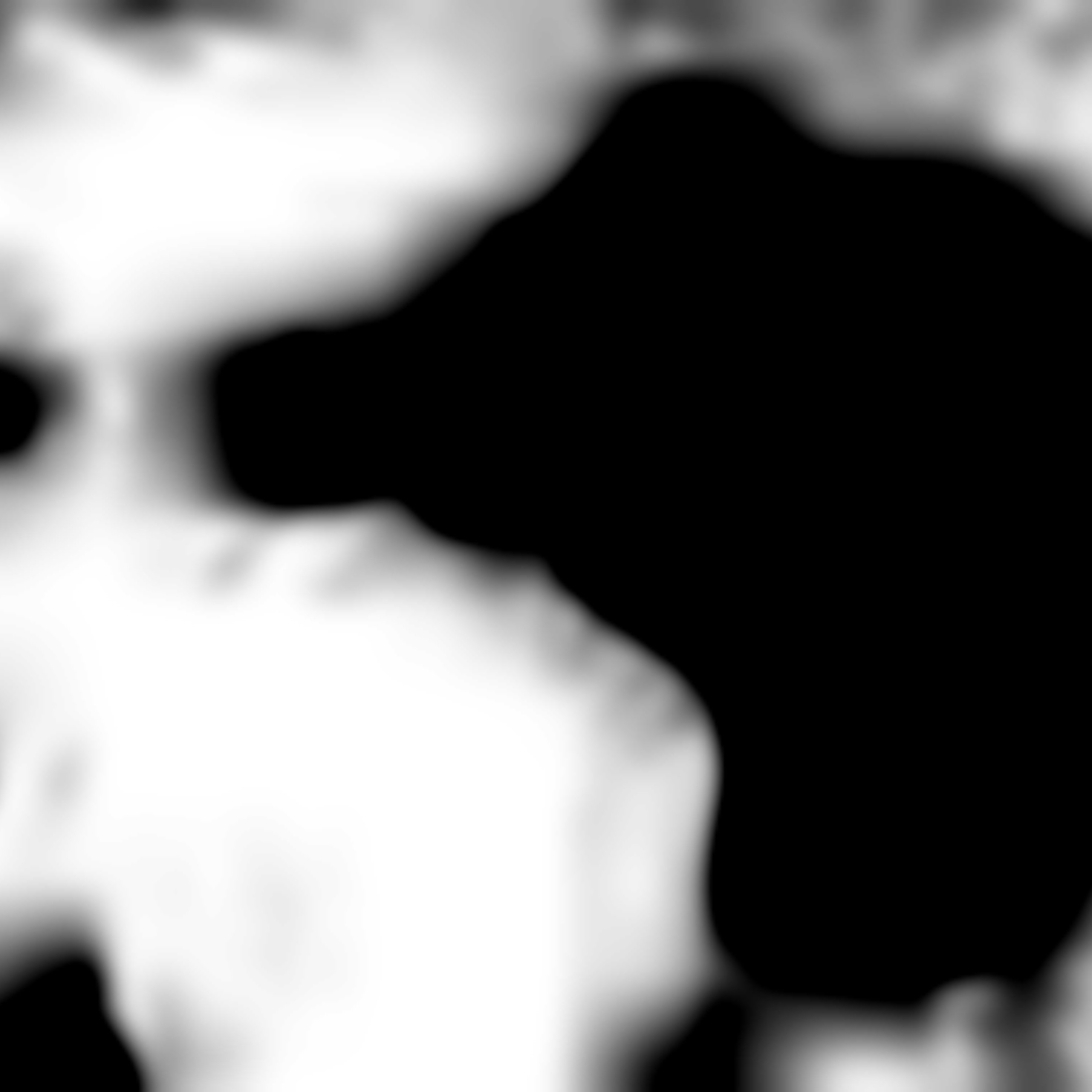}
        \myvspace{-1.5em}
        \subcaption[]{Flow mask}
        \label{subfig:flow_mask}
    \end{subfigure}
    \begin{subfigure}[t]{0.32\columnwidth}
        \centering
        \includegraphics[width=\textwidth]{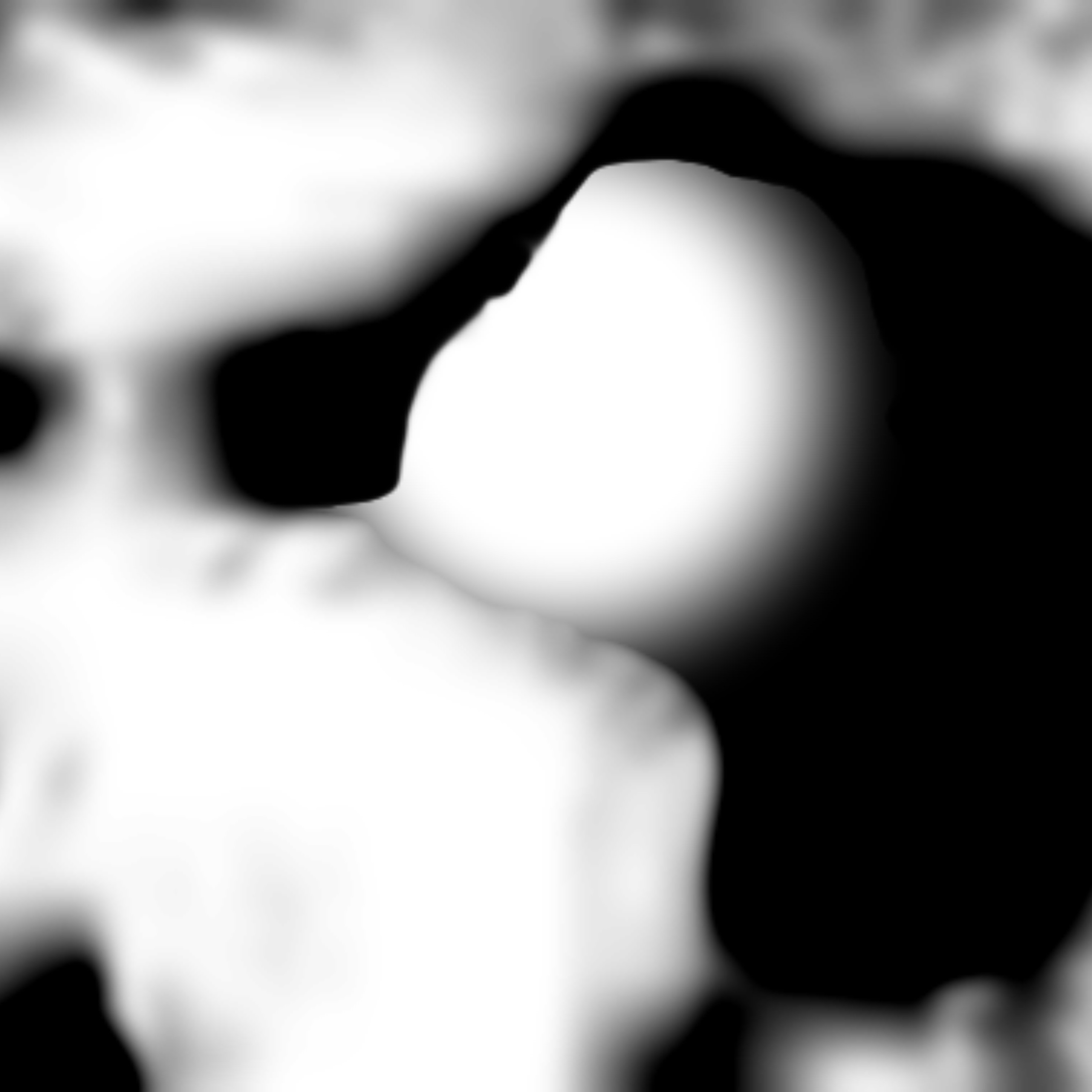}
        \myvspace{-1.5em}
        \subcaption[]{Flow+face mask}
        \label{subfig:flow_face_mask}
    \end{subfigure}\\
    \myvspace{0.5em}
    \begin{subfigure}[t]{0.32\columnwidth}
        \centering
        \includegraphics[width=\textwidth]{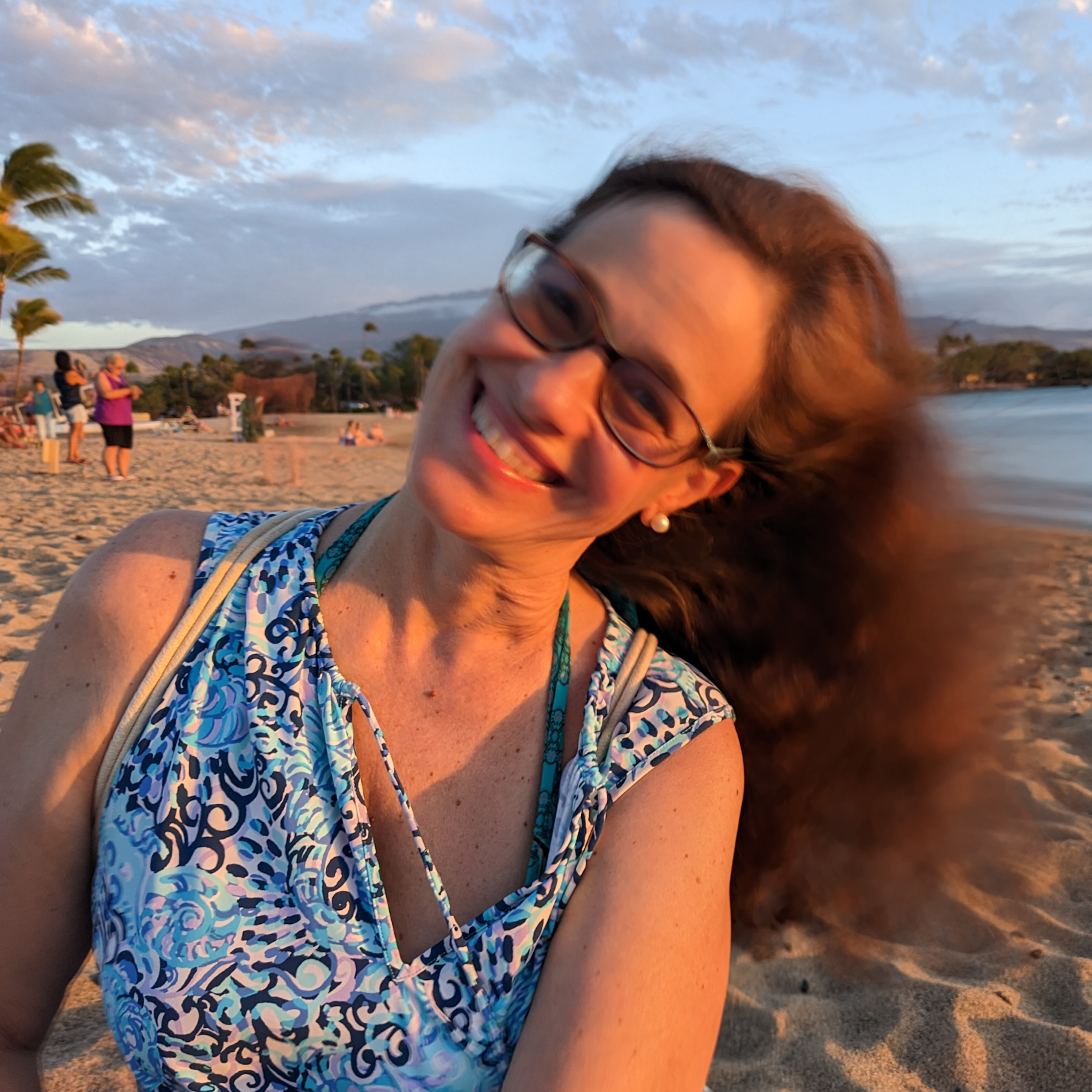}
        \myvspace{-1.5em}
        \subcaption[]{Fully blurred}
        \label{subfig:fully_blurred}
    \end{subfigure}
    \begin{subfigure}[t]{0.32\columnwidth}
        \centering
        \includegraphics[width=\textwidth]{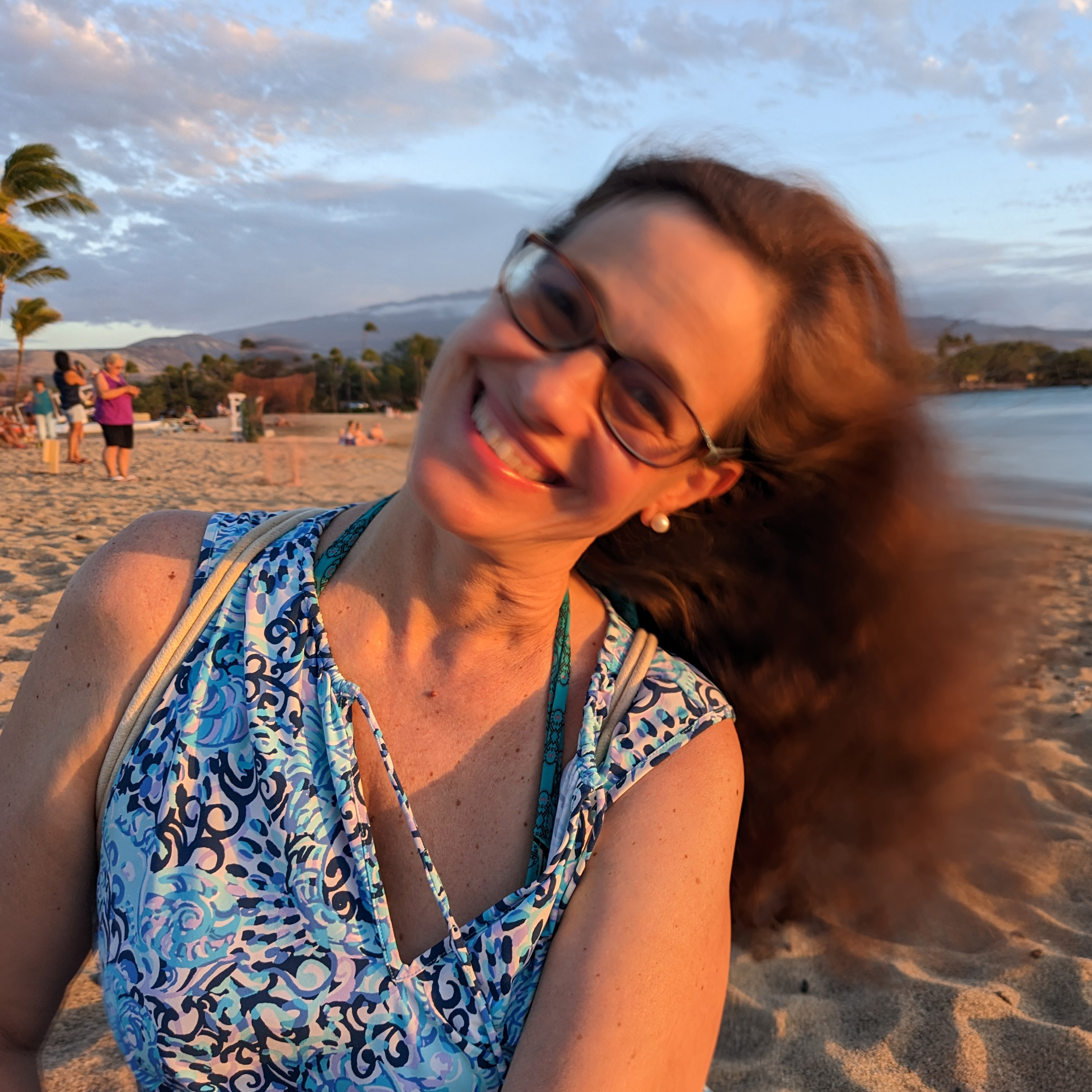}
        \myvspace{-1.5em}
        \subcaption[]{Flow protected}
        \label{subfig:flow_protected}
    \end{subfigure}
    \begin{subfigure}[t]{0.32\columnwidth}
        \centering
        \includegraphics[width=\textwidth]{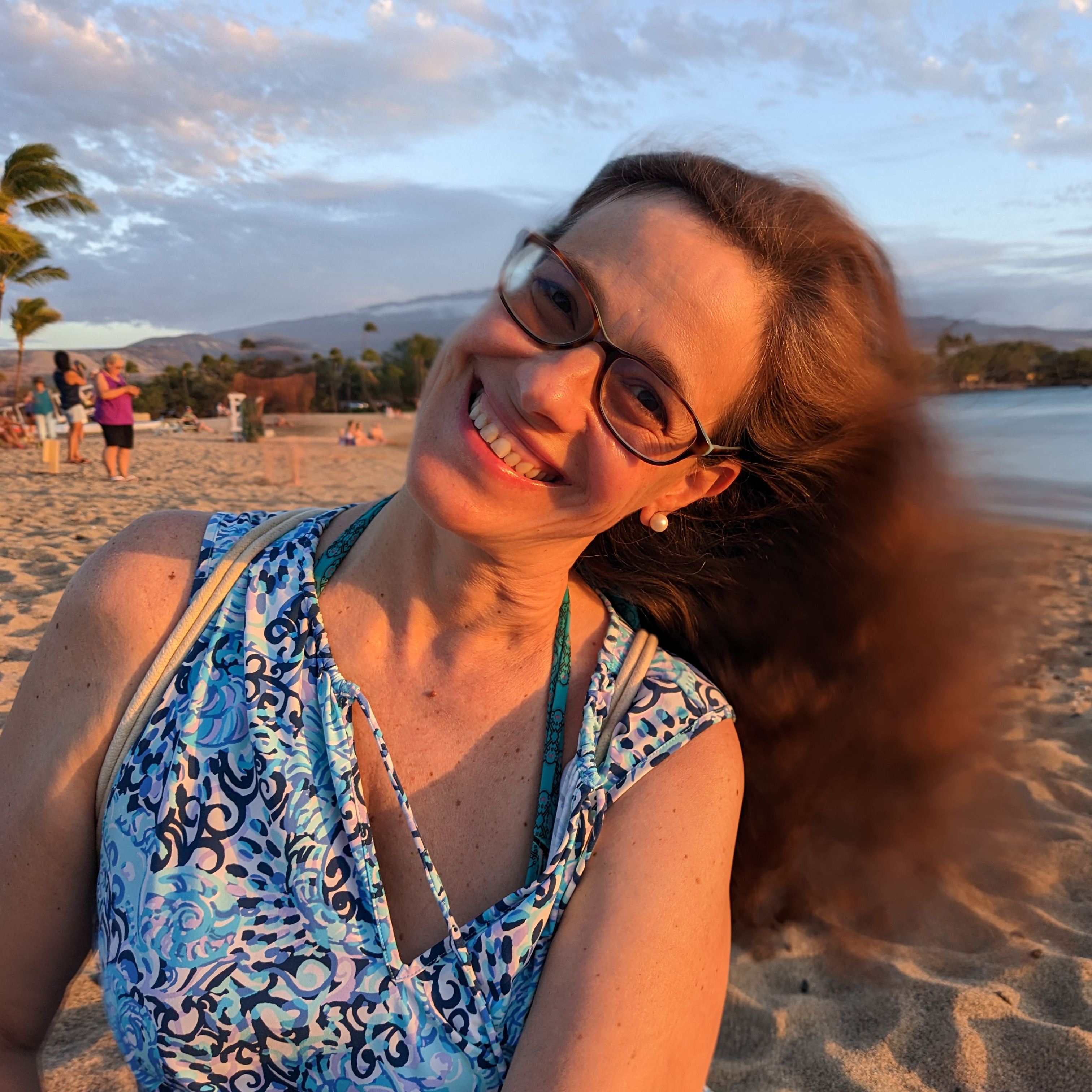}
        \myvspace{-1.5em}
        \subcaption[]{Flow+face protected}
        \label{subfig:flow_face_protected}
    \end{subfigure}
    \caption{Compositing. The regular exposure (\subref{subfig:fully_sharp}) is composited with the synthetically blurred image (\subref{subfig:fully_blurred}) to produce our final output with details preserved. A flow-based mask (\subref{subfig:flow_mask}) protects nearly motionless image regions to produce (\subref{subfig:flow_protected}), note the preserved texture detail in the subject's dress. Further including face signals in the mask (\subref{subfig:flow_face_mask}) also preserves moving, but semantically important image regions (\subref{subfig:flow_face_protected}).}
    \label{fig:compositing}
\end{figure}

The synthetically blurred image described in \sect{sec:rendering} is computed at half resolution to satisfy device memory and latency constraints.
Accordingly, even perfectly aligned, zero-motion regions of the blurred image will lose detail due to \tacc{the upsampling of the result computed at half resolution}. To preserve details, we composite the blurred image with a maximally sharp regular exposure where we expect things to be sharp. \tacc{Two categories need this protection: 1) stationary scene content, and 2) semantically important subjects with little movement, as shown in \fig{fig:compositing}}.

\myvspace{1em}

For category 1, we produce a mask of pixels with very little motion across the entire set of frame pairs, $M_\texttt{flow}$:
    \begin{enumerate}
        \item Compute a per-pixel maximum motion magnitude $|F|$ across all frame pairs.
        \item Compute a reference motion magnitude $|F|_{\texttt{ref}}$ that's effectively a robust max over all pixels in $|F|$ (i.e., 99th percentile).
        \item Rescale and clamp the per-pixel motion magnitudes such that anything below $\alpha|F|_\texttt{ref}$ is mapped to $0$ and anything above $\beta|F|_\texttt{ref}$ is mapped to $1$. \tacc{We use values $0.16$ and $0.32$ for $\alpha$ and $\beta$ respectively.}
        \begin{equation*}
            M_\texttt{flow} = \frac{|F| - \alpha |F|_\texttt{ref}}{\beta|F|_\texttt{ref} - \alpha|F|_\texttt{ref}}
        \end{equation*}
        \item Apply a bilateral blur using the sharp image as a guide~\cite{He13}, to ensure that any edges in $M_\texttt{flow}$ correspond to real edges and minimize artifacts where the flow field is unreliable (e.g., uniform or texture-less regions like skies).
    \end{enumerate}

Category 2 is more complicated and breaks from the physical behavior of optical motion blur in favor of aesthetics. E.g., if a scene has two subjects moving with different trajectories, it would be impossible to sharply align on both simultaneously. Even a single subject can be impossible to align due to movement within the subject, e.g., changes in facial expression, etc. An image with a blurry subject face is a (bad) blurry image. Our solution is to reuse the semantic face signal described in \ref{sec:subject_detection}, modified to only include the faces that have low average feature movement in the aligned reference frame.

Finally, we combine the flow and clipped face masks with a simple $\max$ operator. \fig{fig:compositing} shows the cumulative effect of the two mask types on the final composite.

\begin{figure*}[htb]
  \centering
  \includegraphics[width=1.0\textwidth]{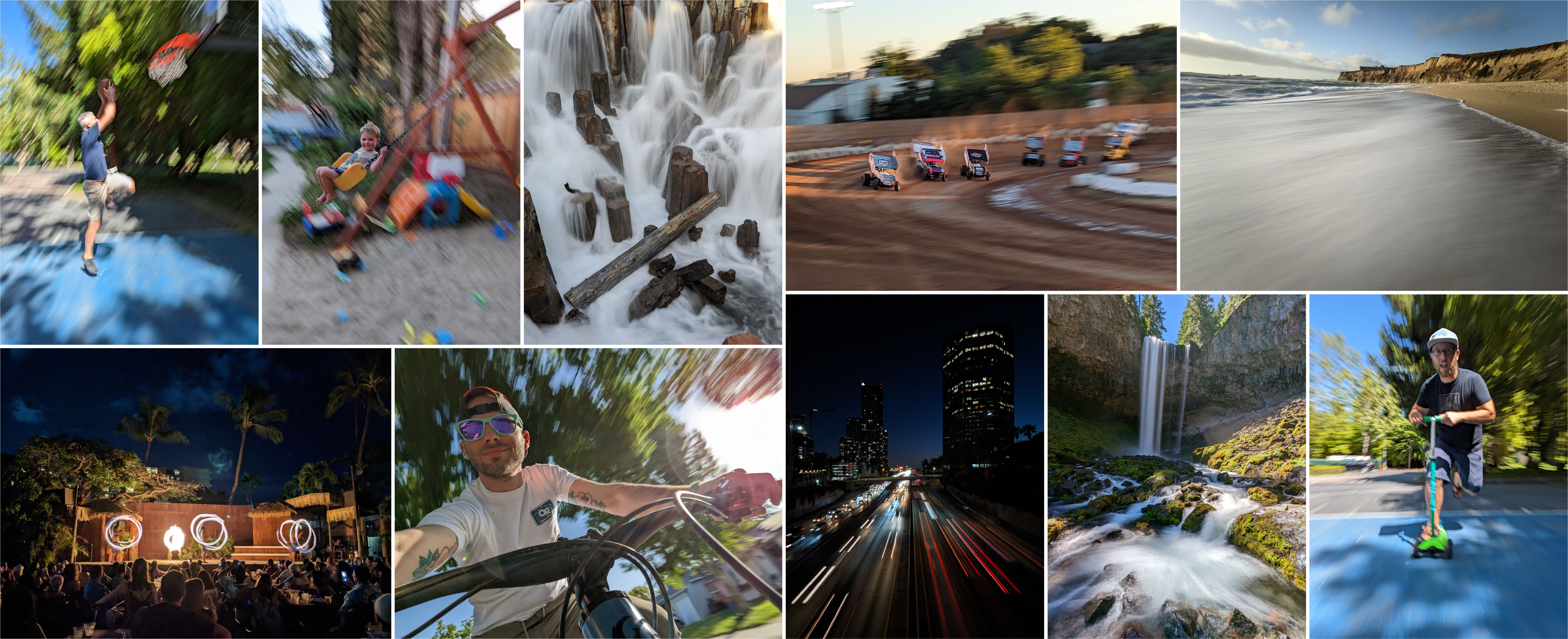}
  \myvspace{-1.5em}
  \caption{Several foreground and background blur examples produced by our system. Several more examples can be found in the supplementary material, along with the corresponding regular exposure\tacc{ and a tone-mapped input burst example. Input RAW burst examples are available on our project webpage: \url{https://motion-mode.github.io/}}.}
  \label{fig:wow}
\end{figure*}

\section{Results}
\label{sec:results}

\fig{fig:wow} shows several foreground and background blur typical use cases, captured and processed using our system. The bursts were all captured hand-held and the results were generated fully automatically, without the need to adjust any settings. In both cases, what makes these long exposure photographs successful is the presence and contrast between sharp and blurry elements.

\tacc{The on-device latency of our system varies according to the number of frames selected for processing. The latency for the main stages (see \fig{fig:system_overview}), measured on 
\ifanonymized
our smartphone (anonymized during review),
\else
a Google Pixel 7 device,
\fi
are as follows. Subject detection, including 8x downsampling and tone-mapping of the base frame: 330ms; motion tracking and alignment, including 8x downsampling and tone-mapping: 55ms per frame; inter-frame motion prediction, including concurrent 2x downsampling and RAW to linear conversion: 77ms per selected frame pair; rendering: 42ms per selected frame pair; final upsampling, compositing and tone-mapping of both image results: 790ms. In the background blur case, a small number of frames are typically selected (e.g. 4), leading to a short total latency (e.g. 1.7s). In the foreground blur case, a higher number of frames are typically selected (e.g. 12) but most of the processing is happening during the extended capture (see \sect{sec:burst_capture}) and the latency is therefore largely hidden from the user.}


\subsection{Track Weights Comparison}
In the following ablation, we compare the effect of including face-region upweighting in motion track weight maps for background-blur alignment quality. (Please refer to \sect{sec:subject_detection} for more details).

\begin{figure}[htb]
    \centering
    \begin{subfigure}[t]{0.49\columnwidth}
        \includegraphics[width=1.0\columnwidth]{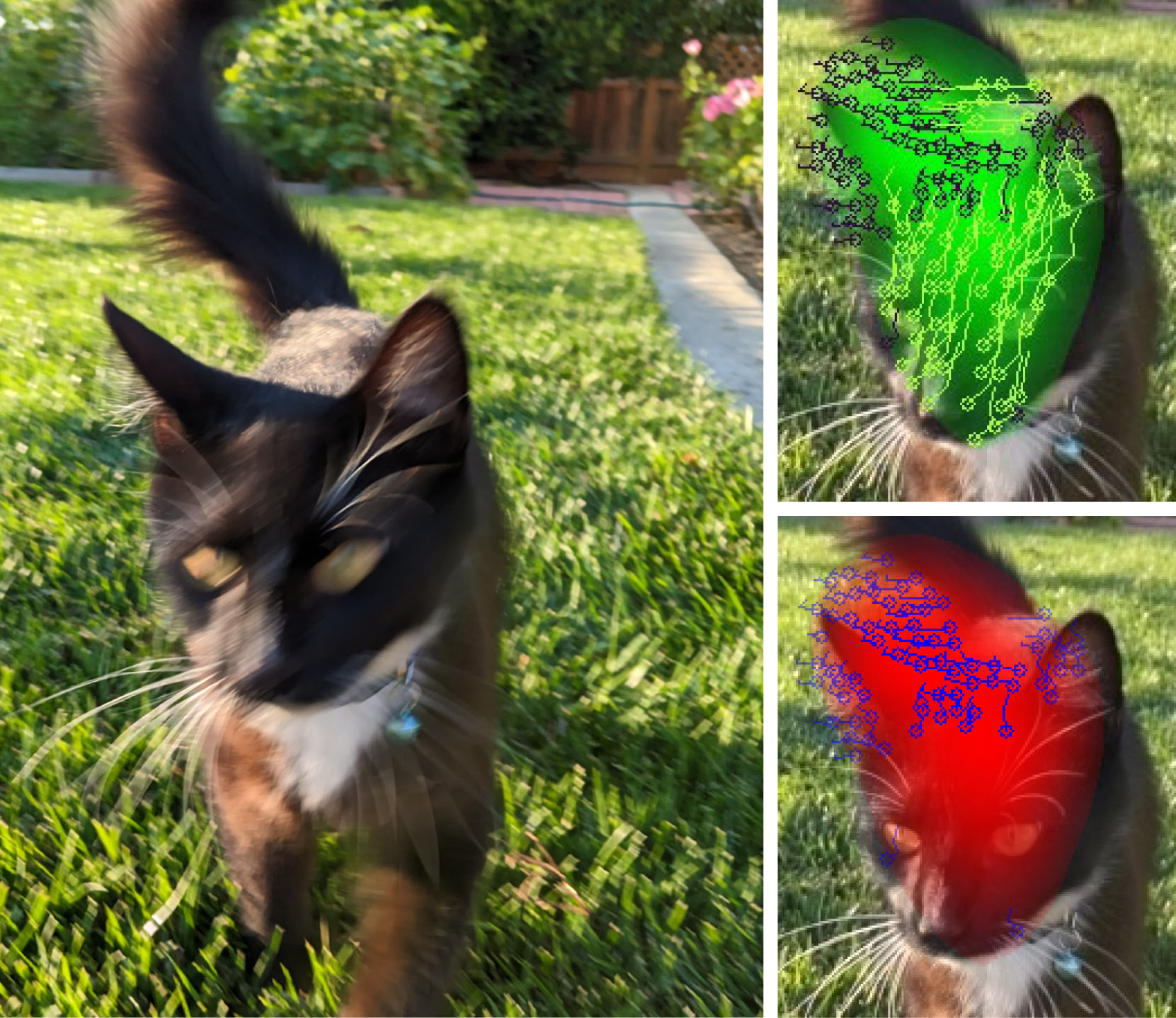}
        \myvspace{-1.25em}
        \subcaption{Without face semantic masking}
        \label{subfig:without_upweighting}
    \end{subfigure}
    \hfill
    \begin{subfigure}[t]{0.49\columnwidth}
        \includegraphics[width=1.0\columnwidth]{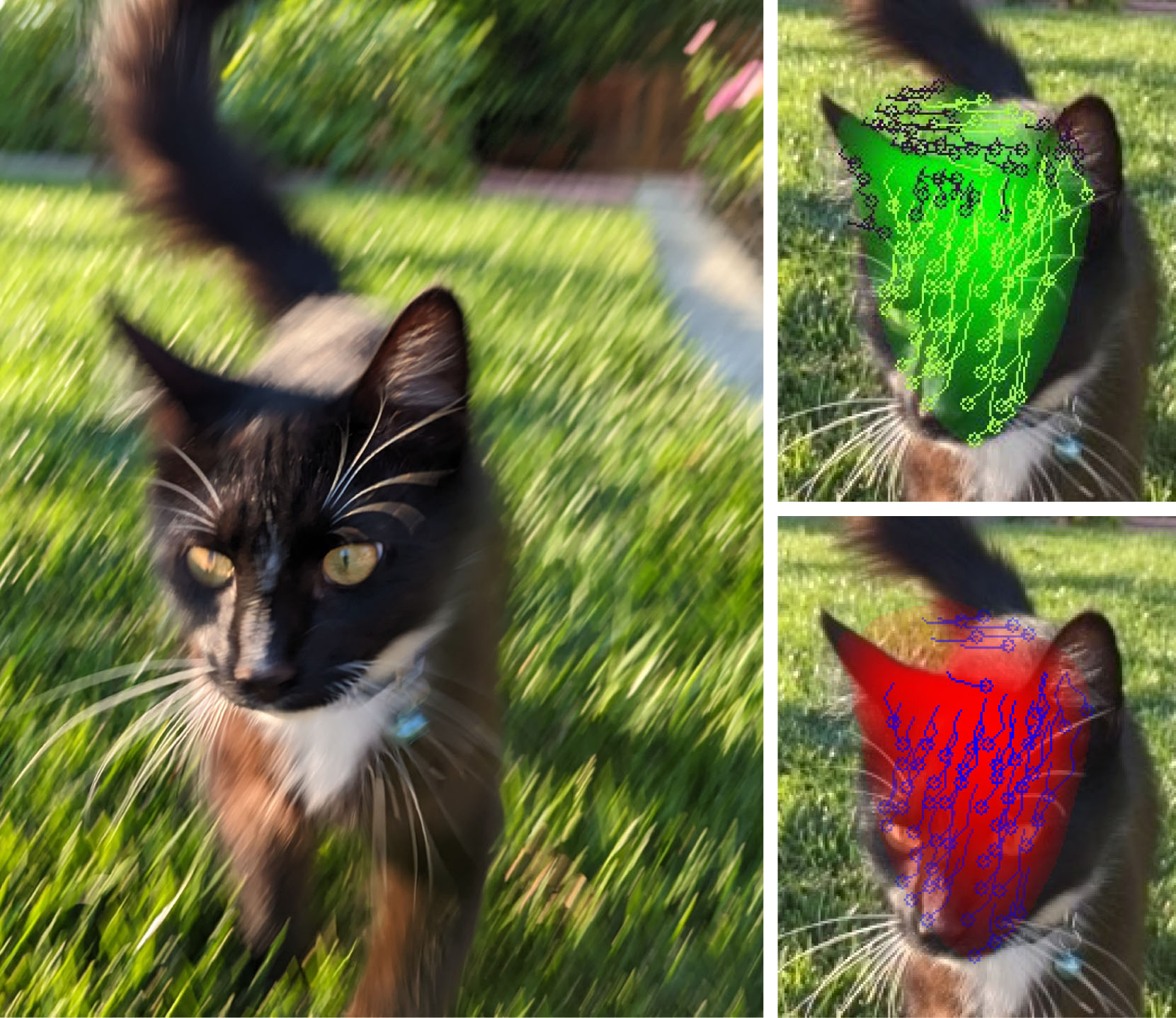}
        \myvspace{-1.25em}
        \subcaption{With face semantic masking}
        \label{subfig:with_upweighting}
    \end{subfigure}
    \myvspace{-0.75em}
    \caption{Up-weighting track weights in faces helps subjects with complex/articulated motion. Larger images are the long exposure results \tacc{rendered with compositing masking disabled (as in \fig{subfig:fully_blurred}) for clarity of the comparison. To their right are the intermediate outputs:} track weights are visualized as added green/red channel values, motion track clusters in the top image, and the highest weight selected cluster in the bottom image. (\subref{subfig:without_upweighting}) Gaze saliency alone peaks the weights on the cat's center, assigning the highest weight to the tracks cluster on the cat's body - resulting in undesirable alignment. (\subref{subfig:with_upweighting}) Our pipeline's results: up-weighting the face-region causes its motion cluster to be selected, resulting in the desired long exposure tracking the cat's face.}
    \myvspace{-1em}
    \label{fig:saliency}
\end{figure}

We find that including both gaze saliency and face detections in the motion track weight map benefits image subjects with complex articulated motion (which can cause the wrong part of the subject to be tracked). A representative example is shown in \fig{fig:saliency}.

\subsection{Image Alignment Comparison}%
\label{sec:alignment_comparison}

In \fig{fig:alignment_ablations}, we showcase additional examples of image alignment on background-blur scenes, comparing the aesthetics of results when the regularization term $E_b$ from Eq.~\ref{eqn:alignment_objective_fn} is excluded (left) and included (right). In the left-hand side column of \fig{fig:alignment_ablations}, we observe that optimizing just for the subject's sharpness $E_b$ doesn't account for the background of the scene. Consequently, sudden changes in transform parameters over time are allowed, resulting in different parts of the field of view having motion blur in completely different directions.
By adding the temporal regularization term $E_b$, we get the results on the right-hand side column of \fig{fig:alignment_ablations} with consistent blur trails. The second example showcases the effect of dampening the rotational parameter, avoiding the blur vortex (green insets).


\subsection{Motion Prediction Comparison}

We compare models described in \sect{sec:motion_prediction} with those from~\cite{Brooks19} that use uniform weights, labelled "BB19-uni.", and that learn \(N = 17\) weights per input image, labelled "BB19".

All the compared models were trained with the same hyper-parameters described in~\cite{Brooks19}. To supervise the training, we generate a bracketed dataset of input image triplets from many videos, as described in~\cite{Reda22}, synthesizing the pseudo ground-truth motion-blurred image using a previously trained FILM frame interpolation model. To evaluate our model we used a test set with 2000 examples and report the PSNR and SSIM, i.e. comparing the synthesized motion-blur image to the pseudo ground-truth, in Table~\ref{table:model_comparison}.

\begin{table}[htb]
  \begin{center}
    \caption{Comparing motion prediction models evaluation performance (PSNR and SSIM) and properties: number of learned parameters in millions (M-Par.), number of floating point operations at the evaluated image input resolution in billions (B-Flop) and receptive field computed based on~\cite{Dumoulin16} in pixels (Rec. Field). Our simplified model to run on mobile devices, shows comparable quality performance to models with 1.7 times as many parameters, both quantitatively and qualitatively (shown in \fig{fig:model_comparisons}).}
    \myvspace{-0.5em}
    \begin{tabular}{l|c|c|c|c|c}
      Model        &  PSNR.  &  SSIM    &  M-Par. &  B-Flop  &  Rec. Field\\
      \hline \hline
      BB19         &  41.78  &  0.9862  &  7.057  &  107.28  &  202\\
      BB19-uni.    &  40.07  &  0.9803  &  7.056  &  106.81  &  202\\
      \hline
      Ours-large.  &  41.32  &  0.9842  &  7.301  &  107.40  &  202\\
      Ours         &  40.78  &  0.9823  &  4.173  &  114.67  &  128\\
      Ours-abl.    &  40.61  &  0.9808  &  4.173  &  114.67  &  128\\
      \hline
    \end{tabular}
    \label{table:model_comparison}
  \end{center}
\end{table}

A visual comparison on 512 x 384 image crops is provided in \fig{fig:model_comparisons} and shows that our model performs visually similarly to "BB19" (e.g. blur smoothness, handling of dis-occlusions), despite the significant simplifications of our implementation to run on mobile devices. It also reveals that both models "Ours-abl." and "BB19-uni." suffer from the same discontinuity artifacts in the middle of blur trails, which are described in \sect{sec:motion_prediction}.

Our model runs in under 80ms on a
\ifanonymized
mobile device's hardware accelerator (anonymized during review)
\else
Google Pixel 7 mobile device's TPU~\cite{Gupta21},
\fi
when given an input image pair from our low resolution image pipeline.


\subsection{Rendering Comparison}

\subsubsection{Motion Interpolation}
\label{subsec:motion_interpolation}

\begin{figure}[htb]
  \centering
  \includegraphics[width=1.0\columnwidth]{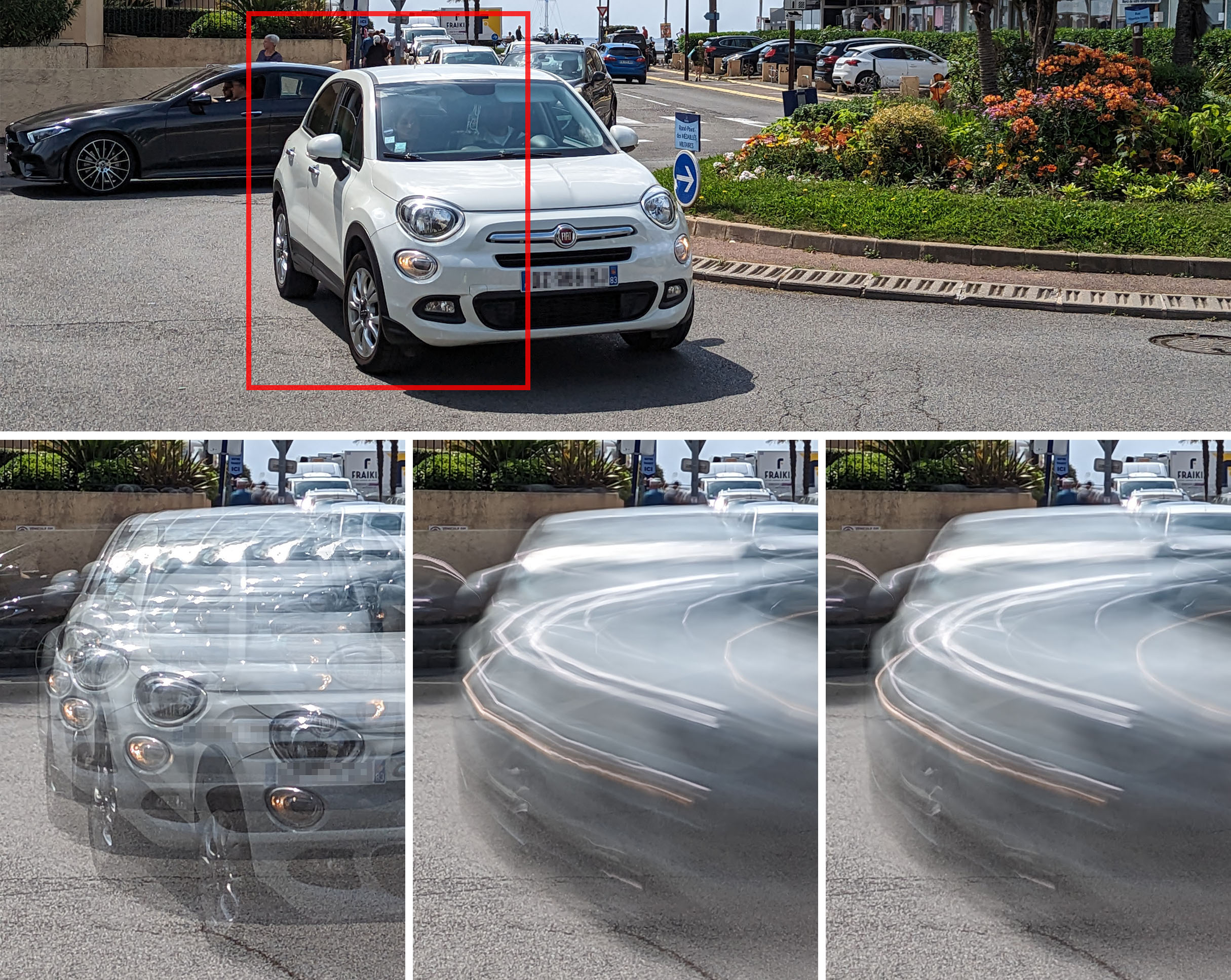}
  \myvspace{-1.5em}
  \caption{Comparison of interpolation methods described in \sect{subsec:spline_interpolation} on a scene with a car traversing a roundabout. Top: An intermediate frame from the input burst. Bottom, left to right: average of several successive input frames, linear flow interpolation, spline flow interpolation. Piecewise-linear motion trails look synthetically generated, revealing the number of input frame pairs used to render the image, whereas curved motion trails look optically generated as in a single continuous long exposure. See supplementary material for more examples.}
  \label{fig:linear_vs_curved}
\end{figure}

In \fig{fig:linear_vs_curved}, we compare the effects of piecewise-linear flow interpolation vs cubic spline interpolation. Particularly when camera or object motion from frame to frame is irregular, spline interpolation can impart a much more natural and photorealistic appearance to the motion trails.

\myvspace{1em}

\begin{figure}[htb]
  \centering
  \begin{subfigure}[t]{0.327\columnwidth}
    \begin{subfigure}[t]{1.0\columnwidth}
      \includegraphics[width=1.0\columnwidth]{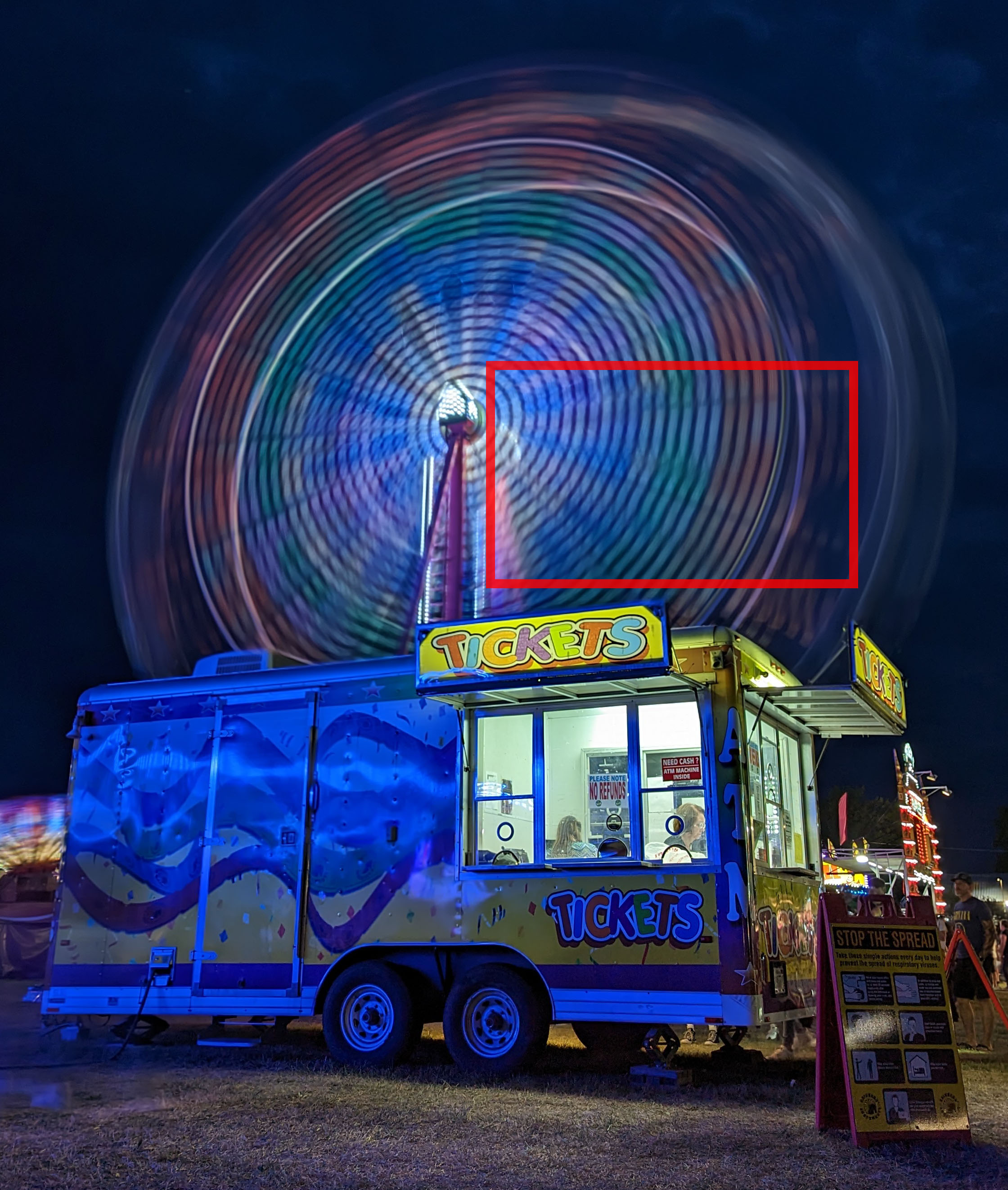}
      \myvspace{-1.05em}
    \end{subfigure}
    \begin{subfigure}[t]{1.0\columnwidth}
      \includegraphics[width=1.0\columnwidth]{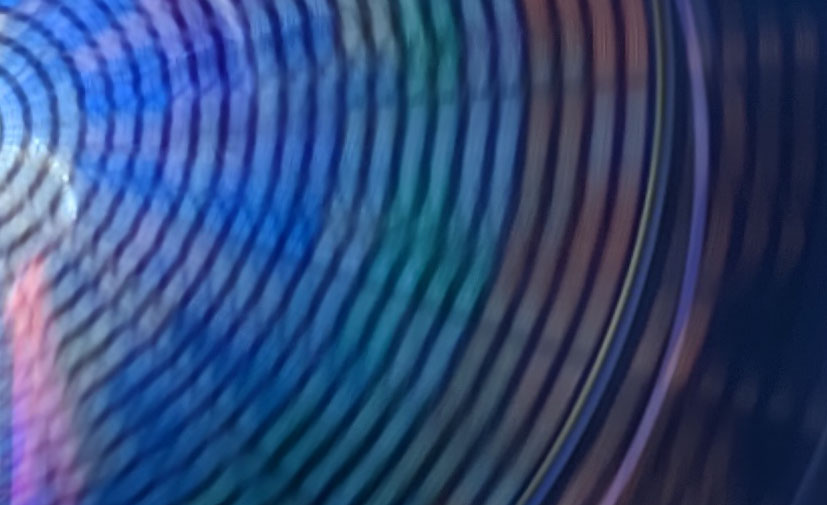} 
    \end{subfigure}
    \myvspace{-1.5em}
    \subcaption[]{sRGB}
    \label{subfig:srgb_colorspace}
  \end{subfigure}
  \hfill
  \begin{subfigure}[t]{0.327\columnwidth}
    \begin{subfigure}[t]{1.0\columnwidth}
      \includegraphics[width=1.0\columnwidth]{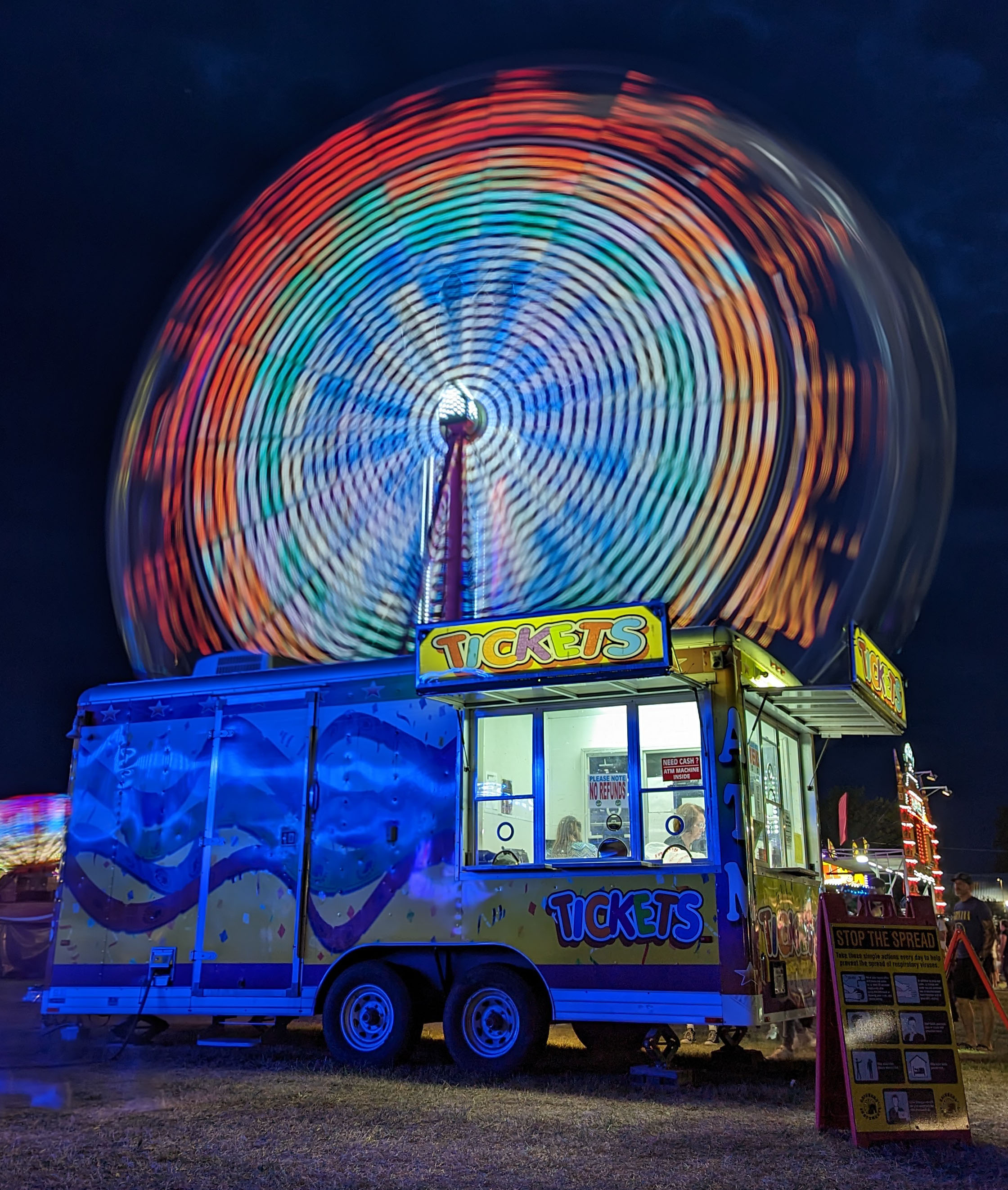}
      \myvspace{-1.05em}
    \end{subfigure}
    \begin{subfigure}[t]{1.0\columnwidth}
      \includegraphics[width=1.0\columnwidth]{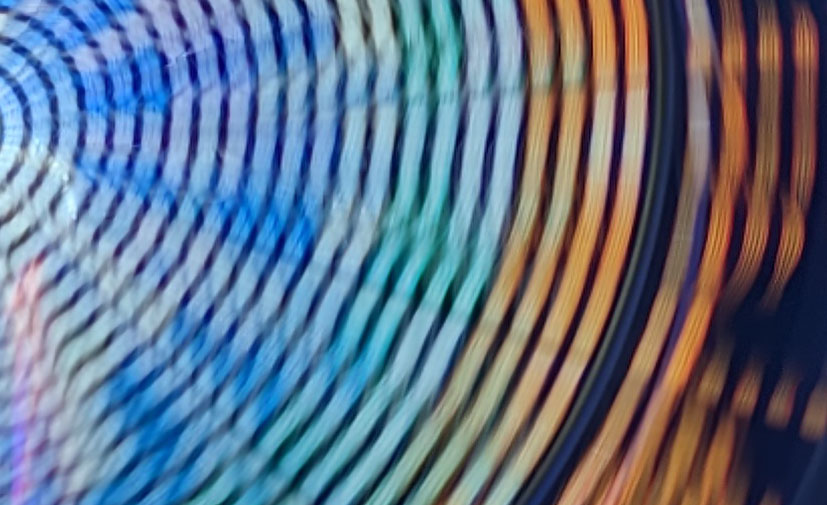}
    \end{subfigure}
    \myvspace{-1.5em}
    \subcaption[]{Linear}
    \label{subfig:linear_colorspace}
  \end{subfigure}
  \hfill
  \begin{subfigure}[t]{0.327\columnwidth}
    \begin{subfigure}[t]{1.0\columnwidth}
      \includegraphics[width=1.0\columnwidth]{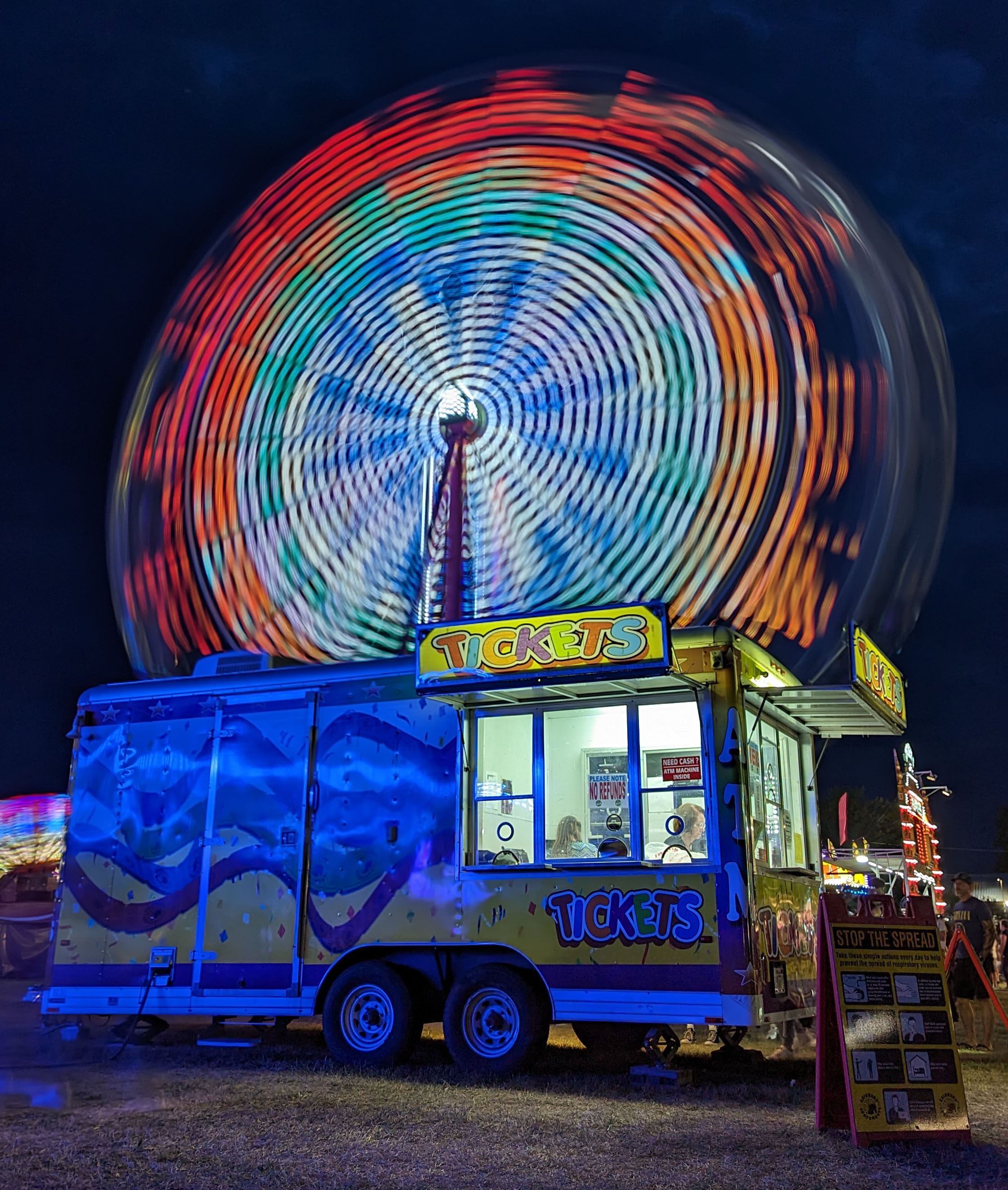}
      \myvspace{-1.05em}
    \end{subfigure}
    \begin{subfigure}[t]{1.0\columnwidth}
      \includegraphics[width=1.0\columnwidth]{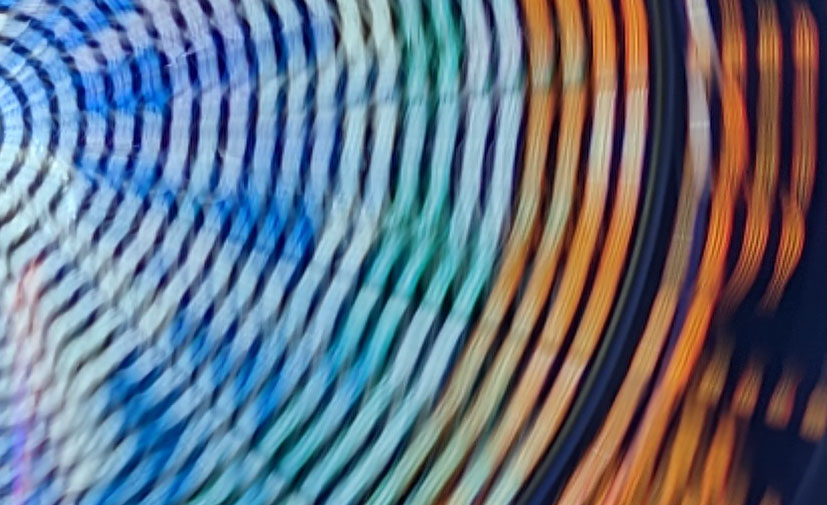}
    \end{subfigure}
    \myvspace{-1.5em}
    \subcaption[]{Soft gamma}
    \label{subfig:soft_gamma_colorspace}
  \end{subfigure}
  \myvspace{-1.0em}
  \caption{\tacc{Colorspace comparison of the blurring operation. (\subref{subfig:srgb_colorspace}) sRGB colorspace blur loses most of the motion-blurred highlights intensity. (\subref{subfig:linear_colorspace}) Linear colorspace is physically correct but produces dull blur trails due to clipping occurring in the sensor prior to blurring. (\subref{subfig:soft_gamma_colorspace}) Soft gamma colorspace blur, described in \sect{subsec:soft_gamma}, is able to preserve strong motion-blurred highlights and increases blur trails contrast and color saturation. See supplementary material for more examples.}}
  \label{fig:linear_vs_gamma}
\end{figure}

\subsubsection{Rendering Colorspace}
\label{subsec:rendering_colorspace}

In \fig{fig:linear_vs_gamma}, we compare the results of performing the blurring operation in a conventional sRGB colorspace, versus \tacc{a linear physically correct colorspace, versus} our non-physical "soft gamma" colorspace, obtained by adjusting the linear-space image in a direction opposite from a usual linear to sRGB color transformation. The figure illustrates how blurring in the soft-gamma colorspace emphasizes and preserves the brightness of the motion trails in the highlights\tacc{ and generally increases their contrast and color saturation}.


\subsection{Comparison to Mobile Phone Camera Applications}
Unlike other works which realize a long exposure effect~\cite{Lee09,Teramoto10,Luo18,Luo20,Lancelle19,Mikamo21}, our pipeline is a responsive mobile phone capture experience. Therefore, we also compare our results to released capture experiences for consumer phones.

\begin{figure}[htb]
    \centering
    \begin{subfigure}[t]{0.495\columnwidth}
        \begin{subfigure}[t]{\columnwidth}
            \includegraphics[width=\columnwidth]{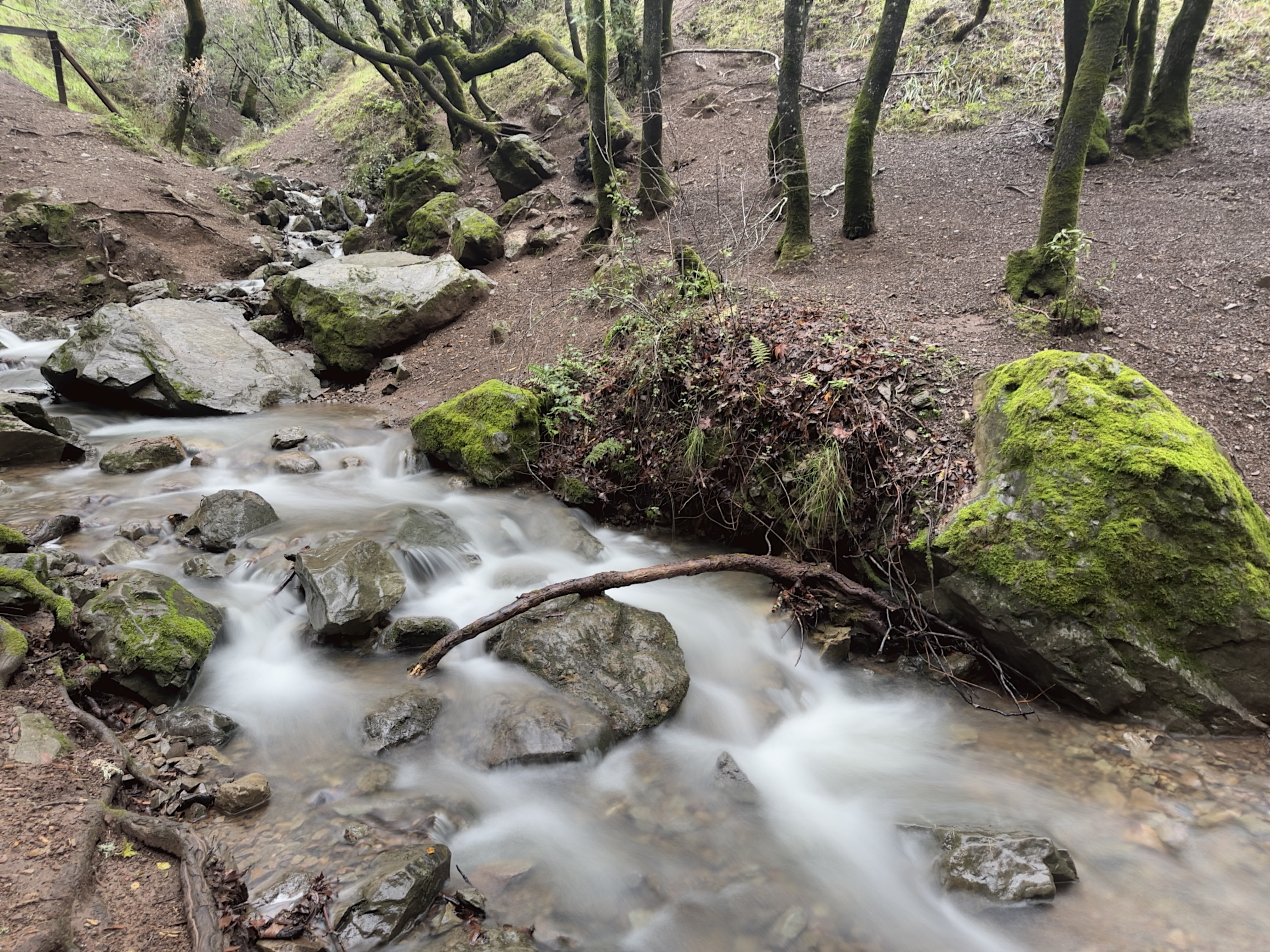}
            \myvspace{-1em}
        \end{subfigure}
        \begin{subfigure}[t]{\columnwidth}
            \includegraphics[width=\columnwidth]{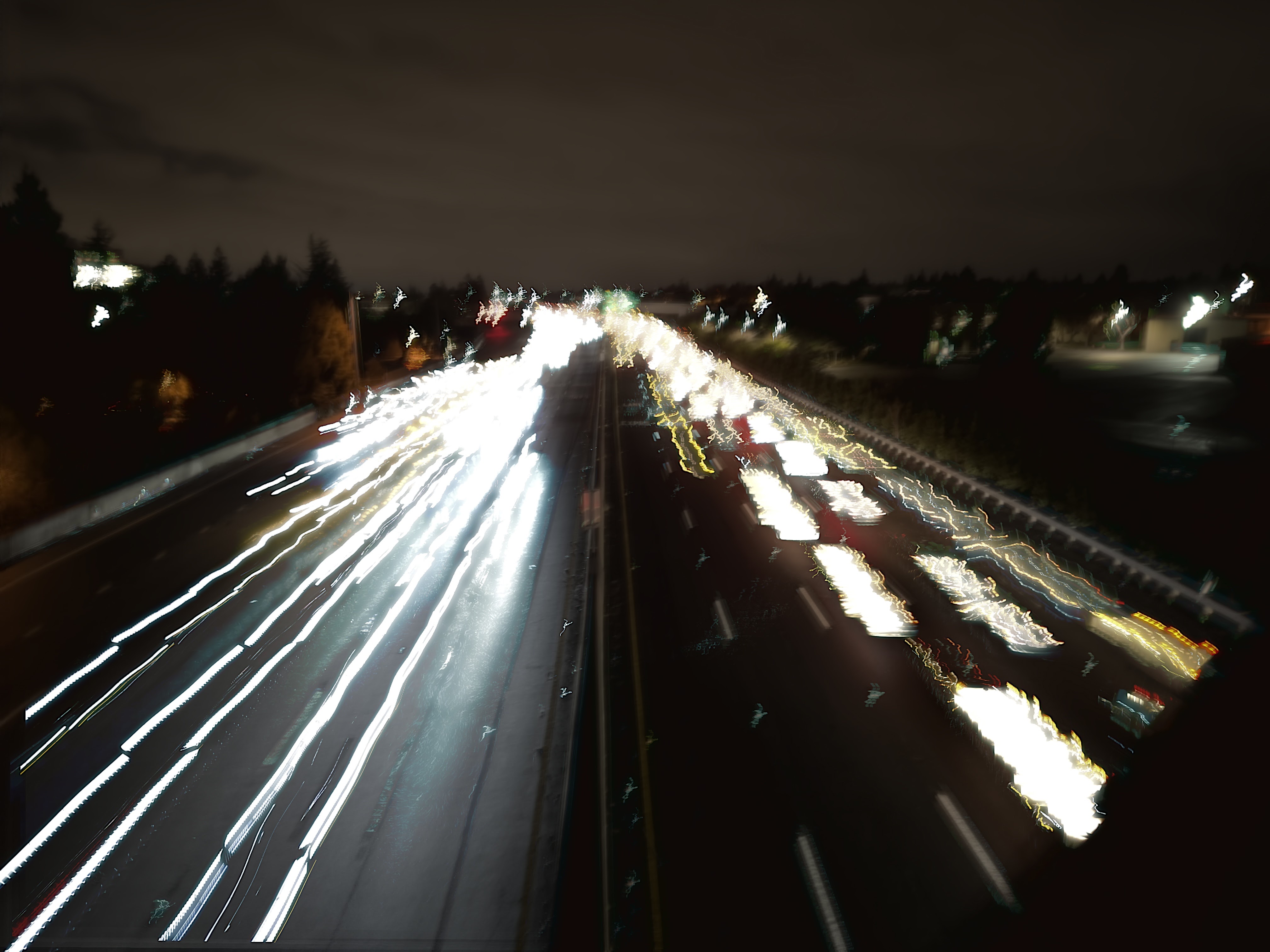}
            \myvspace{-1em}
        \end{subfigure}
        \begin{subfigure}[t]{\columnwidth}
            \includegraphics[width=\columnwidth]{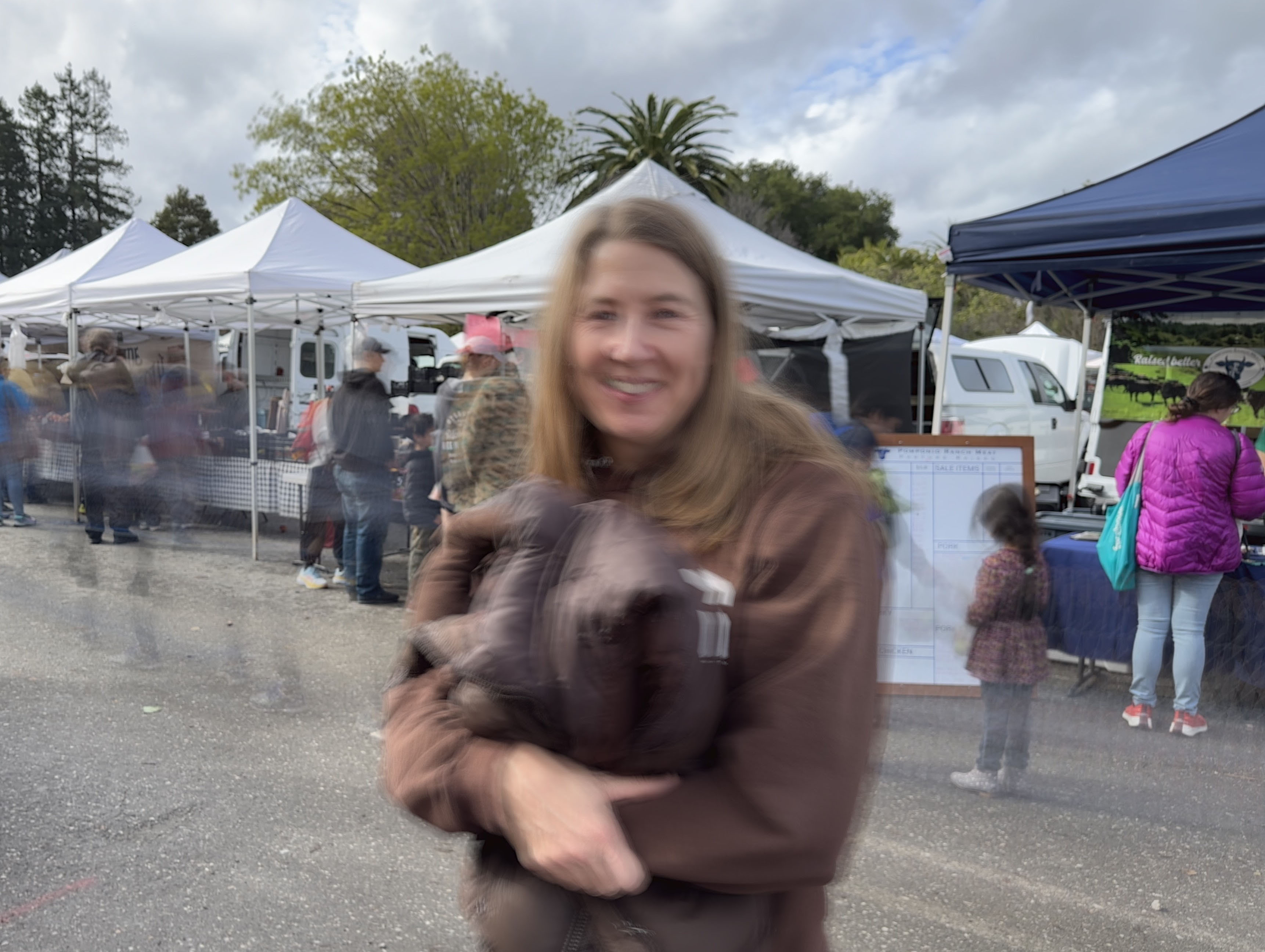}
        \end{subfigure}
        \myvspace{-1.5em}
        \subcaption[]{Spectre}
        \label{subfig:ios_apps_spectre}
    \end{subfigure}
    \hfill
    \begin{subfigure}[t]{0.495\columnwidth}
        \begin{subfigure}[t]{\columnwidth}
            \includegraphics[width=\columnwidth]{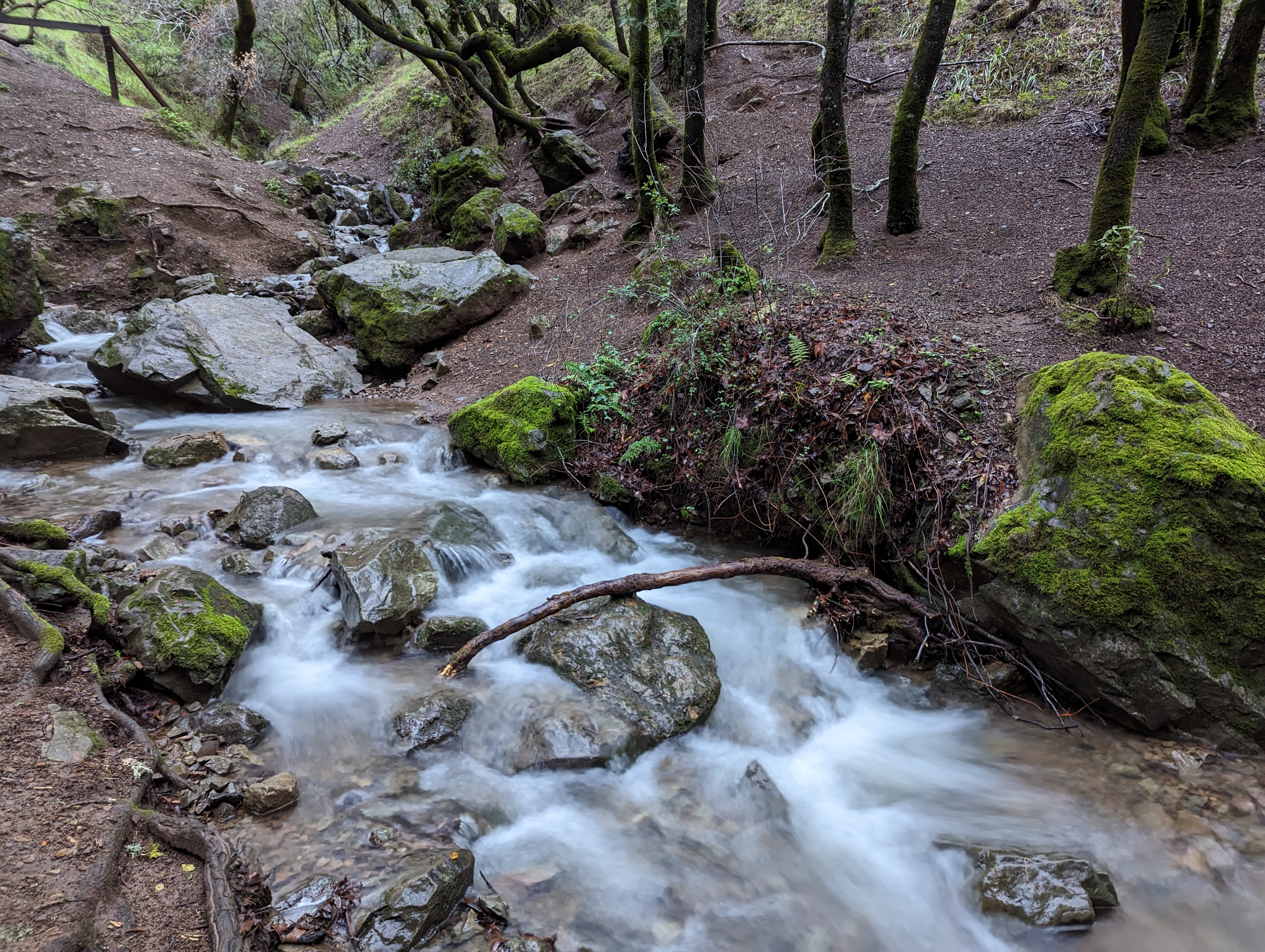}
            \myvspace{-1em}
        \end{subfigure}
        \begin{subfigure}[t]{\columnwidth}
            \includegraphics[width=\columnwidth]{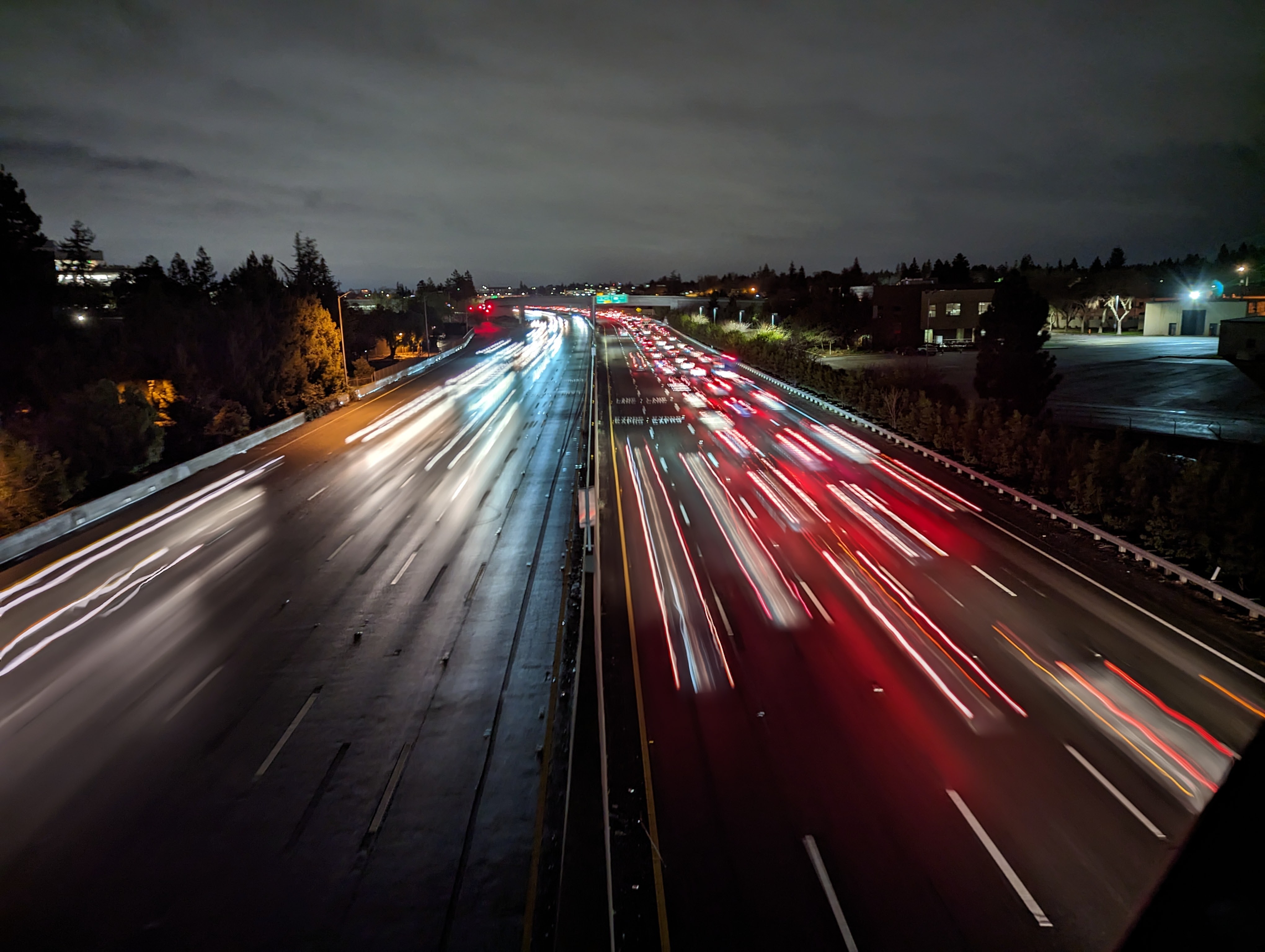}
            \myvspace{-1em}
        \end{subfigure}
        \begin{subfigure}[t]{\columnwidth}
            \includegraphics[width=\columnwidth]{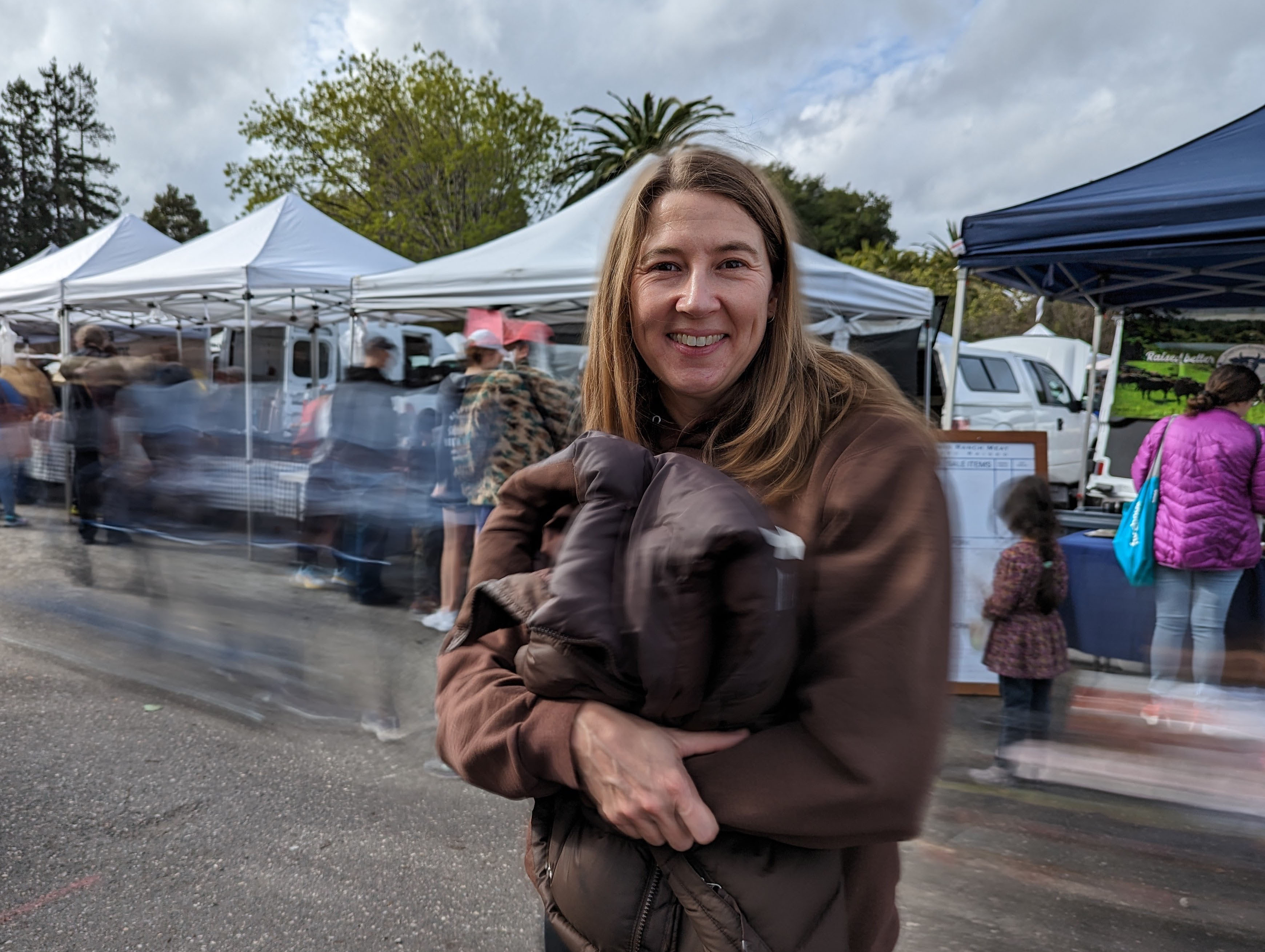}
        \end{subfigure}
        \myvspace{-1.5em}
        \subcaption[]{Ours}
        \label{subfig:ios_apps_ours}
    \end{subfigure}
    \myvspace{-1.0em}
    \caption{\tacc{Comparison of the app Spectre (\subref{subfig:ios_apps_spectre})~\cite{Spectre} vs. our method (\subref{subfig:ios_apps_ours}), on scenes captured hand-held. The light trail scene in the middle row was captured in very windy conditions. Our pipeline shows better background alignment, subject preservation, and more noticeable motion trails. A more extensive comparison with a few additional apps can be found in the supplement.}}
    \label{fig:vs_spectre}
\end{figure}

Several mobile applications allow a more manual control of the capture schedule on mobile phones such as Even Longer, Moment, Neoshot, and Procam 8 (all available in the iOS App Store). These apps do not seem to have frame alignment capability, and therefore require a tripod for capturing sharp long exposure images. Spectre, released on iOS, seems to have stabilization and auto-exposure capabilities. Through a capture study of dozens of scenes, we found the hand-held performance of Spectre to be inconsistent. \fig{fig:vs_spectre} shows representative comparisons of the results of our pipeline with Spectre.

To our knowledge, our pipeline is the only mobile-phone capture experience with all of the following features: background-blur alignment (automatically tracking and keeping a moving subject sharp), robust foreground-blur alignment (keeping the background sharp), motion interpolation for smooth motion trails (compared to results showing temporal undersampling), and face-region sharpness protection (keeping slightly moving subjects sharp).

\tacc{

\subsection{Evaluation on Existing Datasets}
\label{sec:evaluation}

In \fig{fig:evaluation}, we evaluate our method on the publicly available video dataset in~\cite{Liu2014}. The images are also available in the supplement, and can be compared to their results (see their Figure 9), as well as results in~\cite{Lancelle19} (see their Figure 15).

\begin{figure}[htb]
    \centering

    \begin{subfigure}[t]{\columnwidth}
        \begin{subfigure}[t]{0.327\columnwidth}
            \includegraphics[width=1.0\columnwidth]{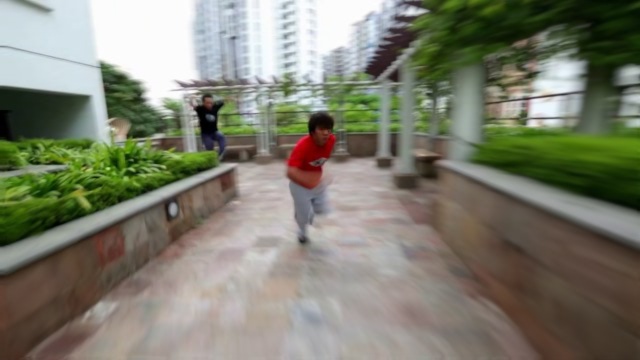}
        \end{subfigure}
        \hfill
        \begin{subfigure}[t]{0.327\columnwidth}
            \includegraphics[width=1.0\columnwidth]{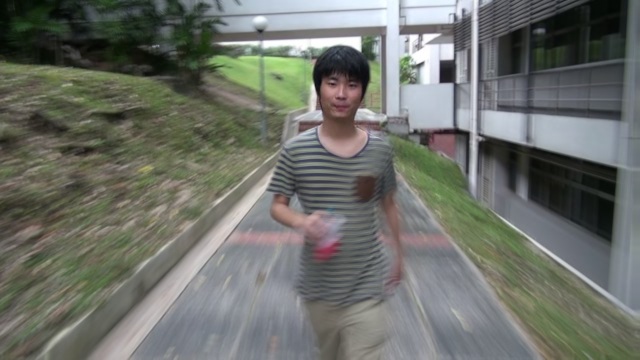}
        \end{subfigure}
        \hfill
        \begin{subfigure}[t]{0.327\columnwidth}
            \includegraphics[width=1.0\columnwidth]{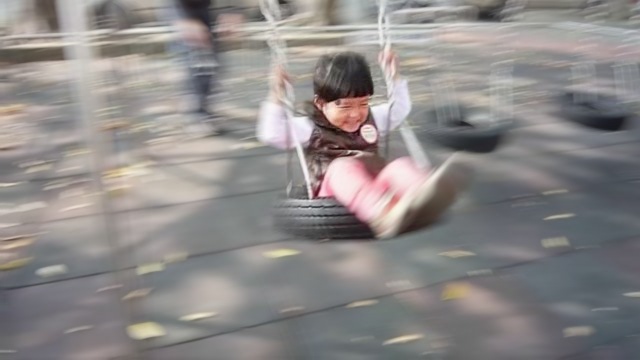}
        \end{subfigure}
    
        \myvspace{0.15em}
    
        \begin{subfigure}[t]{0.327\columnwidth}
            \includegraphics[width=1.0\columnwidth]{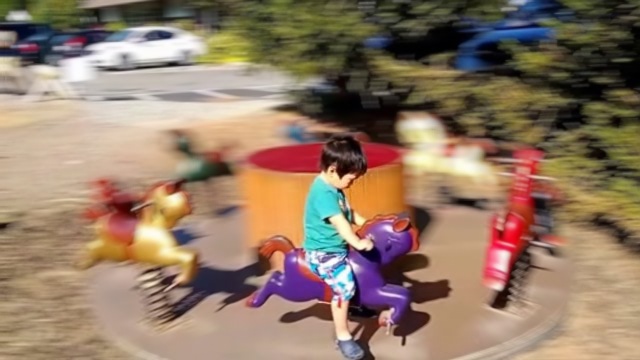}
        \end{subfigure}
        \hfill
        \begin{subfigure}[t]{0.327\columnwidth}
            \includegraphics[width=1.0\columnwidth]{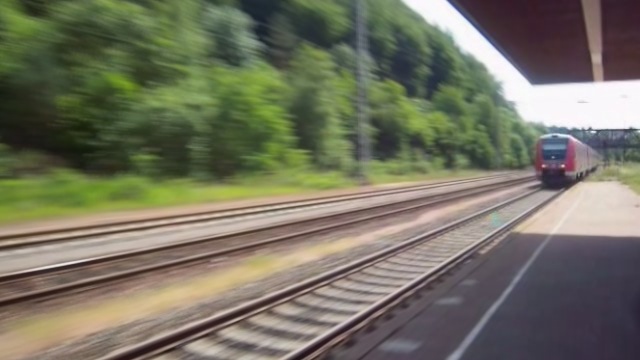}
        \end{subfigure}
        \hfill
        \begin{subfigure}[t]{0.327\columnwidth}
            \includegraphics[width=1.0\columnwidth]{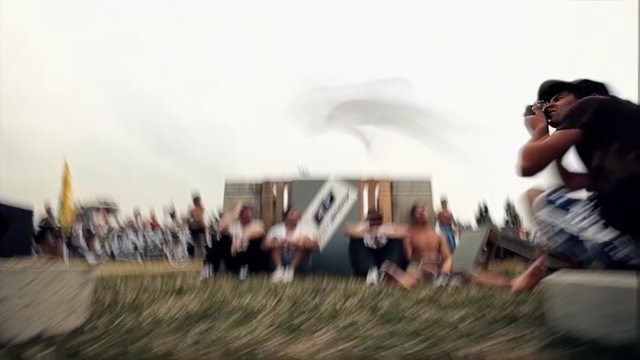}
        \end{subfigure}
    
        \myvspace{0.15em}
    
        \begin{subfigure}[t]{0.327\columnwidth}
            \includegraphics[width=1.0\columnwidth]{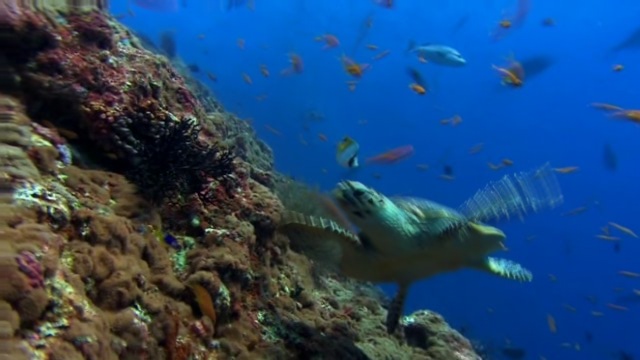}
        \end{subfigure}
        \hfill
        \begin{subfigure}[t]{0.327\columnwidth}
            \includegraphics[width=1.0\columnwidth]{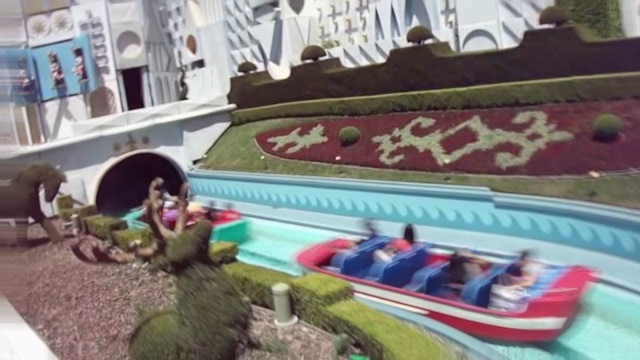}
        \end{subfigure}
        \hfill
        \begin{subfigure}[t]{0.327\columnwidth}
            \includegraphics[width=1.0\columnwidth]{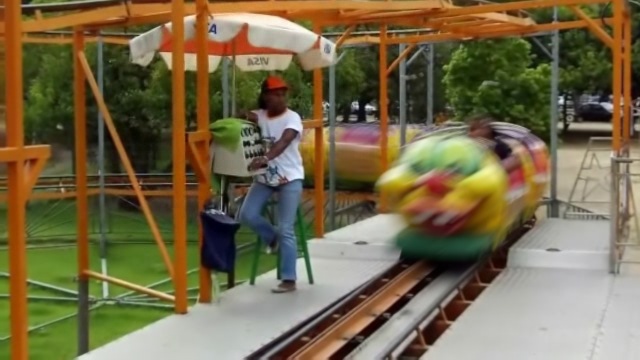}
        \end{subfigure}
        \myvspace{-1.5em}
        \subcaption{Our pipeline, without using the segmentation mask data.}
        \myvspace{0.25em}
        \label{subfig:evaluation_no_mask}
    \end{subfigure}

    \begin{subfigure}[t]{\columnwidth}
        \begin{subfigure}[t]{0.327\columnwidth}
            \includegraphics[width=1.0\columnwidth]{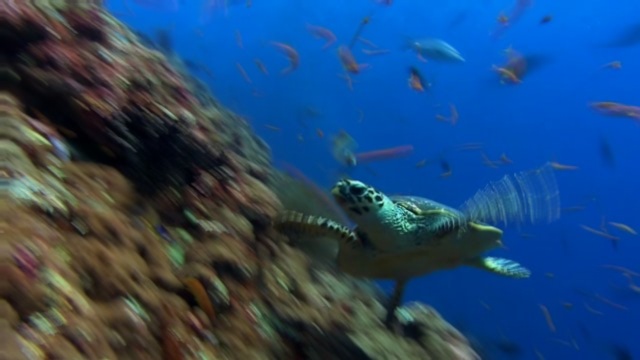}
        \end{subfigure}
        \hfill
        \begin{subfigure}[t]{0.327\columnwidth}
            \includegraphics[width=1.0\columnwidth]{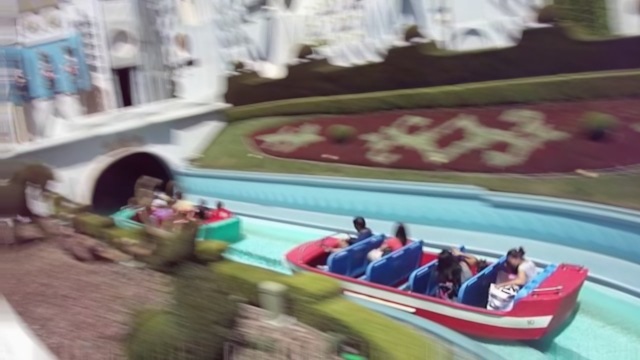}
        \end{subfigure}
        \hfill
        \begin{subfigure}[t]{0.327\columnwidth}
            \includegraphics[width=1.0\columnwidth]{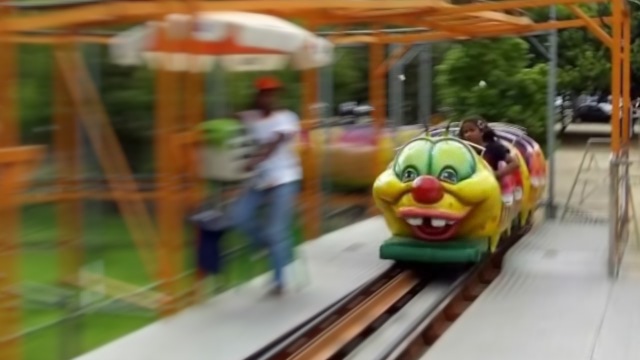}
        \end{subfigure}
        \myvspace{-1.5em}
        \subcaption{Our pipeline, using a manual saliency override mask.}
        \myvspace{-1em}
        \label{subfig:evaluation_saliency_override}
    \end{subfigure}

    \caption{\tacc{Several examples showing our method evaluated on the input video dataset in~\cite{Liu2014}. We include a few examples in (\subref{subfig:evaluation_no_mask}) middle-right and lower row, where our subject detection leads to a different outcome vs. their 3D aware segmentation-driven approach. We add corresponding comparable results in (\subref{subfig:evaluation_saliency_override}), obtained by manually overriding our saliency signal with a subject mask.}}
    \myvspace{-1em}
    \label{fig:evaluation}
\end{figure}

Our automatic subject detection aligns the result on people's faces when they are detected, and on visually salient features otherwise, which matches the selected subject from previous work in many scenes. When multiple faces are detected, our method chooses to align on the largest face, which may lead to a different outcome (\fig{subfig:evaluation_no_mask} middle-right and lower-right examples). We also observe a possible mismatch when no faces are present or are too small to be detected, e.g. while our saliency signal reacts to the most colorful or brightest areas of the image (\fig{subfig:evaluation_no_mask} lower-left and lower-middle examples respectively).

Even though we use a simple 2D image alignment approach (see \sect{subsec:alignment_bg_blur}), our method leads to comparable subject stabilization in most cases. Our similarity transform solver is able to preserve subject sharpness and models a relative virtual camera motion that is similar to that of compared works and is sometimes more accurate (\fig{subfig:evaluation_no_mask} center and \fig{subfig:evaluation_saliency_override} right examples).

Our rendering approach is most similar to~\cite{Lancelle19} but differs in the implementation to interpolate motion between frames. Our method scales to very high resolution images in an efficient manner on a mobile device, and we find the resulting motion-blur quality to be comparable. Both works benefit from integrating motion-blur from the input images spanning the whole time-interval, unlike the approach in~\cite{Liu2014}, which uses a spatially-varying blur of only the base frame. Our method correctly renders the dynamic motion of the scene handling dis-occlusions, showing other moving scene objects, people's moving limbs, and the motion of the background, all relative to the subject and as seen through the virtual camera aligned over time. In contrast, blurring only the base frame assumes the scene is static and only the aligned camera transformation affects the blur. This is most noticeable when comparing our turtle result in \fig{subfig:evaluation_saliency_override}-left to theirs. Our system renders the relative coral motion direction correctly, as can be seen in the input video, and shows the turtle's moving fin and the swirling individual motion of surrounding fish.

The amount of blur is normalized by our frame selection algorithm described in Section ~\ref{sec:frame_selection}. Our stylistic background blur length target is shorter than the results in~\cite{Lancelle19}, and is motivated by the goal to preserve subject sharpness and scene context.

}

\section{Limitations and Future Work}

Background blur scenes with very small subjects tend to significantly increase the occurrence of saliency and portrait mask mispredictions and feature tracking errors, ultimately resulting in an undesirable alignment and preserving the sharpness of the incorrect image region. Although our system can handle reasonably small subjects, this problem can be improved further by refining these predictions using the appropriate sub-region of the input images down-sampled at a higher resolution.

Our motion prediction model with receptive field window of 128 pixels can handle approximately 64 pixels of motion disparity at the chosen input low resolution. In our system, this corresponds to 512 pixels of disparity at full resolution, which is a good practical upper bound when capturing 12 megapixel bursts at 30fps. Larger motion disparities across frame pairs cause inaccurate predicted kernel maps and result in significant artifacts in the synthesized motion-blur. When these rare cases are detected in the motion analysis stage of our pipeline, we decide to output only the sharp exposure.

None of the models we tested perfectly handle motion silhouettes, when the foreground and background are moving between perpendicular and opposite directions or in the presence of large motion, causing swirly looking or disocclusion artifacts. We also notice similar artifacts in areas where a cast shadow moves in a different direction than the object where the shadow is projected onto. Some examples can be found in the supplementary material \tacc{in Figure 1}. Resolving these challenging cases is left for future work.

\myvspace{-0.5em}

\section{Conclusion}

In this paper, we described a long exposure computational photography system, that is able to produce high quality long exposure foreground or background blur effects. Our system operates in a smartphone camera app and outputs both long and conventional exposures fully automatically, in just a few seconds after pressing the shutter button. We described the key elements that make our system successful: adapting the burst capture schedule to scene motion velocity, separating the main subject from the background and tracking their motion, creating custom aesthetic alignments of input burst images, synthesizing smooth curved motion-blur spanning multiple underexposed \tacc{HDR} sharp input images, and compositing sharp and motion-blurred results to protect important scene locations - exceeding what would be physically possible with traditional long-exposure photography. The careful combination of these elements gives our system the ability to produce aesthetically pleasing blur trails and to preserve sharpness where it is most needed. The end-to-end integration of our system into a consumer mobile device camera makes it possible for casual users to access a creative style previously reserved to more advanced photographers.

\myvspace{-0.5em}

\begin{acks}

\ifanonymized
Withheld for anonymous review.
\else

\tacc{"Motion Mode" and the work described in this paper is the result of many individuals' effort in Google Research's Creative Camera team and the Google Camera App team, with help from many collaborating partners in the Gcam, reModel, Pixel and Silicon teams. We are especially thankful to Janne Kontkanen for leading the research on FILM, to Gabriel Nava, Shamvi Punja, Qiurui He, Alex Schiffhauer, Steven Hickson, Po-Ya Hsu, Zhijie Deng, Yen-Hsiang Huang, Kiran Murthy and Sam Hasinoff for their contributions, and to Tim Brooks and Jon Barron for early research. Credit for photography and image quality evaluations goes to our staff photographers Michael Milne, Andy Radin, Nicholas Wilson, Brandon Ruffin, as well as Chris Haney, Sam Robinson, Christopher Farro and Cort~Muller. We are also very grateful to Orly Liba, Cassidy Curtis, Kfir Aberman and David Salesin for their help reviewing early drafts of this paper.}

\fi

\end{acks}

\bibliography{main}

\clearpage

\begin{figure*}[ht]
    \centering
    \hfill
    \begin{subfigure}[t]{0.497\textwidth}
        \begin{subfigure}[t]{\columnwidth}
            \begin{tikzpicture}[node distance=0,outer sep=0,spy using outlines]
              \node[anchor=south](FigA) at (0,0) {\includegraphics[height=4.7cm]{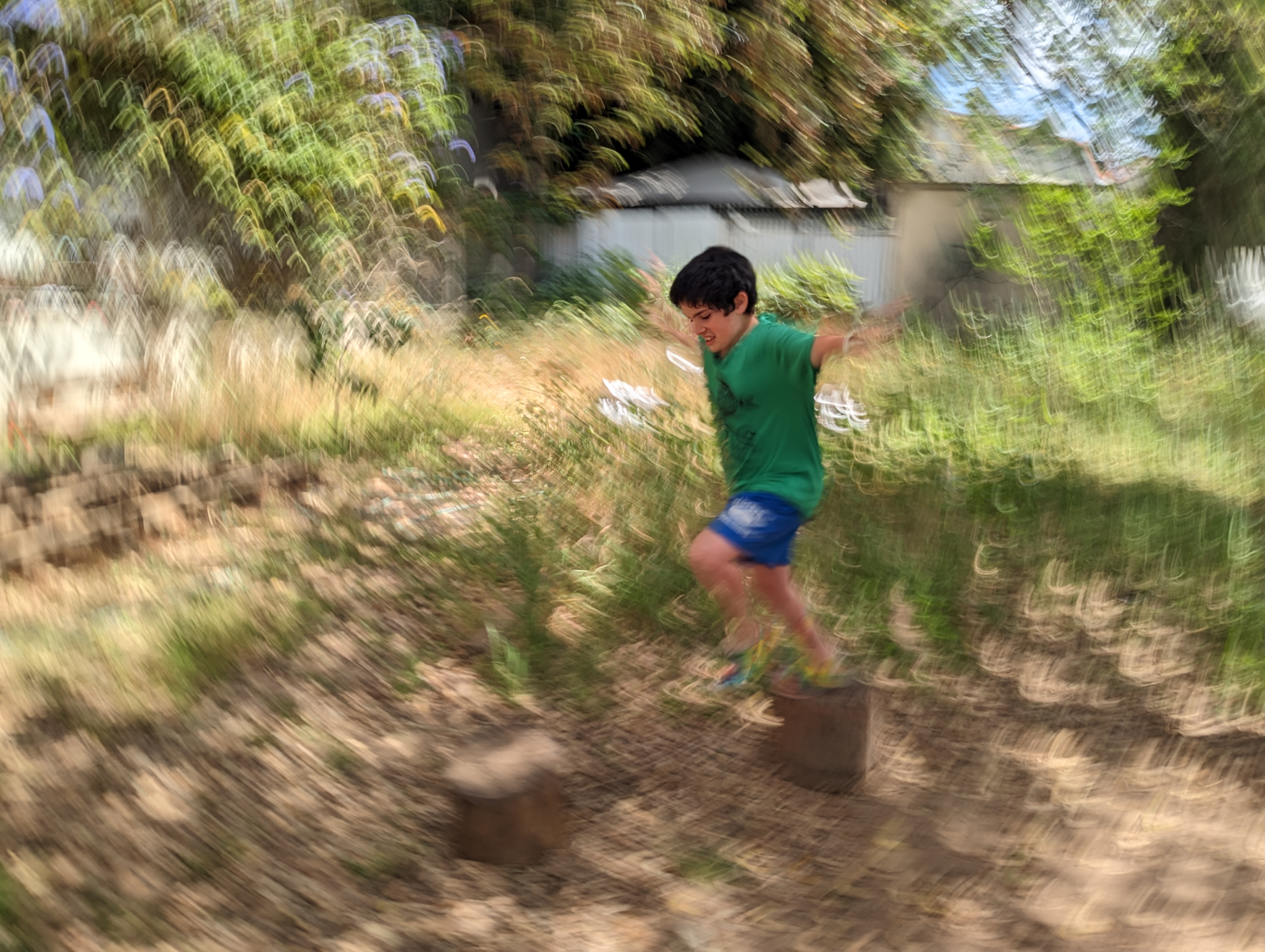}};
              \spy [aalcloseup1,magnification=5, every spy on node/.append style={ultra thick}] on ($(FigA)+( -0.9, 1.45)$) 
                in node[aallargewindow1,anchor=west] at ($(FigA.east)+(0, 1.2)$);
              \spy [aalcloseup2,magnification=5, every spy on node/.append style={ultra thick}] on ($(FigA)+( 1.85, -0.5)$) 
                in node[aallargewindow2,anchor=west] at ($(FigA.east)+(0, -1.2)$);
            \end{tikzpicture}
        \end{subfigure}
        \begin{subfigure}[t]{\columnwidth}
            \begin{tikzpicture}[node distance=0,outer sep=0,spy using outlines]
              \node[anchor=south](FigA) at (0,0) {\includegraphics[height=4.65cm]{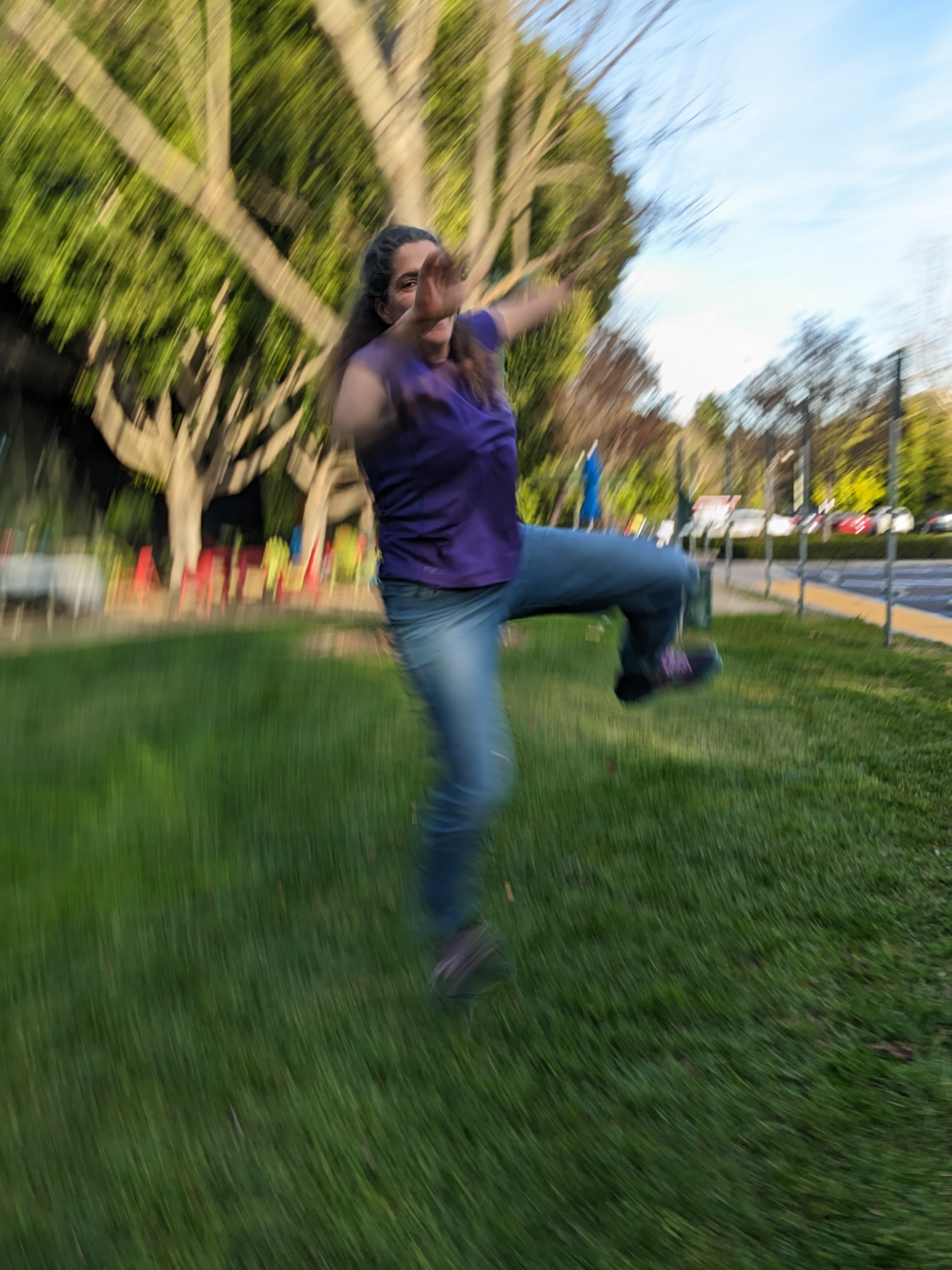}};
              \spy [aapcloseup1,magnification=5, every spy on node/.append style={ultra thick}] on ($(FigA)+( -1.4, 1.95)$) 
                in node[aaplargewindow1,anchor=west] at ($(FigA.east)+(0, 1.2)$);
              \spy [aapcloseup3,magnification=5, every spy on node/.append style={ultra thick}] on ($(FigA)+( 1.45, -0.5)$)
                in node[aaplargewindow3,anchor=west] at ($(FigA.east)+(2.4, 0.0)$);
              \spy [aapcloseup2,magnification=5, every spy on node/.append style={ultra thick}] on ($(FigA)+( -1.4, -1.95)$) 
                in node[aaplargewindow2,anchor=west] at ($(FigA.east)+(0, -1.2)$);
            \end{tikzpicture}
        \end{subfigure}
        \begin{subfigure}[t]{\columnwidth}
            \begin{tikzpicture}[node distance=0,outer sep=0,spy using outlines]
              \node[anchor=south](FigA) at (0,0) {\includegraphics[height=4.7cm]{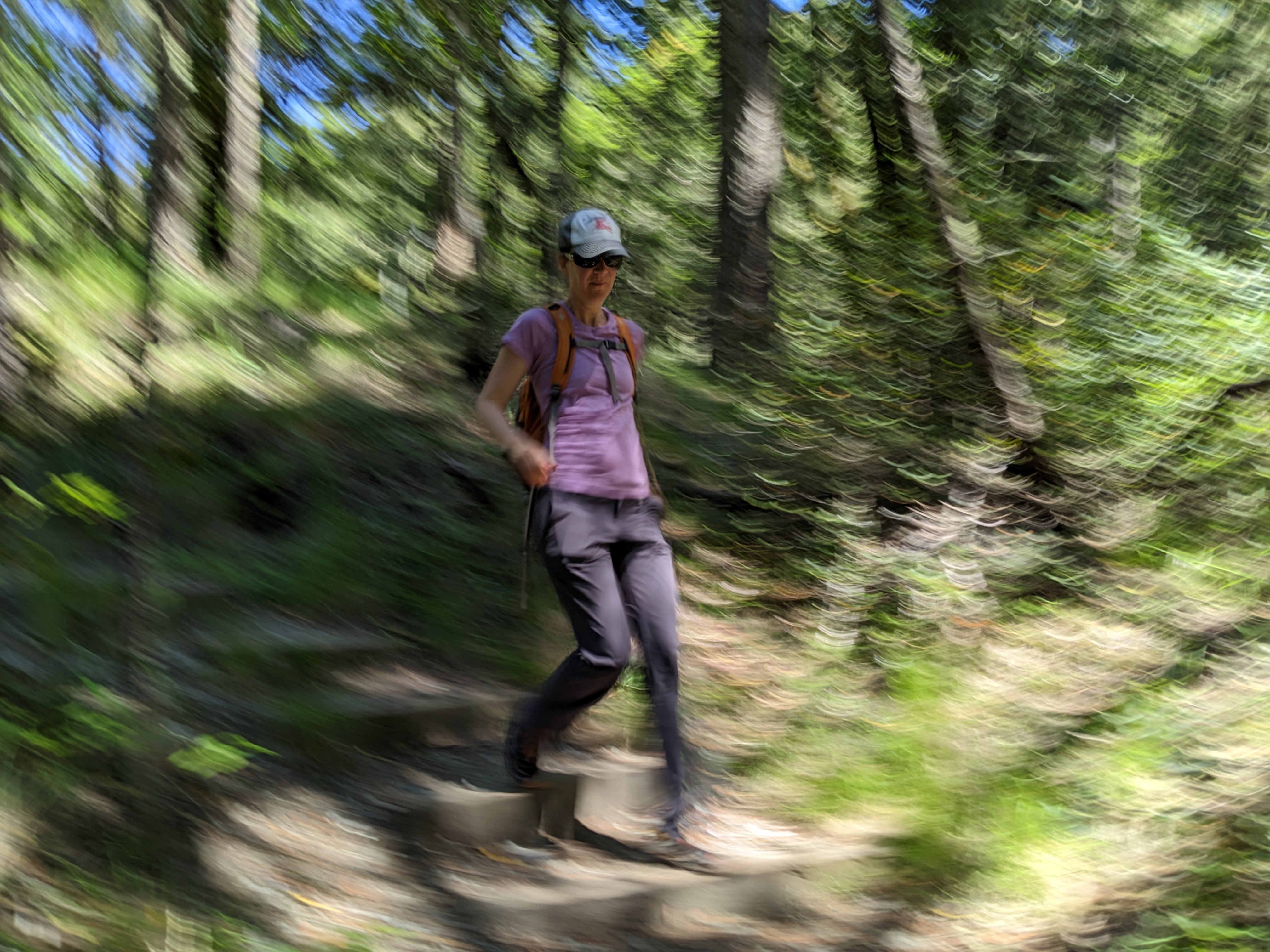}};
              \spy [aalcloseup1,magnification=5, every spy on node/.append style={ultra thick}] on ($(FigA)+( 1.45, 1.9)$) 
                in node[aallargewindow1,anchor=west] at ($(FigA.east)+(0, 1.2)$);
              \spy [aalcloseup2,magnification=5, every spy on node/.append style={ultra thick}] on ($(FigA)+( -2.4, 1.55)$) 
                in node[aallargewindow2,anchor=west] at ($(FigA.east)+(0, -1.2)$);
            \end{tikzpicture}
        \end{subfigure}
    \end{subfigure}
    \begin{subfigure}[t]{0.497\textwidth}
        \begin{subfigure}[t]{\columnwidth}
            \begin{tikzpicture}[node distance=0,outer sep=0,spy using outlines]
              \node[anchor=south](FigA) at (0,0) {\includegraphics[height=4.7cm]{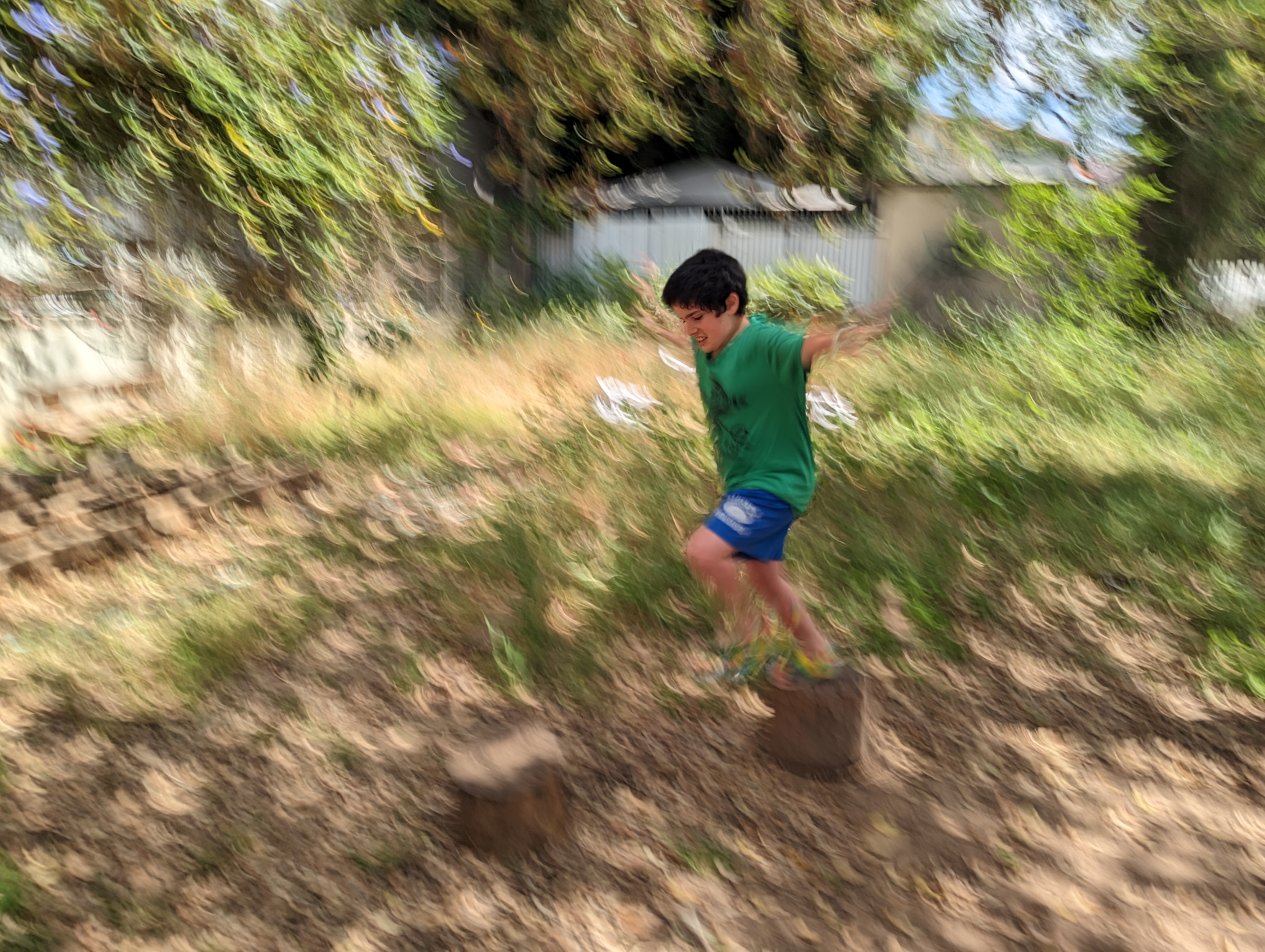}};
              \spy [aalcloseup1,magnification=5, every spy on node/.append style={ultra thick}] on ($(FigA)+( -0.9, 1.45)$) 
                in node[aallargewindow1,anchor=west] at ($(FigA.east)+(0, 1.2)$);
              \spy [aalcloseup2,magnification=5, every spy on node/.append style={ultra thick}] on ($(FigA)+( 1.8, -0.5)$) 
                in node[aallargewindow2,anchor=west] at ($(FigA.east)+(0, -1.2)$);
            \end{tikzpicture}
        \end{subfigure}
        \begin{subfigure}[t]{\columnwidth}
            \begin{tikzpicture}[node distance=0,outer sep=0,spy using outlines]
              \node[anchor=south](FigA) at (0,0) {\includegraphics[height=4.65cm]{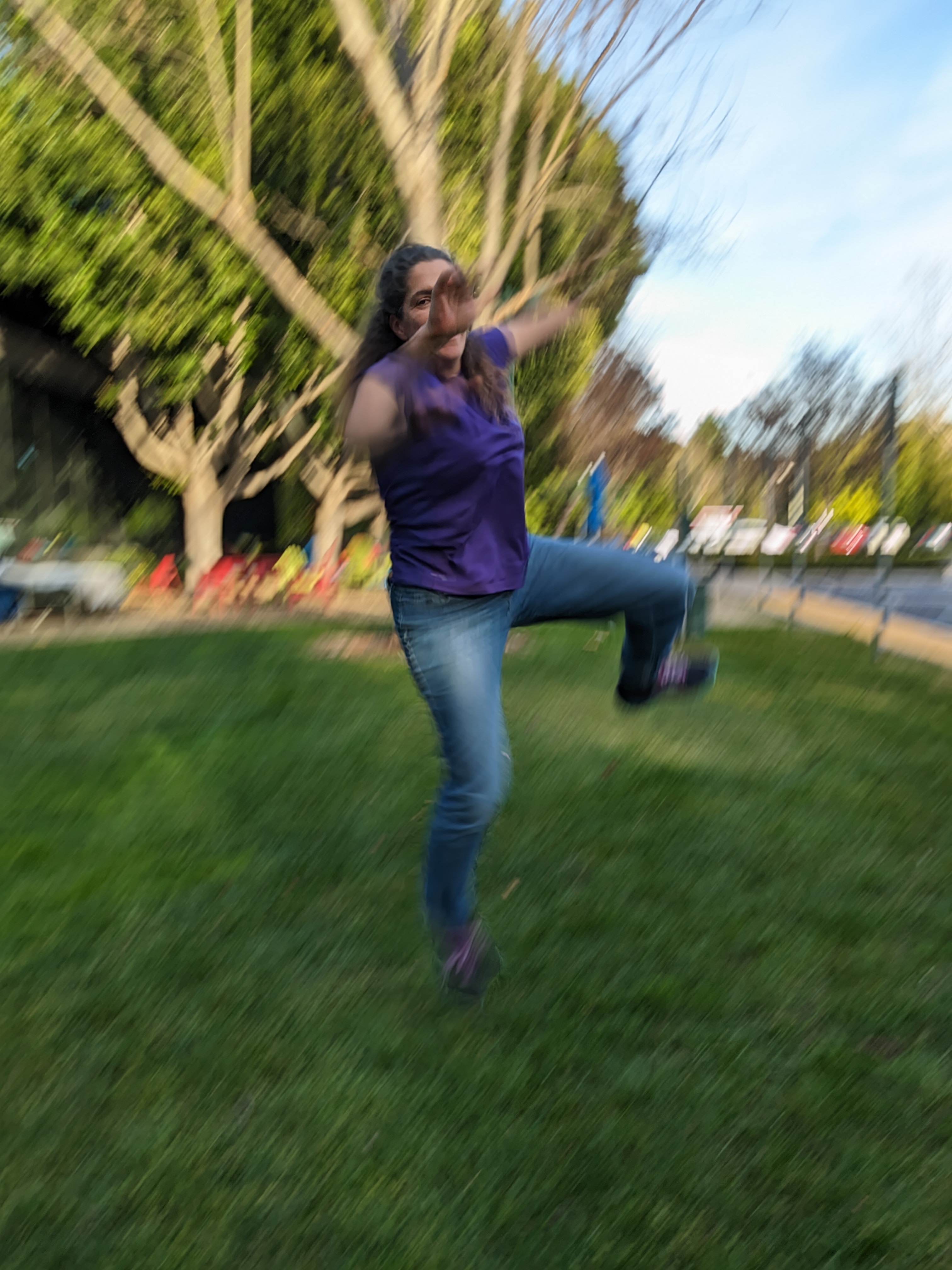}};
              \spy [aapcloseup1,magnification=5, every spy on node/.append style={ultra thick}] on ($(FigA)+( -1.4, 1.95)$) 
                in node[aaplargewindow1,anchor=west] at ($(FigA.east)+(0, 1.2)$);
              \spy [aapcloseup3,magnification=5, every spy on node/.append style={ultra thick}] on ($(FigA)+( 1.45, -0.5)$)
                in node[aaplargewindow3,anchor=west] at ($(FigA.east)+(2.4, 0.0)$);
              \spy [aapcloseup2,magnification=5, every spy on node/.append style={ultra thick}] on ($(FigA)+( -1.4, -1.95)$) 
                in node[aaplargewindow2,anchor=west] at ($(FigA.east)+(0, -1.2)$);
            \end{tikzpicture}
        \end{subfigure}
        \begin{subfigure}[t]{\columnwidth}
            \begin{tikzpicture}[node distance=0,outer sep=0,spy using outlines]
              \node[anchor=south](FigA) at (0,0) {\includegraphics[height=4.7cm]{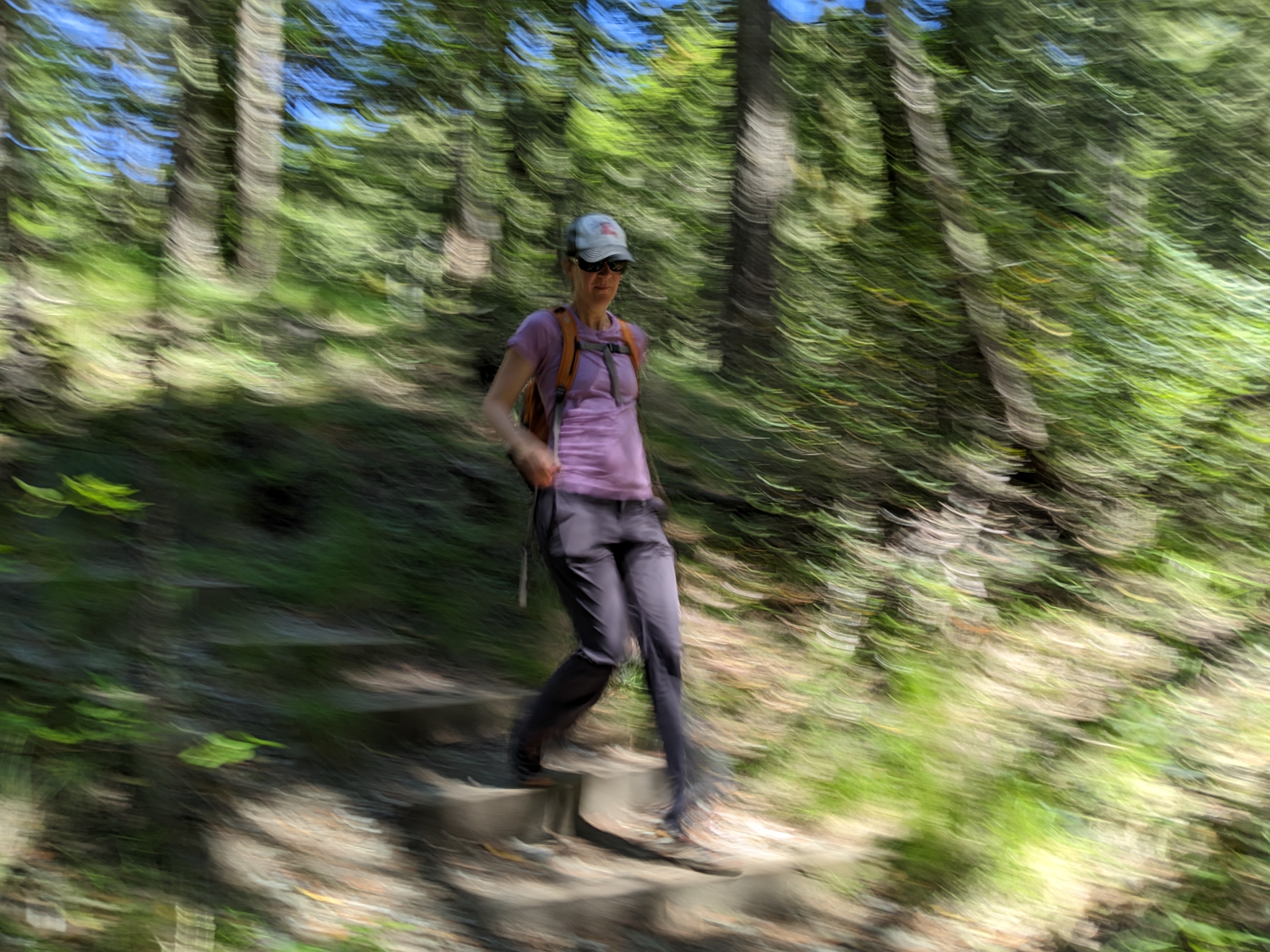}};
              \spy [aalcloseup1,magnification=5, every spy on node/.append style={ultra thick}] on ($(FigA)+( 1.45, 1.9)$) 
                in node[aallargewindow1,anchor=west] at ($(FigA.east)+(0, 1.2)$);
              \spy [aalcloseup2,magnification=5, every spy on node/.append style={ultra thick}] on ($(FigA)+( -2.4, 1.55)$) 
                in node[aallargewindow2,anchor=west] at ($(FigA.east)+(0, -1.2)$);
            \end{tikzpicture}
        \end{subfigure}
    \end{subfigure}
    \hfill
    \setlength{\belowcaptionskip}{-1\baselineskip}
    \caption{Background blur aesthetics visualized across different parts of the field of view (red and blue insets) comparing results without (left), and with (right) the regularization term $E_b$ from Eq.~\ref{eqn:alignment_objective_fn}. By additionally imposing the rotational constraints from \sect{sec:image_alignment}, we remove an undesirable sharp region surrounded by a blur vortex (green inset). Insets are displayed at 5x magnification.}
    \label{fig:alignment_ablations}
\end{figure*}


\begin{figure*}[ht]
    \centering
    \begin{subfigure}[t]{0.163\textwidth}
        \includegraphics[width=1.0\columnwidth]{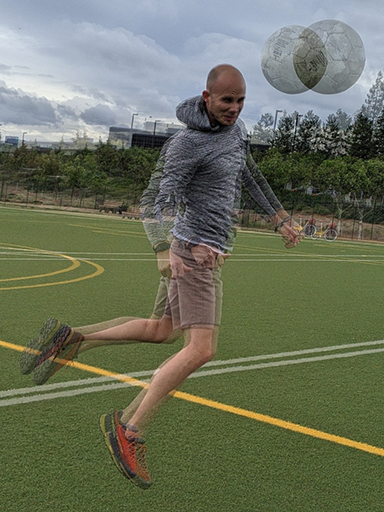}
        \includegraphics[width=1.0\columnwidth]{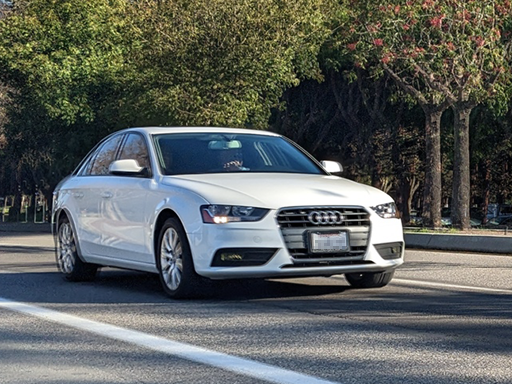}
        \includegraphics[width=1.0\columnwidth]{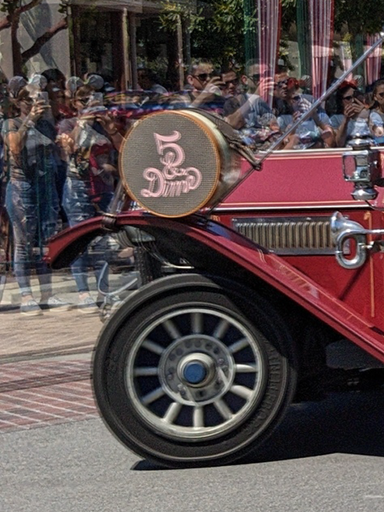}
        \includegraphics[width=1.0\columnwidth]{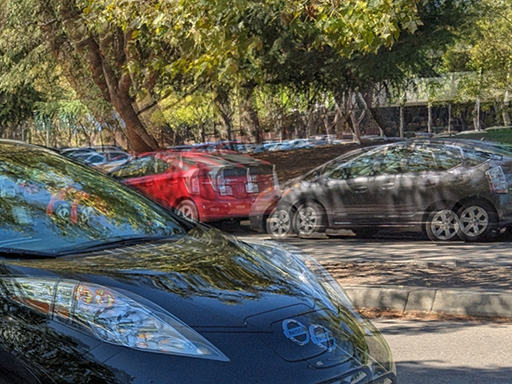}
        \subcaption{Input average}
    \end{subfigure}
    \hfill
    \begin{subfigure}[t]{0.163\textwidth}
        \includegraphics[width=1.0\columnwidth]{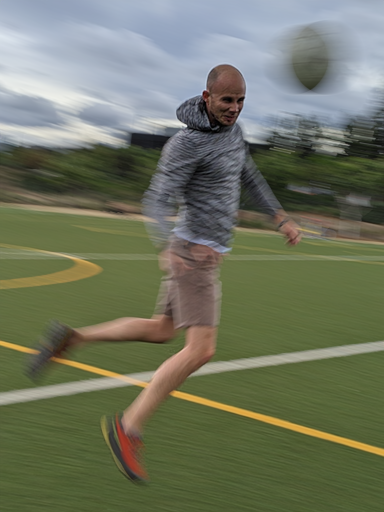}
        \includegraphics[width=1.0\columnwidth]{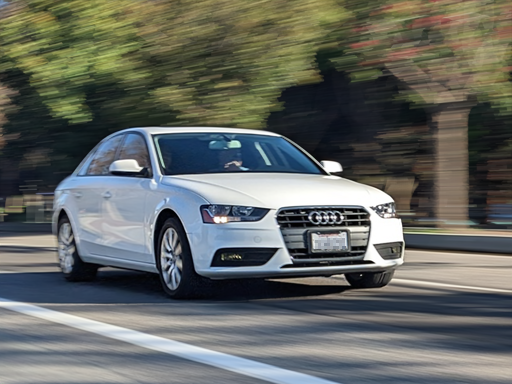}
        \includegraphics[width=1.0\columnwidth]{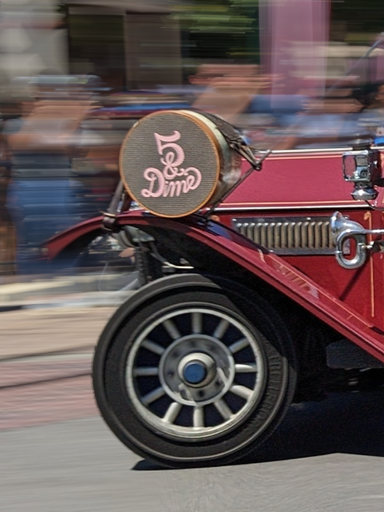}
        \includegraphics[width=1.0\columnwidth]{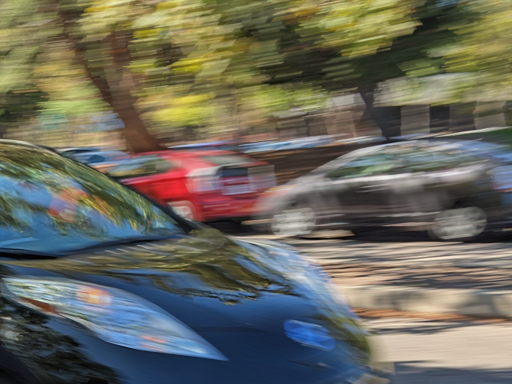}
        \subcaption{FILM (ground truth)}
    \end{subfigure}
    \hfill
    \begin{subfigure}[t]{0.163\textwidth}
        \includegraphics[width=1.0\columnwidth]{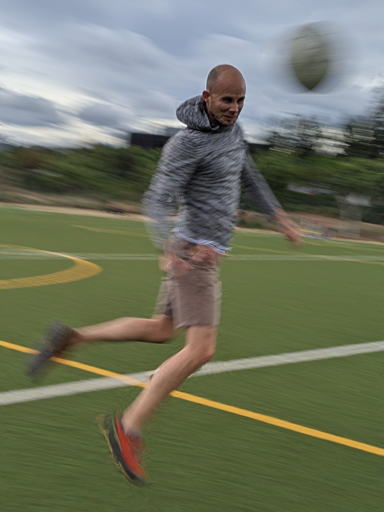}
        \includegraphics[width=1.0\columnwidth]{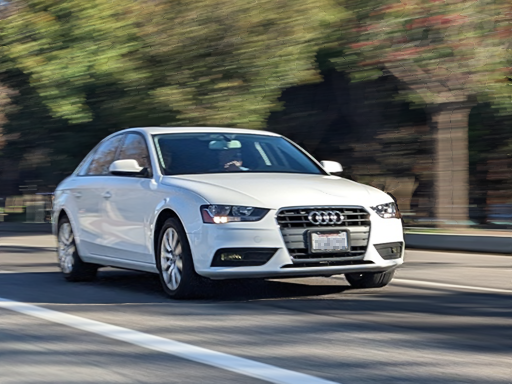}
        \includegraphics[width=1.0\columnwidth]{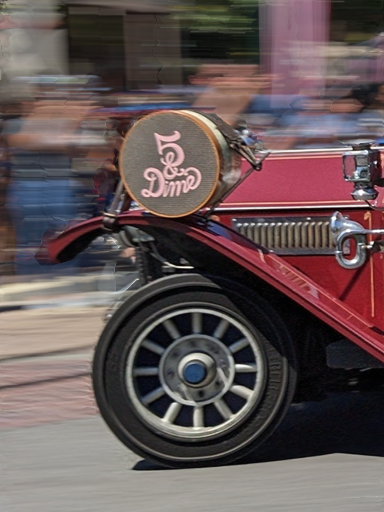}
        \includegraphics[width=1.0\columnwidth]{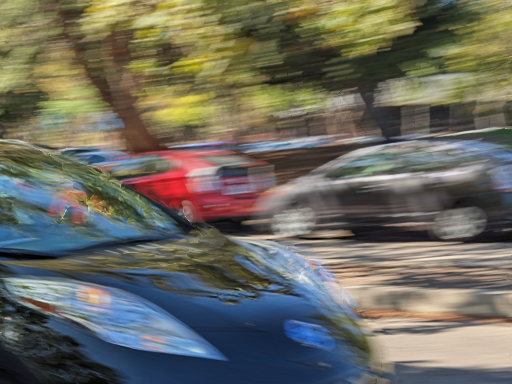}
        \subcaption{BB19-uni.}
    \end{subfigure}
    \hfill
    \begin{subfigure}[t]{0.163\textwidth}
        \includegraphics[width=1.0\columnwidth]{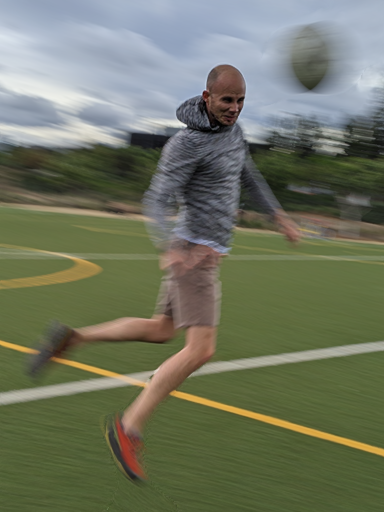}
        \includegraphics[width=1.0\columnwidth]{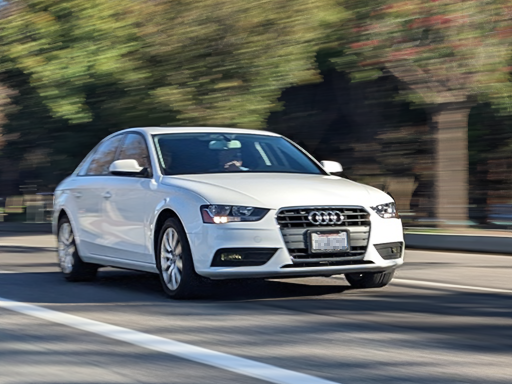}
        \includegraphics[width=1.0\columnwidth]{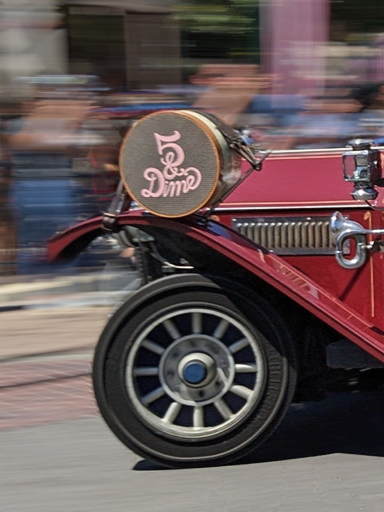}
        \includegraphics[width=1.0\columnwidth]{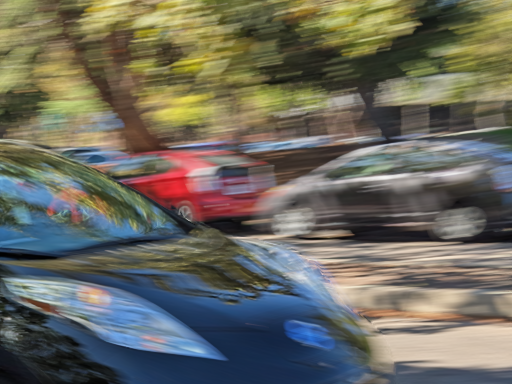}
        \subcaption{BB19}
    \end{subfigure}
    \hfill
    \begin{subfigure}[t]{0.163\textwidth}
        \includegraphics[width=1.0\columnwidth]{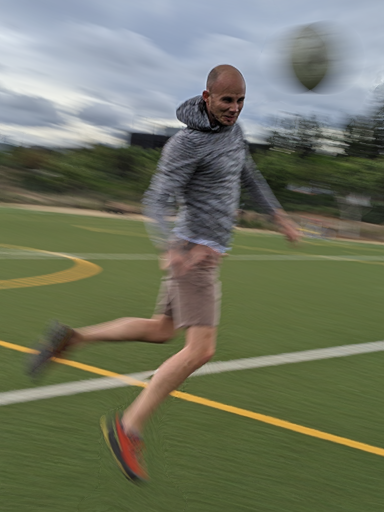}
        \includegraphics[width=1.0\columnwidth]{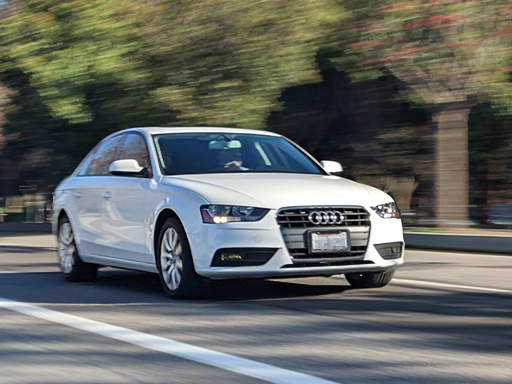}
        \includegraphics[width=1.0\columnwidth]{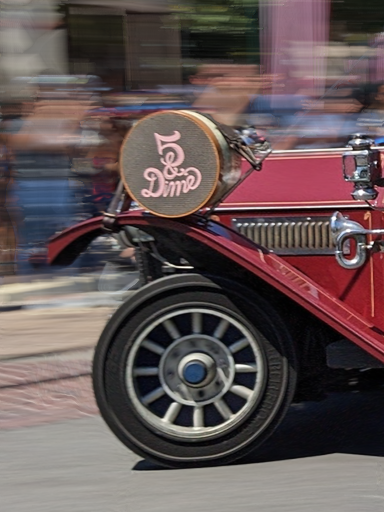}
        \includegraphics[width=1.0\columnwidth]{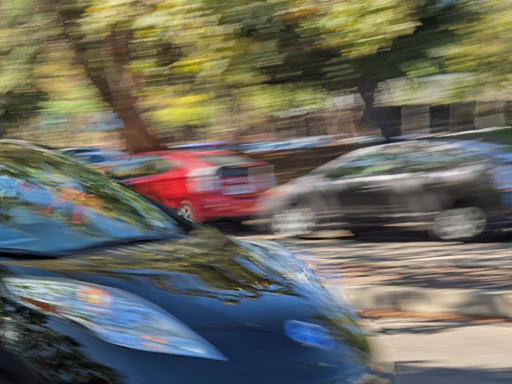}
        \subcaption{Ours}
    \end{subfigure}
    \hfill
    \begin{subfigure}[t]{0.163\textwidth}
        \includegraphics[width=1.0\columnwidth]{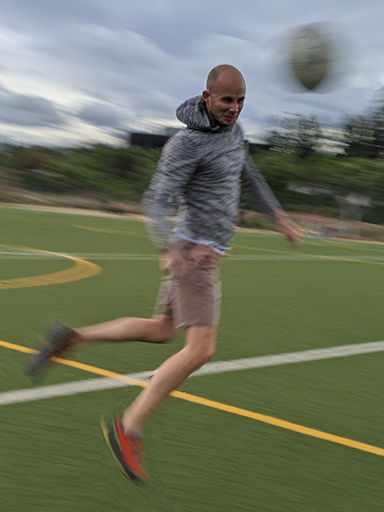}
        \includegraphics[width=1.0\columnwidth]{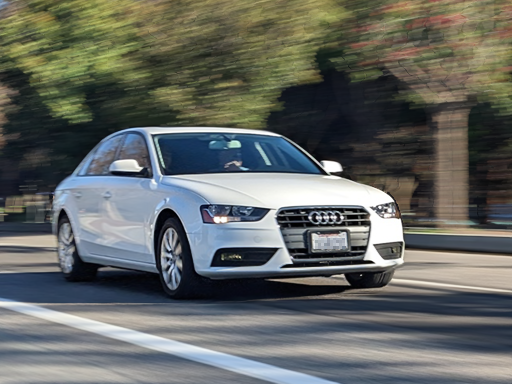}
        \includegraphics[width=1.0\columnwidth]{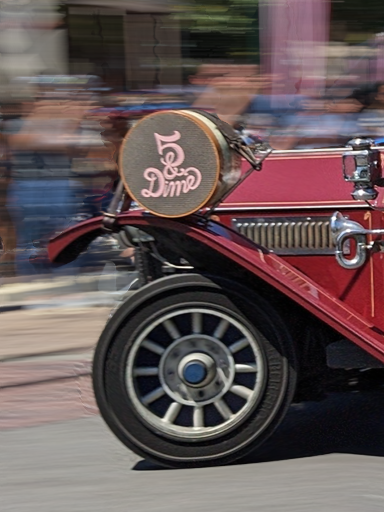}
        \includegraphics[width=1.0\columnwidth]{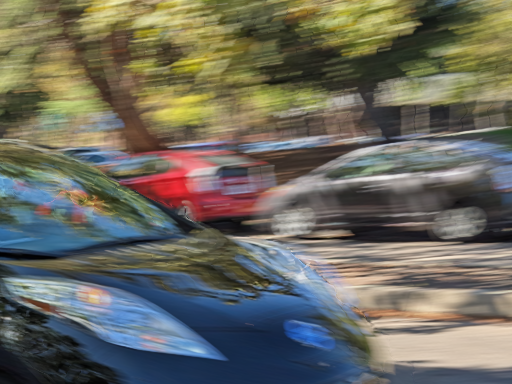}
        \subcaption{Ours-abl.}
    \end{subfigure}
    \caption{Model comparisons from single input image pairs shown overlaid in column (a). Column (b) is rendered using recursive frame interpolation with the FILM model from~\cite{Reda22}, and is used as ground truth when training our models. Columns (c) and (d) are rendered with the model from~\cite{Brooks19}, with uniform and learned weights respectively. Columns (e) and (f) are rendered with our model, with and without the ramp function \(w_n\) respectively. Differences are subtle, showing that our mobile model simplifications do not affect image quality substantially. The examples on each row showcase blur quality and disocclusions with various amount of motion disparity. The last row contains disocclusions with opposite motion. Several additional challenging examples can be found in the supplementary material, we encourage the reader to compare the accompanying images, to see differences more clearly.}
    \label{fig:model_comparisons}
\end{figure*}

\end{document}